\documentclass{article}

\usepackage{amsmath,amsfonts,bm}

\def\eqref#1{equation~\ref{#1}}
\def\Eqref#1{Equation~\ref{#1}}

\def\1{\bm{1}}

\DeclareMathAlphabet{\mathsfit}{\encodingdefault}{\sfdefault}{m}{sl}
\SetMathAlphabet{\mathsfit}{bold}{\encodingdefault}{\sfdefault}{bx}{n}

\DeclareMathOperator*{\argmin}{arg\,min}

\usepackage{microtype}
\usepackage{graphicx}
\usepackage{subfigure}
\usepackage{booktabs} 

\usepackage[hyphens]{url}
\usepackage{hyperref}

\usepackage[accepted]{icml2024}

\usepackage{amsmath}
\usepackage{amssymb}
\usepackage{mathtools}
\usepackage{amsthm}

\usepackage{tabularx}

\definecolor{deeppink}{rgb}{1.0, 0.08, 0.58}
\usepackage{pgf}
\usepackage{pgfplots}\usepackage{pgf}
\pgfplotsset{every tick label/.append style={font=\scriptsize}}
\pgfplotsset{compat=newest}
\pgfplotsset{every axis legend/.append style={font=\footnotesize}}
\usepgfplotslibrary{groupplots,dateplot}
\usetikzlibrary{patterns,shapes.arrows}

\usepackage{colortbl}
\usepackage{multirow}

\usepackage[symbol]{footmisc}

\newtheorem*{problem*}{Problem}
\newtheorem*{orthogonal_assumption*}{Orthogonality Assumption}
\newtheorem*{linear_assumption*}{Linearity Assumption}
\newcommand{\littlespace}{\hspace{0.2cm}}

\errorcontextlines\maxdimen

\newcommand{\samplemeang}[1]{\frac{1}{n_g}\sum^{n_g}_{i=1}d_{i, g}}

\newcommand{\var}[1]{\mathrm{Var}\left( #1 \right) }

\newcommand{\overallestimator}{\bar{d}_{w}}
\newcommand{\overallestimatorsum}{\frac{1}{4}\sum^{4}_{g=1}\bar{d}_g}

\newcommand{\bx}{\boldsymbol{x}}
\newcommand{\ymt}{y_\mathrm{mt}}
\newcommand{\ymtsub}[1]{y_{\mathrm{mt},{#1}}}
\newcommand{\ysp}{y_\mathrm{sp}}
\newcommand{\yspsub}[1]{y_{\mathrm{sp},{#1}}}
\newcommand{\yhatmt}{\hat{y}_\mathrm{mt}}
\newcommand{\yhatmtsub}[1]{\hat{y}_{\mathrm{mt},{#1}}}
\newcommand{\yhatsp}{\hat{y}_\mathrm{sp}}
\newcommand{\yhatspsub}[1]{\hat{y}_{\mathrm{sp},{#1}}}
\newcommand{\xmt}{\boldsymbol{x}_\mathrm{mt}}
\newcommand{\xsp}{\boldsymbol{x}_\mathrm{sp}}

\newcommand{\bz}{\boldsymbol{z}}
\newcommand{\bZ}{\boldsymbol{Z}}
\newcommand{\zmt}{\boldsymbol{z}_\mathrm{mt}}
\newcommand{\zsp}{\boldsymbol{z}_\mathrm{sp}}
\newcommand{\Zmt}{\mathcal{Z}_\mathrm{mt}}
\newcommand{\Zsp}{\mathcal{Z}_\mathrm{sp}}

\newcommand{\Vmt}{\boldsymbol{V}_\mathrm{mt}}
\newcommand{\Vsp}{\boldsymbol{V}_\mathrm{sp}}
\newcommand{\Vhatmt}{\hat{\boldsymbol{V}}_\mathrm{mt}}
\newcommand{\Vhatsp}{\hat{\boldsymbol{V}}_\mathrm{sp}}
\newcommand{\vmt}{\boldsymbol{v}_\mathrm{mt}}
\newcommand{\vsp}{\boldsymbol{v}_\mathrm{sp}}
\newcommand{\vhatmt}{\hat{\boldsymbol{v}}_\mathrm{mt}}
\newcommand{\vhatmtsub}[1]{\hat{\boldsymbol{v}}_{\mathrm{mt},{#1}}}
\newcommand{\vhatsp}{\hat{\boldsymbol{v}}_\mathrm{sp}}
\newcommand{\vhatspsub}[1]{\hat{\boldsymbol{v}}_{\mathrm{sp},{#1}}}
\newcommand{\vspbasis}[1]{\boldsymbol{v}_{\mathrm{sp},{#1}}}

\newcommand{\wmt}{\boldsymbol{w}_\mathrm{mt}}
\newcommand{\wsp}{\boldsymbol{w}_\mathrm{sp}}
\newcommand{\whatmt}{\hat{\boldsymbol{w}}_\mathrm{mt}}
\newcommand{\whatsp}{\hat{\boldsymbol{w}}_\mathrm{sp}}

\newcommand{\bmt}{b_\mathrm{mt}}
\newcommand{\bsp}{b_\mathrm{sp}}
\newcommand{\bhatmt}{\hat{b}_\mathrm{mt}}
\newcommand{\bhatsp}{\hat{b}_\mathrm{sp}}

\newcommand{\dmt}{d_\mathrm{mt}}
\newcommand{\dsp}{d_\mathrm{sp}}

\newcommand{\Zsportho}{\boldsymbol{Z}_\mathrm{sp}^\perp}
\newcommand{\zsportho}{\boldsymbol{z}_{i, \mathrm{sp}}^{\perp^{\top}}}

\newcommand{\Zremain}{\boldsymbol{Z}_\mathrm{remain}}

\tikzstyle{Mytext} = [text width=0.5cm,text centered]

\usepackage[capitalize,noabbrev]{cleveref}

\theoremstyle{plain}

\theoremstyle{definition}

\theoremstyle{remark}

\newlength\figureheight
\newlength\figurewidth

\usepackage[textsize=tiny]{todonotes}

\icmltitlerunning{Removing Spurious Concepts from Neural Network Representations via Joint Subspace Estimation}

\begin{document}

\twocolumn[
\icmltitle{Removing Spurious Concepts from Neural Network \\ Representations via Joint Subspace Estimation}




\begin{icmlauthorlist}
\icmlauthor{Floris Holstege}{UvA,TI}
\icmlauthor{Bram Wouters}{UvA}
\icmlauthor{Noud van Giersbergen}{UvA}
\icmlauthor{Cees Diks}{UvA,TI}

\end{icmlauthorlist}

\icmlaffiliation{UvA}{University of Amsterdam, Department of Quantitative Economics}
\icmlaffiliation{TI}{Tinbergen Institute}

\icmlcorrespondingauthor{Floris Holstege}{f.g.holstege@uva.nl}

\icmlkeywords{Machine Learning, ICML}

\vskip 0.3in
]



\printAffiliationsAndNotice{}  

\begin{abstract}

An important challenge in the field of interpretable machine learning is to ensure that deep neural networks (DNNs) use the correct or desirable input features in performing their tasks. Concept-removal methods aim to do this by eliminating concepts that are spuriously correlated with the main task from the neural network representation of the data. However, existing methods tend to be overzealous by inadvertently removing part of the correct or desirable features as well, leading to wrong interpretations and hurting model performance. We propose an iterative algorithm that separates spurious from main-task concepts by jointly estimating two low-dimensional orthogonal subspaces of the neural network representation. By evaluating the algorithm on benchmark datasets from computer vision (Waterbirds, CelebA) and natural language processing (MultiNLI), we show  it outperforms existing concept-removal methods in terms of identifying the main-task and spurious concepts, while removing only the latter.


\end{abstract}

\section{Introduction}

Although deep neural networks (DNNs) have achieved impressive results in computer vision and language modeling, it is notoriously difficult to control which concepts are being used by DNNs in performing their tasks. A \textit{concept} refers to a representation in the data of a human-defined object or phenomenon \citep{Been_2018}. For example, if a model's main task is to distinguish between images of cows and penguins, the training data might contain a spurious correlation between the concepts of background and animal type (cows typically appearing on grassland, and penguins typically in the snow). DNNs frequently rely on such spurious correlations within the data \citep{gururangan-etal-2018-annotation, pmlr-v119-srivastava20a, Wang_2020,  sagawa_2020B, zhou_2021}. For example, they use the background to identify the animal in the picture. This can be problematic in several ways. It makes models less trustworthy. Penguins on grassland could be classified as cows, for example. It also makes models difficult to interpret, as it is unclear to what extent a model relies on which concept. Finally, this could lead to models using undesirable concepts (e.g., race or gender). Therefore, we would like to control whether or not a model uses a specific concept in performing its task.




\begin{figure}[H]
    \centering
        \includegraphics[scale=0.24]{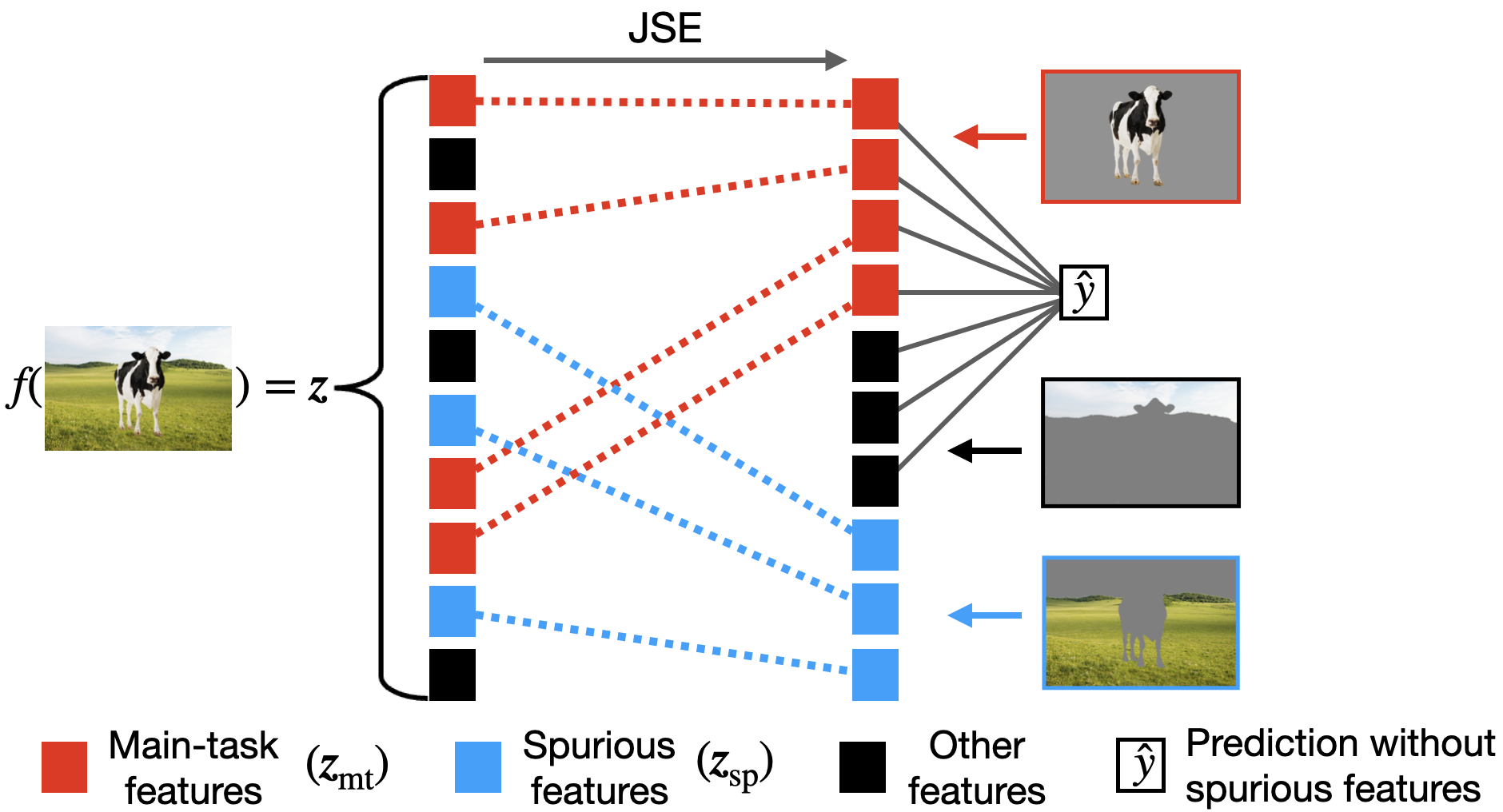}
       \caption{\textbf{High-level overview of Joint Subspace Estimation (JSE) for concept removal}: the input $\boldsymbol{x}$ is fed through a neural network $f(\boldsymbol{x})$, from which we can extract the vector representation $\boldsymbol{z}$. Within the vector representation, two orthogonal subspaces are identified: one related to the \textit{spurious concept} (e.g. the background), and one to the   \textit{main-task concept} (e.g. animal type).}
        \label{fig:high_level_illustration}
\end{figure}

Removing concepts directly from the training data (e.g., images or text) is non-trivial and costly. An alternative is to focus on the embeddings, which are vector representations of the data generated by the neural network. Post-hoc concept-removal methods aim to eliminate a concept from the embeddings, after the parameters of the neural network are frozen. Typically a concept classifier is trained on the embeddings, from which the concept features are then removed \citep{ravfogel_2020, rafvogel_2022}. In the example of distinguishing cows and penguins, we would train a classifier to predict the {spurious concept} of background type and use this to remove background features from the embeddings. Afterwards, a linear classifier is trained on the transformed embeddings to predict the {main-task concept} (e.g.~animal type), preventing the model from using the spurious concept for main-task classification. Post-hoc concept-removal methods in the presence of spurious correlations can be used for out-of-distribution (OOD) generalization \citep{joshi2022spurious} and for making models more interpretable \citep{elazar_2021}. They can also be used to remove sensitive information, such as gender \citep{bolukbasi_2016}.





A drawback of post-hoc concept-removal methods is that they tend to eliminate also main-task features from the embeddings \citep{Belinkov_2021, Kumar_2022}. This is because, due to the spurious correlation, a concept classifier might inadvertently use main-task features to predict the spurious concept. For example, a cow's horns could be used to predict a grassland background. As a consequence, using such a concept classifier to remove the spurious concept from the embeddings will also remove information associated with the main task. This potentially hurts main-task model performance and also could lead to wrong interpretations of the model after concept removal. It therefore limits the application of removal methods in the most relevant cases, namely when the spurious correlation is strong and likely to be exploited by the DNN.


\textbf{Our contribution}
: this paper proposes a novel post-hoc concept-removal method by jointly identifying two low-dimensional orthogonal subspaces, one associated with the spurious concept (e.g.~background) and the other with the main-task concept (e.g.~animal type). This crucially differs from existing methods, which only use the spurious concept to determine the subspace of spurious concept features. Furthermore, we make the identification of the subspaces systematic by introducing statistical tests that attribute directions in the embedding space to either the main-task or the spurious concept. The method, which we call Joint Subspace Estimation (JSE), is shown to be able to remove spurious features while retaining vital main-task information, also in the case of strong spurious correlations. Applied to benchmark datasets for image recognition (Waterbirds, CelebA) and natural language processing (MultiNLI), JSE is shown to outperform existing concept-removal methods in its ability to identify the main-task and spurious concepts, and to remove only the latter. A high-level overview of the method is given in Figure \ref{fig:high_level_illustration}. Our code with an implementation of JSE is publicly available.\footnote[1]{\url{https://github.com/fholstege/JSE}}

\section{Spurious Correlations and Concept Removal}

We consider the random variables $\mathcal{D} = ({y}_{\mathrm{mt}}, {y}_{\mathrm{sp}}, \boldsymbol{x} ),$ where $y_{\mathrm{mt}} \in \{0, 1\}$ is the main-task concept label, $y_{\mathrm{sp}} \in \{0, 1\}$ is the spurious concept label and $\boldsymbol{x} \in \mathcal{X}$ represents the input features. Each input $\boldsymbol{x}$ contains subsets $\boldsymbol{x}_{\mathrm{mt}}$ and $\boldsymbol{x}_{\mathrm{sp}}$ of features corresponding to the main-task and spurious concept, respectively. In the example of cows and penguins, $\xmt$ and $\xsp$ correspond to the pixels showing the animal and the background. It is assumed that $\boldsymbol{x}_{\mathrm{mt}}$ and $\boldsymbol{x}_{\mathrm{sp}}$ causally determine the associated labels $y_{\mathrm{mt}}$ and $y_{\mathrm{sp}}$, respectively. Their joint  joint probability density is then given by:
\begin{equation}\label{eq:joint_pdf}
      p\left(\ymt | \xmt\right)
    p\left(\ysp | \xsp\right)
    p\left(\xmt, \xsp\right).
\end{equation}
This implies that the main-task label $y_{\mathrm{mt}}$ is conditionally independent of the spurious features, $\xsp$, but it does not mean that $y_\mathrm{mt}$ and $\boldsymbol{x}_\mathrm{sp}$ are independent; they can be dependent due to dependence between $\boldsymbol{x}_\mathrm{mt}$ and $\boldsymbol{x}_\mathrm{sp}$. Since they are not causally related, we say they are spuriously correlated. At the level of trained neural networks, this means that a main-task classifier tends to make use of the spurious features $\xsp$ within $\bx$. Previous work offers a number of possible reasons, ranging from stochastic gradient descent (SGD) training dynamics \citep{pezeshki_2021} and overparameterization \citep{sagawa_2020B, amour_underspecify} to inductive biases of DNNs \citep{rahaman_2019}.

We now restrict our analysis to DNNs for classification, which typically is done using a complicated function $f(\boldsymbol{x}): \mathcal{X} \rightarrow \mathbb{R}^d$ mapping the input features to a vector representation, followed by a linear layer. We assume the embedding vectors $\bz \in \mathbb{R}^d$ have a similar structure as the input features $\bx,$ in the sense that each $\bz$ has subsets $\zmt \in \Zmt \subseteq \mathbb{R}^d$ and $\zsp \in \Zsp \subseteq \mathbb{R}^d$ that causally determine the labels $\ymt$ and $\ysp,$ respectively. It should be stressed that this does not necessarily hold in practice. The trained DNN could have mixed the different features, or have one of them removed, because of their predictive ability for the main-task label $\ymt.$ However, there is empirical evidence that when trained on data with a spurious correlation, neural networks  tend to learn both main-task and spurious features \citep{kirichenko_2022, izmailov2022feature, rosenfeld2022domainadjusted}. 

In addition, we assume that $\Zmt$ and $\Zsp$ are linear subspaces of the embedding space $\mathbb{R}^d.$ This sometimes goes under the name of {\it linear subspace hypothesis} \citep{bolukbasi_2016, vargas-cotterell-2020-exploring}. Previous work shows that linear subspaces can encode information about complex concepts \citep{bau_2017}. Moreover, non-linear information about the spurious concept cannot be used by the last layer for binary classification \citep{ravfogel2023loglinear}.

One possible concept-removal approach is to project the embedding vectors $\bz$ onto a linear subspace, before feeding them to a linear classifier. Suppose  $\vspbasis{1}, \vspbasis{2}, \ldots, \vspbasis{d_\mathrm{sp}}$ is an orthonormal basis of the spurious embedding subspace $\Zsp \subseteq \mathbb{R}^d$ and $\Vsp$ is the matrix $(d \times d_{\mathrm{sp}})$ whose columns are the basis vectors. Then the transformation from $\bz $ to $ (\boldsymbol{I} - \Vsp \Vsp^{\top}) \bz$ is the projection onto the orthogonal complement of  $\Zsp$ and thereby removes the spurious features from the representation. A linear classifier that uses the transformed embeddings to predict the binary main-task label $\ymt$ will not use the spurious features.

In practice, however, it is highly non-trivial to estimate (the basis of) the subspace $\Zsp.$ Due to the spurious correlations, classifiers that use the embedding vectors to predict the spurious label $\ysp$ also make use of the main-task embeddings $\zmt$. As a consequence, an estimate of $\Zsp$ will also contain directions that are actually part of $\Zmt.$ Projecting out the estimate of $\Zsp$ removes main-task information and therefore hurts the performance of the resulting main-task classifier \citep{ravfogel_2020, Belinkov_2021, Kumar_2022}. Our JSE method addresses this problem by estimating not only $\Zsp,$ but also the main-task embedding space $\Zmt.$ 

\subsection{Related Work} \label{sec:related_work}

\textbf{Concept-removal methods}: Concept removal was initially based on adversarial approaches \citep{goodfellow2014generative}, typically to mitigate undesirable biases \citep{Edwards_2016, lemoine_2018,balanced_dataset_not_enough, wang-etal-2021-dynamically}. This is frequently referred to as adversarial removal (ADV). However, the ability of these methods to remove concepts has been called into question \citep{elazar-goldberg-2018-adversarial}. An alternative is to remove a linear subspace from the embeddings \citep{bolukbasi_2016, ethayarajh_2019, dev_2019, dev-etal-2021-oscar}. A key method in this category is iterative null-space projection (INLP, \citeauthor{ravfogel_2020}, \citeyear{ravfogel_2020}), in which a linear classifier predicts the concept labels, and the coefficients of the classifier are orthogonally projected from the embeddings. This is repeated until the concept can no longer be predicted. A follow-up method is relaxed linear adversarial concept erasure (RLACE), in which an orthogonal projection matrix is trained such that the concept cannot be predicted \citep{rafvogel_2022}.  More recently, least-squares concept erasure (LEACE) was proposed, which provably prevents all linear classifiers from predicting concept labels \citep{belrose2023leace}.

\textbf{Interpretability}: An interpretable DNN obeys domain-specific constraints, allowing it to be better understood by humans \citep{Rudin_2021}. Concept-removal methods contribute to interpretability by allowing for greater control of which features are used by the DNN, as well as a better understanding of which features are used. In terms of control, concept-removal methods can be used to remove protected attributes, or sensitive information \citep{Xu_2017}. Regarding understanding, concept-removal methods can be used by observing how a neural network responds after the concept has been removed from its embeddings \citep{elazar_2021, ravfogel-etal-2021-counterfactual}. However, this approach has been called into question because they encode other information in addition to the concept \citep{Belinkov_2021, Kumar_2022}. JSE addresses this limitation, allowing greater control over spurious concepts, and understanding whether these are used by DNNs.

\textbf{Spurious correlations}: The problem of neural networks relying on spurious correlations has arisen in both computer vision \citep{geirhos2018imagenettrained, madry_2020_background, singla2022salient} and NLP \citep{Kao_2019, kaushik_2018, mccoy-etal-2019-right}. There is a wide range of methods addressing spurious correlations in neural networks, including data augmentation \citep{Hermann_2019}, invariant learning \citep{arjovsky2019, Ahuja_2021}, or instance-reweighting \citep{sagawa_2020, sagawa_2020B, idrissi2022simple}. The latter category is most akin to concept-removal methods, as it uses (limited) availability of spurious concept labels, but lacks the advantage of interpretability.


\section{Joint Subspace Estimation}

We will now introduce Joint Subspace Estimation (JSE) in which the  spurious and main-task embedding subspaces $\Zsp$ and $\Zmt$ are estimated simultaneously.  Section~\ref{sec:est} explains how to simultaneously estimate individual basis vectors for $\Zsp$ and $\Zmt$. Section~\ref{sec:iterative}  introduces an iterative procedure to find multiple basis vectors for  $\Zsp$ and $\Zmt$. Section~\ref{sec:testing} presents two statistical tests to stop the iterative procedure and determine the dimensions of $\Zsp$ and $\Zmt$.

\subsection{Estimating Spurious and Main-task Concept Vectors}
\label{sec:est}

As a starting point, consider simultaneously estimating one vector $\vsp \in \Zsp$ and another vector $\vmt \in \Zmt.$ A usual approach for estimating $\vsp$ is to train a logistic regression on the embeddings, $\hat{y}_\mathrm{sp} = \mathrm{Logit}^{-1} (\bz^{\top} \wsp + b_{\mathrm{sp}}),$ and then use the (normalized) coefficients $\vsp = \wsp / || \wsp ||$ as a so-called concept vector $\vsp$ that contains information about the concept \citep{Been_2018}. However, if we perform logistic regression in a sample where the spurious and main-task features are correlated, the estimate of $\vsp$ might have components in the direction of main-task features. To discourage the estimate of $\vsp$ to use main-task features (and vice versa for the estimate of $\vmt$), we make the following assumption about the relation between the two subspaces.

\begin{orthogonal_assumption*}
\label{assumption:orthogonality}
The linear subspaces $\Zsp$ and $\Zmt$ are orthogonal, i.e.\ each vector $\vsp \in \Zsp$ is perpendicular to each vector $\vmt \in \Zmt$.
\end{orthogonal_assumption*}

This is consistent with the assumption that the features determining the labels $\ysp$ and $\ymt$ are distinct \citep{park2023linear}, and the empirical observation that high-level concepts are distinctly represented in the embeddings \citep{kirichenko_2022}. We emphasize that orthogonality does not imply independence between main-task and spurious features, as assumed in earlier work \citep{rudin_2020}. 

\begin{figure*}
\centering
\input{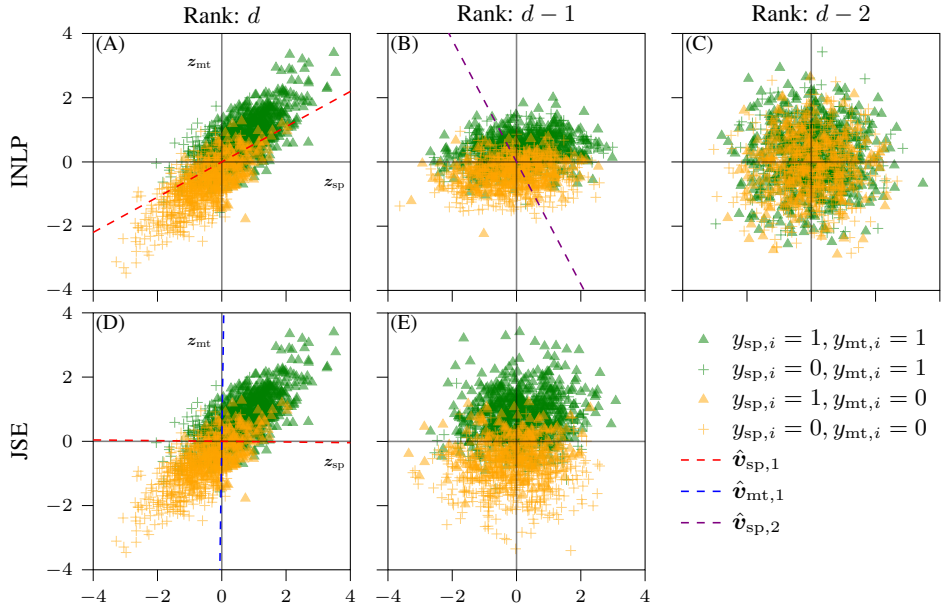}
\vspace{-0.25cm}
    \caption{\textbf{Illustration of JSE, in comparison to INLP}: based on the $d({=}20)$-dimensional Toy dataset (see Section \ref{sec:datasets}) with $\rho =0.8 $ and sample size $n =  $2,000. Two-dimensional slices of $\boldsymbol{z}$ are shown. Panels A and D have the spurious feature on the x-axis and the main-task feature on the y-axis. The remaining panels show the axes that best separate the main-task labels. JSE identifies a single spurious vector (panel D) and the remaining class separation is attributed to the main-task concept (panel E). INLP identifies (superpositions of) the main-task and spurious directions as spurious (panels A and B), and the main-task information is removed (panel C).}
    \label{fig:illustrate_toy} 
\end{figure*}

An alternative perspective is that $\Zsp$ and $\Zmt$ are not necessarily orthogonal, but that we aim to identify subspaces of $\Zsp$ and $\Zmt$ that are orthogonal to each other and informative about the respective labels. If $\Zsp$ and $\Zmt$ are high-dimensional (applicable in most realistic settings), there are enough degrees of freedom for these subspaces to cover significant parts of $\Zsp$ and $\Zmt$ in terms of their ability to predict $\ysp$ and $\ymt.$ We illustrate the effect of the orthogonality assumption in Figure~\ref{fig:illustrate_toy} (panels A and D) for Toy data, and analyse how JSE behaves when the assumption does not hold in  Appendix~\ref{sec:add_results_Toy}.

\setlength\figureheight{5cm}
\setlength\figurewidth{5cm}

We thus simultaneously perform a logistic regression on the embeddings $\bz$ for $\ysp$ and $\ymt,$ subject to the constraint of orthogonality of $\wsp$ and $\wmt.$ This means that for a sample $\left\{ \ymtsub{i}, \yspsub{i}, \bz_i \right\}_{i=1}^n$ the estimates $\wsp, \wmt, \bsp, \bmt$ are obtained by performing the following optimization:\begin{equation}\label{eq:estimation}
    \argmin_{\substack{ \wsp, \wmt, \bsp, \bmt \\ (\wsp \perp \wmt) }}
    \sum^{n}_{i = 1} 
    \mathcal{L}_{\mathrm{BCE}}(\yhatspsub{i}, \yspsub{i}) +   \mathcal{L}_{\mathrm{BCE}}(\yhatmtsub{i}, \ymtsub{i}),
\end{equation}
where $ \mathcal{L}_{\mathrm{BCE}}$ is the binary cross-entropy ($\mathrm{BCE}$). Furthermore, $\yhatspsub{i} = \mathrm{Logit}^{-1} \left(\bz_i^T \wsp + b_{\mathrm{sp}}\right),$ and similarly for $\yhatmtsub{i}.$ The estimated spurious and main-task concept vectors are then $\vhatsp = \whatsp / || \whatsp ||$ and $\vhatmt = \whatmt / || \whatmt ||.$

\subsection{Iteratively Estimating Multiple Concept and Main-task Vectors}
\label{sec:iterative}

The subspaces $\Zsp$ and $\Zmt$ will generally not be one-dimensional. Thus, the estimated spurious concept vector $\vhatsp$ could still contain main-task components (and vice versa for $\vhatmt$). To address this, we propose an iterative procedure to estimate the orthonormal bases of $\Zsp$ and $\Zmt,$ which are guaranteed to be orthogonal to each other. 

\begin{algorithm}[tb]
    \caption{JSE algorithm to estimate orthonormal bases for $\Zsp$ and $\Zmt.$ The conditions in the $\textbf{if}$-statements are discussed in Section~\ref{sec:testing}.}
    \label{alg:JSE_basic}
\begin{algorithmic}
    \STATE {\bfseries Input:} a sample $\left\{ \ymtsub{k}, \yspsub{k}, \bz_k \right\}_{k=1}^n$ consisting of two binary labels and a vector $\bz_k \in \mathbb{R}^d.$
    \STATE Initialize embedding matrix $\bZ = (\bz_1 \, \bz_2 \, \cdots \, \bz_n )^{\top}.$
    \STATE Initialize $\Zsportho \leftarrow \bZ.$
    \FOR{$i = 1, ..., d$}
        \STATE $\Zremain \leftarrow \Zsportho$
        \FOR{$j =1, ..., d$}
            \STATE Estimate $\whatsp,$ $\whatmt$ with \Eqref{eq:estimation} (use $\Zremain$).
            \STATE $\vhatspsub{i} \leftarrow \whatsp / || \whatsp ||$ and $\vhatmtsub{j} \leftarrow \whatmt / || \whatmt ||.$
            \IF{$\vhatmtsub{j}$ is a proper main-task direction}
                \STATE Projection $\Zremain \leftarrow \Zsportho (\boldsymbol{I} - \Vhatmt\Vhatmt^{\top}),$ with $\Vhatmt = (\vhatmtsub{1} \, \vhatmtsub{2} \, \cdots \, \vhatmtsub{j} ).$
            \ELSE
                \STATE \textbf{break}
            \ENDIF
        \ENDFOR
        \IF{$\vhatspsub{i}$ is a proper spurious direction}
            \STATE Projection $\Zsportho \leftarrow \bZ (\boldsymbol{I} - \Vhatsp\Vhatsp^{\top}),$ where \\ $\Vhatsp = (\vhatspsub{1} \, \vhatspsub{2} \, \cdots \, \vhatspsub{i} ).$
            \STATE $\hat{\boldsymbol{v}}'_{\mathrm{mt},{\ell}} \leftarrow \vhatmtsub{\ell},$ for $\ell=1,2,\ldots,\ell_{0},$ with $\ell_{0} = j.$
        \ELSE
            \STATE \textbf{break}
        \ENDIF
    \ENDFOR
\STATE \textbf{return} Bases $\{\vhatspsub{m}\}_{m=1}^{i-1}$ and $\{\hat{\boldsymbol{v}}'_{\mathrm{mt},{\ell}}\}_{\ell=1}^{\ell_{0}-1}.$
\end{algorithmic}
\end{algorithm}
For now, let us focus on estimating a vector $\vsp \in \Zsp.$  By applying the procedure of \Eqref{eq:estimation} gives a $\vhatsp$ and $\vhatmt,$ where $\vhatsp$ may still have components in $\Zmt$ and $\vhatmt$ may have components in $\Zsp.$ We propose to project out the direction $\vhatmt$ from the embeddings and to repeat the optimization of \Eqref{eq:estimation} for the resulting subspace. Doing this multiple times will eventually remove all main-task information, guaranteeing that the estimated vector $\vhatspsub{1}$ is orthogonal to the (estimated) main-task subspace.

By projecting out $\vhatspsub{1}$ from the embeddings $\boldsymbol{z}$ and repeating the whole procedure $\dsp$ times, we estimate an orthonormal basis $\vhatspsub{1}, \vhatspsub{2}, \ldots, \vhatspsub{\dsp}$ of $\Zsp$ that is orthogonal to the (main-task) subspace. The method described is a nested for-loop (see Algorithm~\ref{alg:JSE_basic}), where the inner loop finds and projects out main-task vectors, and the outer loop finds and projects out spurious vectors. After having found a basis of $\Zsp,$ one can repeat the inner loop one last time to find $\dmt = \dim (\Zmt)$ vectors $\vhatmtsub{1}, \vhatmtsub{2}, \ldots, \vhatmtsub{\dmt}$ constituting an estimated basis for $\Zmt.$ The computational cost of the double for-loop is limited, as the number of light-weight optimizations (\mbox{\Eqref{eq:estimation}}) is at most $\dmt \times \dsp.$  For a  detailed description of the algorithm, see Appendix~\ref{sec:details_algorithm}.

So far, we have treated the subspaces $\Zsp$ and $\Zmt$ equally. This symmetry is broken in Algorithm~\ref{alg:JSE_basic}, as the main-task directions are identified in the inner loop and the concept directions in the outer loop. In Appendix~\ref{sec:details_algorithm} we give empirical evidence that swapping the loops has little effect on the outcome of the JSE method.

One can remove the spurious concept by projecting the embeddings $\boldsymbol{z}$ on the orthogonal complement of the spurious concept subspace, $\Zsp^\perp$. The transformed embeddings can then be used for main-task prediction by a linear classifier.  In Appendix~\ref{sec:details_algorithm}, we show that training the linear classifier on $\Zmt$ instead of $\Zsp^\perp$ gives similar performance. 

\subsection{Testing when to Stop Adding Vectors} \label{sec:testing}
So far, in the description of the iterative algorithm we have assumed the dimensions $\dsp$ and $\dmt$ of the respective subspaces $\Zsp$ and $\Zmt$  to be known. In practice, the dimensions must be estimated via stopping criteria of the (nested) \textbf{for} loops in Algorithm~\ref{alg:JSE_basic}. Let us focus on the condition in the outer loop. The condition in the inner loop is, {\it mutatis mutandis}, the same. For a given (normalized) direction $\vsp \in \mathbb{R}^d$ in the embedding space, the statement ``$\vsp$ is a proper spurious direction'' means that two criteria are met:
\begin{enumerate}
    \item \textbf{The direction $\vsp$ is informative about the spurious label $\ysp$}, meaning that the embeddings projected onto $\vsp$ are able to predict the spurious label. To be concrete, a logistic regression based on the projected embeddings should have a higher accuracy than a classifier that just predicts the majority class, which we refer to as a `random classifier'. 
    \item \textbf{The direction $\vsp$ should be more predictive of the spurious concept than of the main-task concept.} Due to the spurious correlation, a vector in $\Zsp$ is likely also predictive for the main-task concept. We nonetheless associate it with the spurious subspace $\Zsp,$ as long as its prediction accuracy for the spurious label is higher than for the main-task label.
\end{enumerate}
Note that the first criterion is already used in \citet{ravfogel_2020}, while the second is novel and addresses the problem of inadvertently removing main-task information in existing concept-removal methods. This is illustrated in Figure~\ref{fig:illustrate_toy} (panel B), where the INLP-method of \citet{ravfogel_2020} removes a feature that is more predictive of the main-task concept than the spurious concept.

To make these criteria operational, we introduce two statistical tests in terms of differences between BCE's. For the first criterion we compare the BCE of $\yhatsp^{(\vsp)} = \mathrm{Logit}^{-1}\left(\gamma_\mathrm{sp} \bz^{\top} \vsp + b_\mathrm{sp}\right),$ which is a  predictor for the label $\ysp$ based on the embeddings projected onto $\vsp,$ and the BCE of a majority-rule `random classifier'. The model parameters $\gamma_\mathrm{sp}$ and $b_\mathrm{sp}$ will be trained by minimizing the BCE. For the second criterion we compare the BCE of $\yhatsp^{(\vsp)}$ with the analogously defined $\yhatmt^{(\vsp)},$ which is a predictor of $\ymt.$ Both tests are performed using a $t$-statistic, using a  weighted average of the BCE's over the four combinations of $\ysp$ and $\ymt$. For a precise definition of the hypotheses, test statistics, and their properties, see Appendix \ref{sec:details_tests}.

\section{Experiments} \label{sec:experiments}
This section gives experimental evidence that JSE is able to identify the main-task and spurious concepts, and that in these respects JSE outperforms existing concept-removal methods. Since for realistic datasets there exists no ground truth about how main-task and spurious concepts are represented in the last-layer embeddings, we mostly need to rely on proxy experiments. A first experiment is the problem of OOD generalization (see Section \ref{sec:OOD_generalization}), where models are trained on data with a spurious correlation, and tested on data without. A good performance of a model after concept removal indicates that the spurious concept was removed adequately, while main-task features have remained intact. 

We proceed with an experiment on Toy data in Section \ref{sec:Toy_data}, for which we know the true main-task and spurious features. To further substantiate our claims, we present Grad-CAM results in Section \ref{sec:gradcam_main} and conduct experiments similar to \citet{ravfogel_2020} and \citet{Kumar_2022} to test the validity of concept-removal methods, in Sections \ref{sec:Rafvogel_experiment_main} and \ref{sec:Kumar_experiment} respectively. In this results section, we compare JSE with other last-layer concept-removal methods mentioned in Section\mbox{~\ref{sec:related_work}}: iterative null-space projection (INLP), relaxed linear adversarial concept erasure (RLACE), adversarial removal based on a single linear adversary (ADV) and least-squares concept erasure (LEACE). Details about the datasets, experimental setup and parameter selection are in Appendix\mbox{~\ref{sec:param_details}}. For numerical details, see Appendix\mbox{~\ref{sec:full_results}}.

\subsection{Datasets}
\label{sec:datasets}

JSE and existing concept-removal methods are applied to a Toy dataset, for which knowledge about the true main-task and spurious features is available, as well as benchmark datasets from computer vision (Waterbirds, CelebA) and natural language processing (MultiNLI). A brief description of the datasets and how we use them is given here. See Appendix \ref{app:datasets} for detailed information.

\textbf{Toy data: } We create $d$-dimensional embeddings drawn from a multivariate normal distribution with a block correlation matrix, $\boldsymbol{z} \sim \mathcal{N}(\boldsymbol{\mu} = \boldsymbol{0}, \boldsymbol{\Sigma}),$ where
\begin{equation*}
   \boldsymbol{\Sigma} = \begin{bmatrix}
        \boldsymbol{\Sigma}_{\mathrm{sp}, \mathrm{mt}} &  \boldsymbol{0} \\
        \boldsymbol{0} & \boldsymbol{I}
    \end{bmatrix}, \quad 
     \boldsymbol{\Sigma}_{\mathrm{sp}, \mathrm{mt}} = \begin{bmatrix}
         1 & \rho \\
         \rho & 1
     \end{bmatrix}.
\end{equation*}
Note that the embeddings here are not neural network representations of underlying input features. We set $\Zsp$ and $\Zmt$ to be one-dimensional subspaces, with direction vectors $\boldsymbol{w}_{\mathrm{sp}} = ( \gamma_{\mathrm{sp}}, 0,  0, \dots, 0 )^{\top}$ and $\boldsymbol{w}_{\mathrm{mt}} = ( 0, \gamma_{\mathrm{mt}},  0, \dots, 0 )^{\top}$, respectively. We define binary labels $\ysp$ and $\ymt$,
\begin{equation*}
    p(\ysp = 1 | \bz) = \mathrm{Logit}^{-1}\left(\boldsymbol{z}^{\top}\boldsymbol{w}_{\mathrm{sp}} + b_{\mathrm{sp}}\right), 
\end{equation*}
and likewise for $p(\ymt = 1 | \bz).$
Throughout the simulations we take $d = 20$, $b_{\mathrm{sp}} = b_{\mathrm{mt}} = 0$ and $\gamma_{\mathrm{sp}} = \gamma_{\mathrm{mt}}= 3.$ Parameter $\rho$ is the correlation between the spurious and main-task features,  determining the spurious relation $p_{\mathrm{train}}(y_{\mathrm{mt}} | \boldsymbol{z}_{\mathrm{sp}})$ between the main-task label and the spurious feature. 

\textbf{Vision:} We use two common datasets containing a spurious correlation. The first is the Waterbirds dataset~\citep{sagawa_2020}, where the main-task concept is bird type (waterbird vs.~landbird) and the spurious concept is background (water vs.~land). The second is the CelebA dataset, where the  main-task concept is hair color (blond vs.~non-blond) and the spurious concept is sex (female vs.~male). We use a pre-trained ResNet50 architecture  \citep{He_2015}, which is then finetuned on the respective dataset, after which the concept removal is applied to the last layer. 

\textbf{NLP:} We use the MultiNLI dataset \citep{Bowman_MultiNLI}, which contains pairs of sentences. The main-task concept is, whether or not the first sentence contradicts the second. Following an experiment from \citet{joshi2022spurious}, we use as spurious concept the presence or absence of punctuation marks (`!!') at the end of the second sentence. For exemplary pairs of sentences, see Appendix~\ref{app:datasets}. Each run in our experiments starts with finetuning a BERT model, after which concept removal is applied to the [CLS] embeddings. 

For the vision and NLP datasets, we cannot directly control the strength of the spurious relation between the main-task label and the spurious features, $p_{\mathrm{train}}(\ymt | \zsp).$ We therefore use $p_{\mathrm{train}}(\ymt = y | \ysp = y)$, with $y \in \{0, 1\},$ as a proxy. 
To increase the precision of our method and for computational efficiency, we reduce the dimension of the last-layer embeddings to $d=300$ for Waterbirds and CelebA, and $d=100$ for multiNLI via Principal Component Analysis (PCA).

\subsection{Out-of-distribution (OOD) Generalization} \label{sec:OOD_generalization}

We consider the problem of OOD generalization, where the training and OOD test data have a different dependence between main-task and spurious features (i.e.~$p_{\mathrm{train}}(\xmt , \xsp) \neq p_{\mathrm{OOD}}(\xmt , \xsp)) $. This generally leads to $p_{\mathrm{train}}(\ymt | \xsp) \neq p_{\mathrm{OOD}}(\ymt | \xsp),$ with similar discrepancies at the level of the embeddings. A DNN using the spurious features to predict the main-task label will see performance loss when applied to the OOD data. Similarly, if a concept-removal method has inadvertently also removed main-task features, the resulting DNN will also observe a deterioration in performance.

\begin{figure*}
\centering
\input{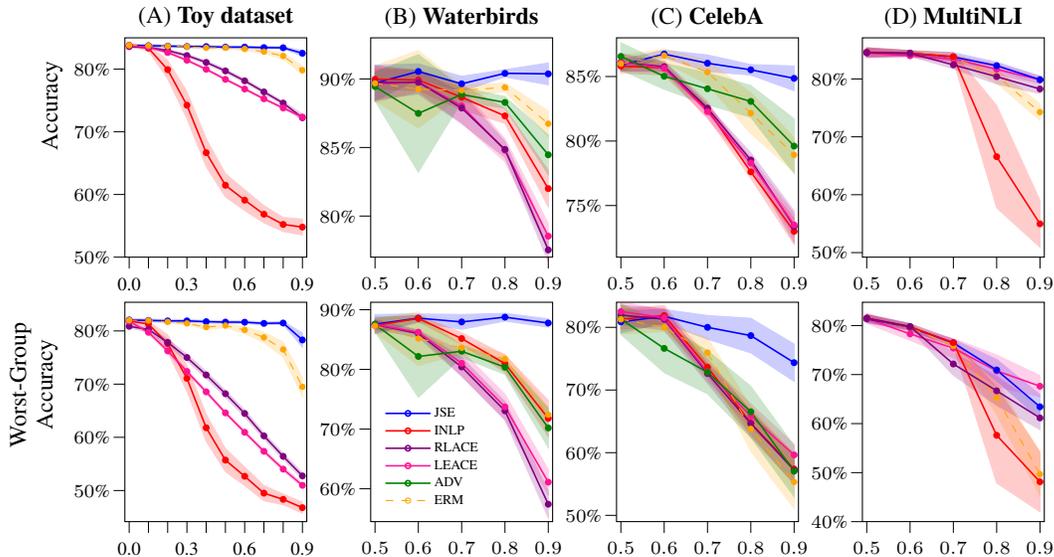}
\vspace{-0.3cm}
\caption{\textbf{OOD generalization, compared to other concept-removal methods}: We plot the (worst-group) accuracy on a test set without spurious correlation, as a function of the spurious correlation in the training set ($\rho$ for the Toy dataset, $p_\mathrm{train}(\ymt = y  | \ysp = y )$ for the other datasets). Averages based on 100, 5, 5 and 5 runs, respectively. The shaded area reflects the 95\% confidence interval.}
\label{fig:OOD_generalization_for_concept_removal}
\vspace{-0.25cm}
\end{figure*}

Figure~\ref{fig:OOD_generalization_for_concept_removal} shows the results of applying different concept-removal methods to the problem of OOD generalization. JSE outperforms the other methods for the Toy and vision datasets, especially when the spurious correlation is strong. When applied to the text dataset, it performs similar to the best other concept-removal methods (LEACE and RLACE). We suspect that during finetuning of BERT the main-task and spurious concepts become overlapping in the [CLS] embeddings, in line with previous work by \citet{dalvi2022discovering}. 

The performance loss of INLP, RLACE and LEACE is explained by the fact that they also inadvertently remove main-task features. This behaviour is illustrated for the Toy dataset in Figure\mbox{~\ref{fig:illustrate_toy}}. For ADV the performance deteriorates because the spurious features remain present after the training procedure, in line with previous work \mbox{\citep{Belinkov_2021, rafvogel_2022}}. JSE also shows performance loss for the Toy and vision datasets, albeit to a lesser extent. As the spurious correlation increases, we suspect our method becomes more sensitive to finite-sample estimation noise. We illustrate this further in Appendix\mbox{~\ref{sec:finite_sample_noise}}.

As an additional benefit, JSE is the first concept-removal method that is (for all datasets) competitive with several instance-reweighting techniques such as group-weighted ERM, group-distributional robust optimization (GDRO), and just train twice (JTT). For an overview of these techniques and a comparison with JSE, see Appendix \ref{sec:OOD_other_algorithms}. We emphasise that this class of techniques is purely aimed at OOD generalization, while the primary focus of JSE is to improve the interpretability of the model.


\subsection{Regression Analysis for Toy Data} \label{sec:Toy_data}

\begin{figure}[H]
    \centering
    \input{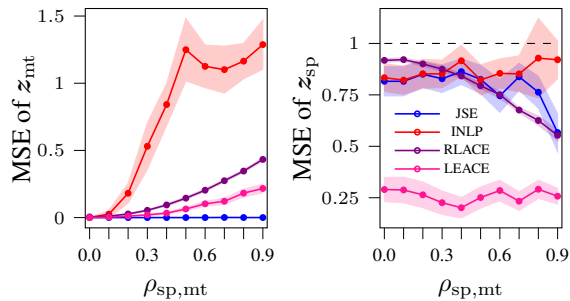}
    \vspace{-0.3cm}
    \caption{\textbf{Ability to reconstruct main-task and spurious concept features after concept removal.}  We show the mean-squared error (MSE) of predicting $\zmt, \zsp$ via OLS on the transformed embeddings. The dotted line on the right plot indicates the MSE when there is no information of $\zsp$ left. Averages based on 100 runs, and shaded area reflects the 95\% confidence interval.} 
    \label{fig:MSE_plot}
\end{figure}

Because we have access to the ground truth for the Toy dataset, we can quantify whether spurious features are removed and main-task features are preserved. Let $\Tilde{\boldsymbol{z}}$ denote the embedding vectors after being transformed by a concept-removal method. We aim to find linear models that use $\Tilde{\boldsymbol{z}}$ to predict $\zmt$ or $\zsp.$ Estimation is conducted via ordinary least squares (OLS) on a test set without correlation between $\zmt$ and $\zsp,$ to prevent the regression from predicting one with the other. A good concept-removal method creates embeddings $\Tilde{\boldsymbol{z}}$ that can reconstruct $\zmt$ (low MSE) and cannot reconstruct $\zsp$ (high MSE).
Figure \ref{fig:MSE_plot} shows that JSE preserves the main-task feature, while other methods remove (part of) it. The spurious feature is removed to a similar extent by JSE, INLP and RLACE. Interestingly, while LEACE prevents the prediction of $\ysp$ with $\Tilde{\boldsymbol{z}},$ it appears that information related to $\zsp$ remains present.

\subsection{Grad-CAM for Waterbirds} 
\label{sec:gradcam_main}

To obtain insight into the workings of the concept-removal methods on the Waterbirds dataset, we use Grad-CAM \citep{Selvaraju_2016} to highlight parts of the images used by the models (after concept removal). Figure \ref{fig:WB_gradcam} shows the results (see Appendix \ref{sec:gradcam_fig} for more images and concept-removal methods). JSE produces a model that relies predominantly on the bird features and neglects the background.   ERM and INLP use the background for their correct and incorrect  predictions. Interestingly, INLP, LEACE and RLACE perform much worse on images that appear more frequently in the training set (e.g.~landbirds on land) than on images from minority groups (e.g.~landbirds on water). We observe a similar pattern for the other datasets. We posit that this is because features get mixed, as described by \citet{Kumar_2022}, leading INLP, LEACE and RLACE to associate spurious features with the main-task concept. This is visible in Figure~\ref{fig:WB_gradcam}, where after INLP the model classifies a waterbird as a landbird by using the water background.

\begin{figure}[h]
    \centering
            \vspace{-0.24cm}
    \includegraphics[scale=0.38]{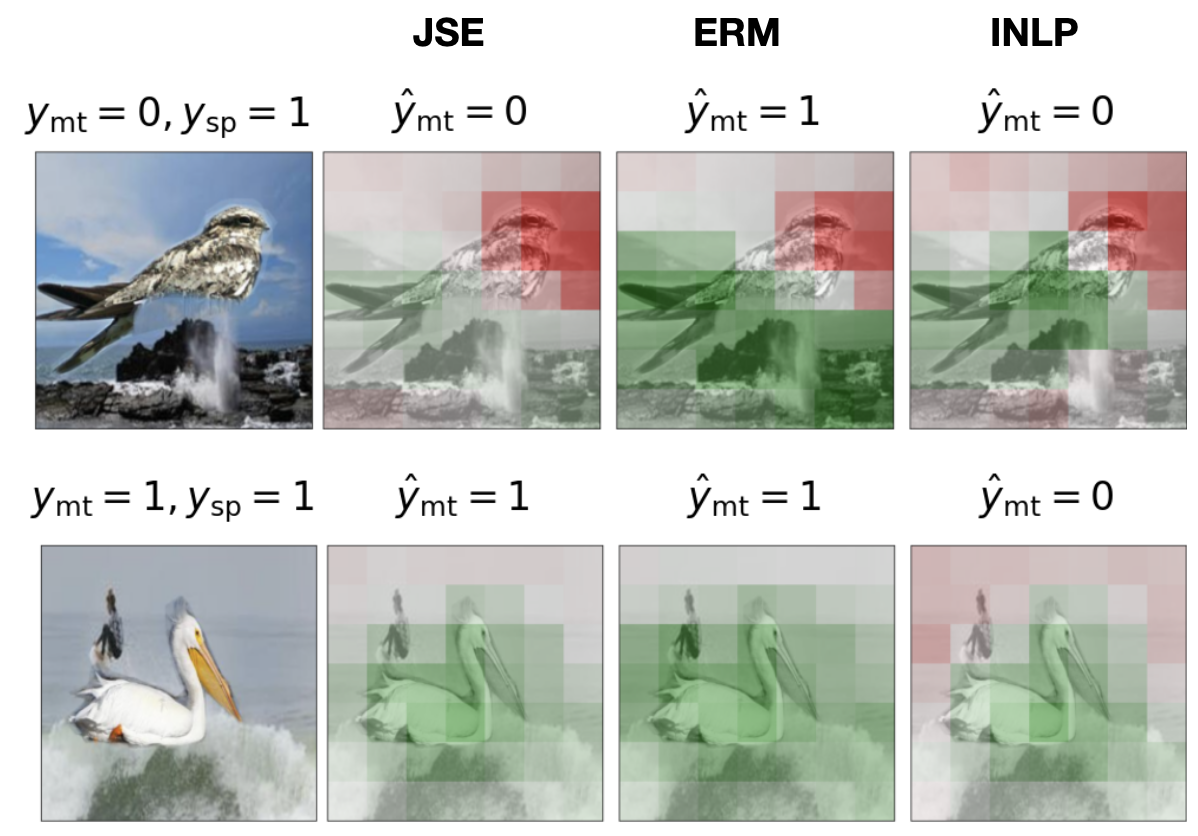}
        \vspace{-0.25cm}
    \caption{\textbf{Grad-CAM for the last layer of Resnet-50 predicting the main-task label}: Red (green) patches indicate a contribution towards a prediction $\ymt=0$ ($\ymt=1$). }
    \label{fig:WB_gradcam}
\end{figure}

\subsection{Removing Concepts from Original Images of CelebA} \label{sec:Rafvogel_experiment_main}

We conduct an experiment for the CelebA dataset that is similar to one from \citet{rafvogel_2022}. The goal is to qualitatively show what features are removed by a concept-removal method. Instead of working with the embeddings, we downscale the images to 50 by 50 grey-scale images, flatten them to a 2,500-dimensional vector, and apply concept-removal methods to the raw pixels. The main-task concept is the presence of 'glasses', while the spurious concept is 'smiling'. We set the spurious correlation at $p(\ymt = y | \ysp = y) = 0.8.$  The results are shown in Figure \ref{fig:CelebA_removal}, with more images and methods in Appendix \ref{sec:CelebA_removal}. Other concept-removal methods  change pixels that relate to both the smile and the glasses. JSE is the only method that primarily focuses on the smile.

\subsection{Preserving Main-task Features in Text Data } \label{sec:Kumar_experiment}

We use the MultiNLI dataset to perform an experiment similar to one from \citet{Kumar_2022}. First, we train a BERT model on a subset of the data with only one value of the spurious concept label (we choose $\ysp = 0,$ meaning there is no `!!' at the end of the second sentence). We refer to this as a `clean' BERT model. According to \citet{ravichander-etal-2021-probing}, a model trained in this manner should not use spurious features. As a consequence, applying a concept-removal method to a clean BERT model should not affect the main-task classification. We test this by training concept-removal methods on data with both spurious labels present and measure how well they generalize OOD.

\begin{figure}[H]
    \centering
    \vspace{-0.1cm}
    \includegraphics[scale=0.27]{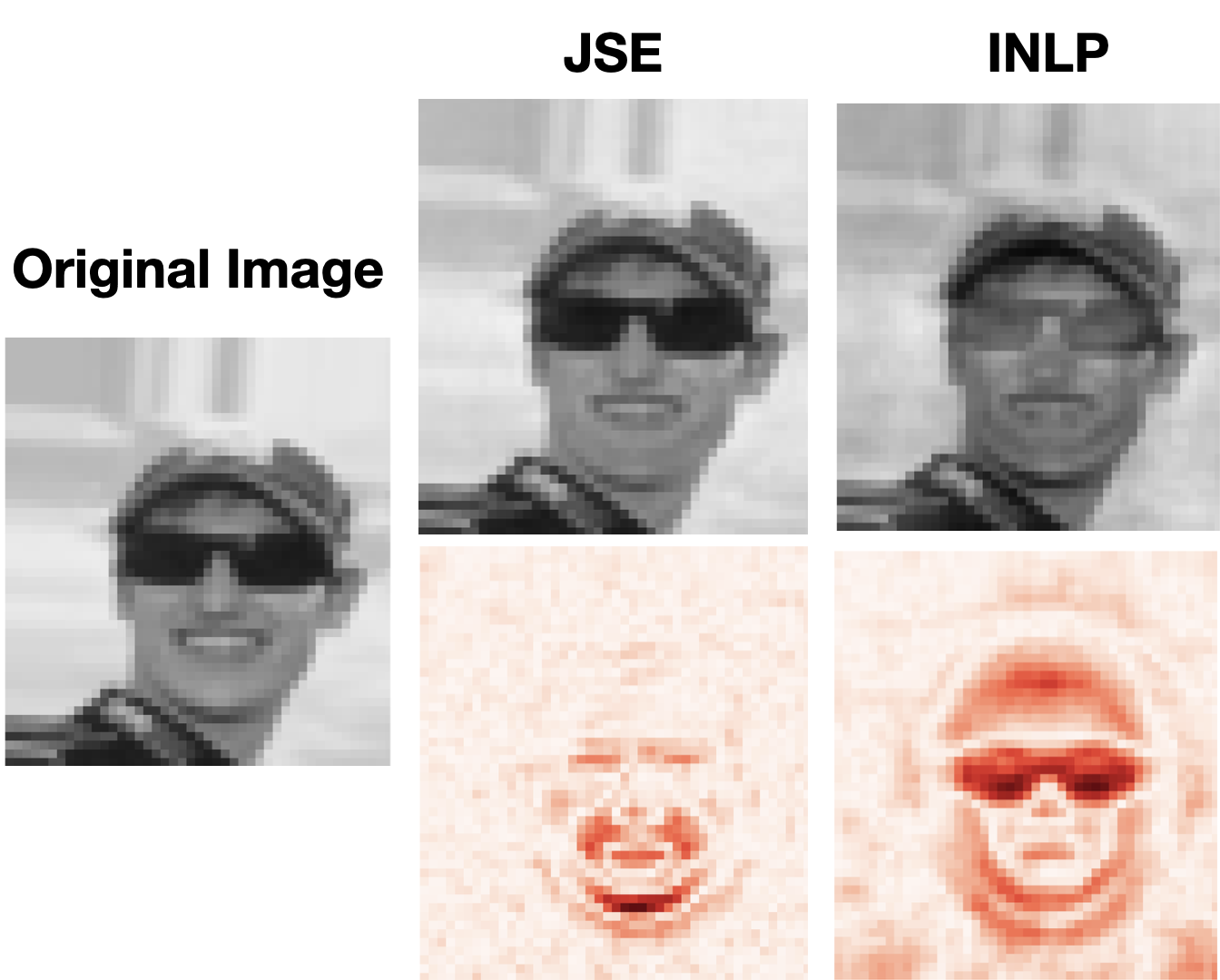}
    \vspace{-0.2cm}
    \caption{\textbf{Application of concept-removal methods to raw pixel data}:  The first row shows the image after it is transformed by the concept-removal method. The second row shows the absolute difference between the transformed and original image to indicate which pixels have been changed.}
    \label{fig:CelebA_removal}
\end{figure}

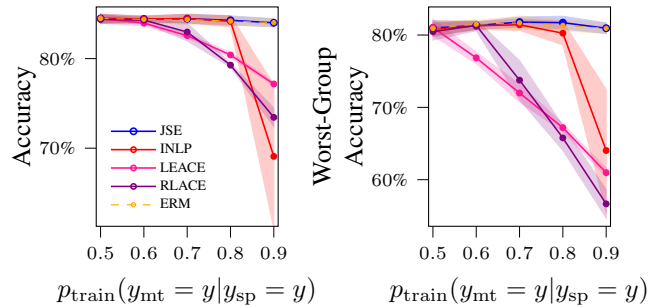
\begin{figure}[H]
    \centering
\begin{tikzpicture}

\definecolor{darkgray176}{RGB}{176,176,176}
\definecolor{deeppink}{RGB}{255,20,147}
\definecolor{lightgray204}{RGB}{204,204,204}
\definecolor{purple}{RGB}{128,0,128}
\definecolor{orange}{RGB}{255,165,0}

\setlength\figureheight{4.5cm}
\setlength\figurewidth{4.0cm}

\begin{groupplot}[group style={group size=2 by 1, horizontal sep=2cm}]
\nextgroupplot[
height=\figureheight,
legend cell align={left},
legend style={
  fill opacity=0.0,
  draw opacity=1,
  text opacity=1,
  at={(0.03,0.03)},
  nodes={scale=0.6},
  anchor=south west,
  draw=none
},
tick align=outside,
tick pos=left,
width=\figurewidth,
x grid style={darkgray176},
xlabel={\(\displaystyle p_{\mathrm{train}}(y_{\mathrm{mt}}=y | y_{\mathrm{sp}} = y)\) },
xmin=0.49, xmax=0.91,
xtick style={color=black},
y grid style={darkgray176},
ylabel style ={xshift=0cm, yshift=-0.2cm}, 
ylabel={Accuracy},
ymin=61.279567867595, ymax=85.7689129432894,
ytick style={color=black},
ytick={60, 70,  80},
yticklabel={\pgfmathprintnumber\tick\%}
]
\path [draw=blue, fill=blue, opacity=0.2]
(axis cs:0.5,84.9599405344414)
--(axis cs:0.5,84.0000609342171)
--(axis cs:0.6,84.1413919485781)
--(axis cs:0.7,83.9476983057375)
--(axis cs:0.8,83.7842940679668)
--(axis cs:0.9,83.5401715339457)
--(axis cs:0.9,84.4598279892171)
--(axis cs:0.9,84.4598279892171)
--(axis cs:0.8,84.7677064546467)
--(axis cs:0.7,84.9642999662047)
--(axis cs:0.6,84.7226078950193)
--(axis cs:0.5,84.9599405344414)
--cycle;

\path [draw=red, fill=red, opacity=0.2]
(axis cs:0.5,84.914789367175)
--(axis cs:0.5,83.9652107472659)
--(axis cs:0.6,84.0698293601754)
--(axis cs:0.7,84.0153130587189)
--(axis cs:0.8,83.5306549937716)
--(axis cs:0.9,61.0801684193108)
--(axis cs:0.9,77.0878351398017)
--(axis cs:0.9,77.0878351398017)
--(axis cs:0.8,84.7973441212186)
--(axis cs:0.7,84.9206868115813)
--(axis cs:0.6,84.794170483422)
--(axis cs:0.5,84.914789367175)
--cycle;

\path [draw=deeppink, fill=deeppink, opacity=0.2]
(axis cs:0.5,84.9758448529851)
--(axis cs:0.5,83.9841542314875)
--(axis cs:0.6,83.639231642684)
--(axis cs:0.7,82.3362214043748)
--(axis cs:0.8,80.119247714722)
--(axis cs:0.9,76.9758398135483)
--(axis cs:0.9,77.3521607319534)
--(axis cs:0.9,77.3521607319534)
--(axis cs:0.8,80.696751793182)
--(axis cs:0.7,82.7997773215163)
--(axis cs:0.6,84.2887702379618)
--(axis cs:0.5,84.9758448529851)
--cycle;

\path [draw=purple, fill=purple, opacity=0.2]
(axis cs:0.5,84.8183525692085)
--(axis cs:0.5,83.9336477626701)
--(axis cs:0.6,83.9535736726012)
--(axis cs:0.7,81.9806124789068)
--(axis cs:0.8,79.0389013062419)
--(axis cs:0.9,72.4278739646352)
--(axis cs:0.9,74.4621225639903)
--(axis cs:0.9,74.4621225639903)
--(axis cs:0.8,79.5051002730428)
--(axis cs:0.7,83.9553878682307)
--(axis cs:0.6,84.6464276625429)
--(axis cs:0.5,84.8183525692085)
--cycle;

\path [draw=orange, fill=orange, opacity=0.2]
(axis cs:0.5,84.9659867628785)
--(axis cs:0.5,83.9940147057799)
--(axis cs:0.6,84.0683672857608)
--(axis cs:0.7,83.7939367154561)
--(axis cs:0.8,83.70797949712)
--(axis cs:0.9,83.5776618217797)
--(axis cs:0.9,84.4543382430702)
--(axis cs:0.9,84.4543382430702)
--(axis cs:0.8,84.6760193737296)
--(axis cs:0.7,84.9420609613933)
--(axis cs:0.6,84.6916319894467)
--(axis cs:0.5,84.9659867628785)
--cycle;

\addplot [semithick, blue, mark=*, mark size=1.2, mark options={solid}]
table {%
0.5 84.4800007343292
0.6 84.4319999217987
0.7 84.4559991359711
0.8 84.2760002613068
0.9 83.9999997615814
};
\addlegendentry{JSE}
\addplot [semithick, red, mark=*, mark size=1, mark options={solid}]
table {%
0.5 84.4400000572205
0.6 84.4319999217987
0.7 84.4679999351501
0.8 84.1639995574951
0.9 69.0840017795563
};
\addlegendentry{INLP}
\addplot [semithick, deeppink, mark=*, mark size=1, mark options={solid}]
table {%
0.5 84.4799995422363
0.6 83.9640009403229
0.7 82.5679993629456
0.8 80.407999753952
0.9 77.1640002727509
};
\addlegendentry{LEACE}

\addplot [semithick, purple, mark=*, mark size=1, mark options={solid}]
table {%
0.5 84.3760001659393
0.6 84.300000667572
0.7 82.9680001735687
0.8 79.2720007896423
0.9 73.4449982643127
};
\addlegendentry{RLACE}

\addplot [dashed, orange, mark=*, mark size=0.8, mark options={solid}]
table {%
0.5 84.4800007343292
0.6 84.3799996376038
0.7 84.3679988384247
0.8 84.1919994354248
0.9 84.0160000324249
};
\addlegendentry{ERM}

\nextgroupplot[
height=\figureheight,
tick align=outside,
tick pos=left,
width=\figurewidth,
x grid style={darkgray176},
xlabel={\(\displaystyle p_{\mathrm{train}}(y_{\mathrm{mt}}=y | y_{\mathrm{sp}} = y)\) },
xmin=0.49, xmax=0.91,
xtick style={color=black},
y grid style={darkgray176},
ylabel style ={xshift=0cm, yshift=-0.2cm, align=center}, 
ylabel={Worst-Group \\ Accuracy},
ymin=53.5547731399536, ymax=83.8848821640015,
ytick={60, 70,  80},
yticklabel={\pgfmathprintnumber\tick\%},
ytick style={color=black}
]
\path [draw=blue, fill=blue, opacity=0.2]
(axis cs:0.5,82.001091003418)
--(axis cs:0.5,79.9509048461914)
--(axis cs:0.6,80.795280456543)
--(axis cs:0.7,81.1020812988281)
--(axis cs:0.8,80.8241729736328)
--(axis cs:0.9,80.2128982543945)
--(axis cs:0.9,81.6751022338867)
--(axis cs:0.9,81.6751022338867)
--(axis cs:0.8,82.5998229980469)
--(axis cs:0.7,82.4819183349609)
--(axis cs:0.6,81.828727722168)
--(axis cs:0.5,82.001091003418)
--cycle;

\path [draw=red, fill=red, opacity=0.2]
(axis cs:0.5,81.7879638671875)
--(axis cs:0.5,79.8120269775391)
--(axis cs:0.6,80.7391052246094)
--(axis cs:0.7,80.6567993164062)
--(axis cs:0.8,78.6224212646484)
--(axis cs:0.9,55.5942344665527)
--(axis cs:0.9,72.4697647094727)
--(axis cs:0.9,72.4697647094727)
--(axis cs:0.8,81.8895874023438)
--(axis cs:0.7,82.1911926269531)
--(axis cs:0.6,81.8209075927734)
--(axis cs:0.5,81.7879638671875)
--cycle;

\path [draw=deeppink, fill=deeppink, opacity=0.2]
(axis cs:0.5,82.1815185546875)
--(axis cs:0.5,79.8984832763672)
--(axis cs:0.6,76.0693511962891)
--(axis cs:0.7,70.7262725830078)
--(axis cs:0.8,66.6682739257812)
--(axis cs:0.9,60.4054527282715)
--(axis cs:0.9,61.5465431213379)
--(axis cs:0.9,61.5465431213379)
--(axis cs:0.8,67.7317352294922)
--(axis cs:0.7,73.2417297363281)
--(axis cs:0.6,77.6266479492188)
--(axis cs:0.5,82.1815185546875)
--cycle;

\path [draw=purple, fill=purple, opacity=0.2]
(axis cs:0.5,81.623046875)
--(axis cs:0.5,79.3049468994141)
--(axis cs:0.6,80.5965576171875)
--(axis cs:0.7,71.0409698486328)
--(axis cs:0.8,64.051628112793)
--(axis cs:0.9,54.7037620544434)
--(axis cs:0.9,58.6162300109863)
--(axis cs:0.9,58.6162300109863)
--(axis cs:0.8,67.500373840332)
--(axis cs:0.7,76.4790191650391)
--(axis cs:0.6,82.0594329833984)
--(axis cs:0.5,81.623046875)
--cycle;

\path [draw=orange, fill=orange, opacity=0.2]
(axis cs:0.5,81.9264907836914)
--(axis cs:0.5,79.8975143432617)
--(axis cs:0.6,80.9925308227539)
--(axis cs:0.7,80.7638626098633)
--(axis cs:0.8,80.5946273803711)
--(axis cs:0.9,80.2050170898438)
--(axis cs:0.9,81.6509857177734)
--(axis cs:0.9,81.6509857177734)
--(axis cs:0.8,82.0933609008789)
--(axis cs:0.7,82.6281356811523)
--(axis cs:0.6,82.0154647827148)
--(axis cs:0.5,81.9264907836914)
--cycle;

\addplot [semithick, blue, mark=*, mark size=1.2, mark options={solid}]
table {%
0.5 80.9759979248047
0.6 81.3120040893555
0.7 81.7919998168945
0.8 81.7119979858398
0.9 80.9440002441406
};
\addplot [semithick, red, mark=*, mark size=1, mark options={solid}]
table {%
0.5 80.7999954223633
0.6 81.2800064086914
0.7 81.4239959716797
0.8 80.2560043334961
0.9 64.0319976806641
};
\addplot [semithick, deeppink, mark=*, mark size=1, mark options={solid}]
table {%
0.5 81.0400009155273
0.6 76.8479995727539
0.7 71.984001159668
0.8 67.2000045776367
0.9 60.9759979248047
};

\addplot [semithick, purple, mark=*, mark size=1, mark options={solid}]
table {%
0.5 80.463996887207
0.6 81.327995300293
0.7 73.7599945068359
0.8 65.7760009765625
0.9 56.6599960327148
};

\addplot [dashed, orange, mark=*, mark size=0.8, mark options={solid}]
table {%
0.5 80.9120025634766
0.6 81.5039978027344
0.7 81.6959991455078
0.8 81.343994140625
0.9 80.9280014038086
};
\end{groupplot}

\end{tikzpicture}
        \vspace{-0.65cm}
    \caption{\textbf{Result of applying concept removal methods to a `clean' BERT model}: We plot the (worst-group) accuracy on a test set without spurious correlation, as a function of the spurious correlation in the dataset that was used for the concept removal method. The BERT model is finetuned on a dataset without variation in the spurious concept ($\ysp = 0$). Averages are based on 5 runs, and the shaded area reflects the 95\% confidence interval.}
    \label{fig:kumar_plot}
\end{figure}

The results of this experiment in Figure \ref{fig:kumar_plot} show that only JSE is applied without performance loss, with respect to the clean BERT model (ERM). This indicates that it has been able to remove features related to the spurious concept label, without removing main-task features. Similar to \citet{Kumar_2022}, for other concept-removal methods, the main-task features are removed to a greater extent as the spurious correlation increases, affecting main-task performance.

\section{Conclusion and Discussion}
This paper has introduced and empirically tested joint subspace estimation (JSE), a novel post-hoc concept-removal method that improves the interpretability and control of neural network representations in the presence of spurious correlations. JSE outperforms existing concept-removal methods, despite making assumptions (linearity, orthogonality) about the structure of the embedding space.

Future work will be needed to develop tests for these assumptions, or see if they can be relaxed. One example is jointly estimating subspaces based on their non-linear relationship with the spurious and main-task labels, as done by \citet{rafvogel_2022_kernel}. Our results highlight the difficulty of separating different concepts in BERT's [CLS] embeddings. This underlines the need to better disentangle concepts in  embeddings of large language models, for instance through different training procedures, as done by \citet{zhang-etal-2021-disentangling}.

\section*{Impact Statement}

Our work has the potential to give individuals greater control over which features are used by DNNs. This potentially has societal consequences: ensuring that a DNN uses the correct or desirable features is crucial before we can apply it in high-stake domains such as medicine or finance \citep{rudin_2019}. However, as highlighted in the paper, our work is not without limitations. Care should be taken to measure the effectiveness of the approach in removing spurious concepts in the context in which it is to be deployed. 

\section*{Acknowledgements}
We are very grateful to participants at various seminars, including the  Tinbergen Institute (TI) Econometrics Seminar, the Center for Economics and Non-linear Dynamics in Economics and Finance (CENDEF) seminar at the Amsterdam School of Economics, the Business Analytics (BA) seminar of the Amsterdam Business school, as well as the Netherlands Econometrics Study Group (NESG) 2024, for their insightful comments and feedback.

\bibliography{icml2024}
\bibliographystyle{icml2024}

\newpage
\onecolumn
\appendix

\section*{Appendix Outline}
In Section \ref{sec:full_results}, we provide details on the results in Section \ref{sec:OOD_generalization} for the four datasets (Waterbirds, CelebA, MultiNLI and Toy). Section \ref{sec:add_results_Toy} contains  additional results for the Toy dataset. In Section \ref{sec:details_algorithm} we present additional details on the implementation of the JSE algorithm. In Section 
\ref{sec:details_tests} we lay out the testing procedure for the JSE algorithm. Section \ref{sec:gradcam_fig} and \ref{sec:CelebA_removal} provide additional details and images for the experiments conducted in Section \ref{sec:gradcam_main} and \ref{sec:Rafvogel_experiment_main} respectively.  In Section \ref{sec:OOD_other_algorithms}, we compare JSE to other instance-reweighting methods. In Section \ref{sec:multiple_concepts} we illustrate how JSE can be used to deal with multiple spurious concept labels. 
Finally, Section \ref{sec:param_details} provides a more detailed description of the datasets, as well as implementation details for the experiments.  

\section{Full Set of Results for Waterbirds, CelebA, MultiNLI and Toy Dataset}
\label{sec:full_results}
\newcommand{\tablesize}{\fontsize{9.5}{10}\selectfont}
\setlength{\extrarowheight}{1pt}

\vspace{-2.cm}
\begin{table}[H]
\caption{\textbf{Results for the Waterbirds dataset}: Table shows the average, worst-group, and per-group accuracy on a test set where $p_{\mathrm{OOD}}(\ymt = y | \ysp = y) = 0.5$, with $y \in \{0, 1\}$,  as a function of $p_\mathrm{train}(\ymt = y  | \ysp = y ).$ Each accuracy is obtained by averaging over 5 runs. Standard error is reported between brackets. }
\vskip 0.15in
\label{tab:waterbirds}
 \centering
 \tablesize
\begin{tabular}{ll|rrrrr}
\toprule
\multicolumn{1}{l}{\textbf{}}       & \textbf{}                      & \multicolumn{5}{c}{$p_{\mathrm{train}}(\ymt = y | \ysp = y)$}              \\
\multicolumn{1}{l}{\textbf{Method}} & \textbf{Accuracy}              & 0.5          & 0.6          & 0.7          & 0.8          & 0.9          \\
\hline
\multirow{6}{*}{JSE}   & $\ymt = 0$, $\ysp = 0$ & 90.72 (0.83) & 92.43 (0.44) & 90.58 (0.27) & 91.46 (0.76) & 91.85 (0.75) \\
                       & $\ymt = 0$, $\ysp = 1$ & 88.16 (1.08) & 89.05 (0.50) & 87.96 (0.64) & 89.64 (0.50) & 89.56 (0.76) \\
                       & $\ymt = 1$, $\ysp = 0$  & 91.43 (0.24) & 90.16 (0.46) & 91.25 (0.28) & 90.25 (0.41) & 89.63 (0.81) \\
                       & $\ymt = 1$, $\ysp = 1$  & 89.75 (0.57) & 89.56 (0.33) & 90.69 (0.32) & 89.60 (0.53) & 88.75 (0.66) \\
                       & Worst-group            & 87.57 (0.83) & 88.60 (0.36) & 87.96 (0.64) & 88.76 (0.33) & 87.77 (0.36) \\
                       & Average                & 89.70 (0.66) & 90.55 (0.28) & 89.64 (0.28) & 90.41 (0.13) & 90.37 (0.40) \\
                       \hline
\multirow{6}{*}{ERM}   & $\ymt = 0$, $\ysp = 0$ & 90.11 (0.72) & 92.63 (1.70) & 94.12 (0.28) & 96.91 (0.06) & 98.44 (0.16) \\
                       & $\ymt = 0$, $\ysp = 1$ & 88.85 (1.16) & 85.34 (2.35) & 83.62 (0.51) & 83.17 (0.38) & 77.47 (1.44) \\
                       & $\ymt = 1$, $\ysp = 0$  & 91.71 (0.23) & 89.31 (0.51) & 87.94 (0.42) & 81.74 (0.45) & 72.93 (0.87) \\
                       & $\ymt = 1$, $\ysp = 1$  & 89.10 (0.64) & 91.09 (0.84) & 92.49 (0.21) & 92.40 (0.22) & 91.96 (0.43) \\
                       & Worst-group            & 87.31 (0.55) & 85.17 (2.29) & 83.62 (0.51) & 81.74 (0.45) & 72.40 (0.62) \\
                       & Average                & 89.69 (0.65) & 89.25 (1.44) & 89.17 (0.20) & 89.38 (0.16) & 86.73 (0.48) \\
                                              \hline
\multirow{6}{*}{INLP}  & $\ymt = 0$, $\ysp = 0$ & 92.32 (0.49) & 90.26 (0.60) & 85.65 (0.91) & 80.97 (0.73) & 71.80 (1.52) \\
                       & $\ymt = 0$, $\ysp = 1$ & 87.36 (0.63) & 89.39 (0.62) & 90.47 (0.49) & 92.82 (0.31) & 89.79 (0.40) \\
                       & $\ymt = 1$, $\ysp = 0$  & 90.69 (0.35) & 91.62 (0.32) & 93.05 (0.25) & 93.71 (0.17) & 92.74 (0.46) \\
                       & $\ymt = 1$, $\ysp = 1$  & 90.22 (0.30) & 88.88 (0.35) & 88.54 (0.52) & 83.80 (0.40) & 79.69 (0.84) \\
                       & Worst-group            & 87.36 (0.63) & 88.53 (0.26) & 85.17 (0.50) & 80.97 (0.73) & 71.80 (1.52) \\
                       & Average                & 89.98 (0.36) & 89.92 (0.39) & 88.66 (0.46) & 87.30 (0.28) & 82.00 (0.68) \\
                                              \hline
\multirow{6}{*}{LEACE} & $\ymt = 0$, $\ysp = 0$ & 89.32 (0.85) & 87.35 (0.69) & 81.67 (1.41) & 73.72 (1.09) & 61.11 (1.07) \\
                       & $\ymt = 0$, $\ysp = 1$ & 89.43 (1.05) & 92.12 (0.50) & 93.29 (0.55) & 94.08 (0.25) & 91.14 (0.38) \\
                       & $\ymt = 1$, $\ysp = 0$  & 92.12 (0.28) & 92.96 (0.35) & 94.74 (0.23) & 94.95 (0.31) & 95.42 (0.45) \\
                       & $\ymt = 1$, $\ysp = 1$  & 88.72 (0.50) & 87.07 (0.35) & 85.23 (0.78) & 80.97 (0.49) & 78.57 (1.12) \\
                       & Worst-group            & 87.55 (0.33) & 86.24 (0.14) & 81.02 (1.06) & 73.72 (1.09) & 61.11 (1.07) \\
                       & Average                & 89.61 (0.65) & 89.80 (0.38) & 88.04 (0.63) & 84.80 (0.43) & 78.53 (0.47) \\
                                              \hline
\multirow{6}{*}{RLACE} & $\ymt = 0$, $\ysp = 0$ & 89.84 (0.79) & 86.97 (0.88) & 81.19 (1.36) & 73.06 (0.77) & 57.41 (1.14) \\
                       & $\ymt = 0$, $\ysp = 1$ & 89.19 (1.07) & 92.29 (0.40) & 93.46 (0.44) & 94.88 (0.22) & 92.93 (0.27) \\
                       & $\ymt = 1$, $\ysp = 0$  & 92.06 (0.25) & 93.15 (0.36) & 94.61 (0.29) & 95.39 (0.23) & 96.11 (0.24) \\
                       & $\ymt = 1$, $\ysp = 1$  & 88.91 (0.58) & 87.23 (0.23) & 85.02 (0.83) & 80.56 (0.54) & 75.48 (0.63) \\
                       & Worst-group            & 87.51 (0.42) & 86.00 (0.30) & 80.41 (0.81) & 73.06 (0.77) & 57.41 (1.14) \\
                       & Average                & 89.73 (0.64) & 89.75 (0.43) & 87.88 (0.58) & 84.86 (0.29) & 77.53 (0.44) \\
                                              \hline
\multirow{6}{*}{ADV}   & $\ymt = 0$, $\ysp = 0$ & 89.69 (0.55) & 90.58 (2.55) & 93.86 (0.29) & 96.42 (0.18) & 97.70 (0.30) \\
                       & $\ymt = 0$, $\ysp = 1$ & 88.55 (1.01) & 82.17 (3.50) & 83.04 (0.52) & 80.38 (0.62) & 70.89 (2.02) \\
                       & $\ymt = 1$, $\ysp = 0$  & 91.93 (0.24) & 90.53 (0.73) & 88.10 (0.26) & 82.80 (0.46) & 76.76 (1.29) \\
                       & $\ymt = 1$, $\ysp = 1$  & 89.25 (0.49) & 92.24 (0.74) & 92.55 (0.23) & 92.99 (0.32) & 93.36 (0.53) \\
                       & Worst-group            & 87.53 (0.62) & 82.17 (3.50) & 83.04 (0.52) & 80.37 (0.62) & 70.19 (1.67) \\
                       & Average                & 89.44 (0.54) & 87.49 (2.19) & 88.87 (0.26) & 88.29 (0.22) & 84.47 (0.73)\\
                                    \bottomrule
\end{tabular}

\end{table}

\begin{table}[H]
\caption{\textbf{Results for the CelebA dataset}: Table shows the average, worst-group, and per-group accuracy on a test set where $p_{\mathrm{OOD}}(\ymt = y | \ysp = y) = 0.5$, with $y \in \{0, 1\}$, as a function of $p_\mathrm{train}(\ymt = y  | \ysp = y )$. Each accuracy is obtained by averaging over 5 runs. Standard error is reported between brackets. }
\vskip 0.15in

 \centering \tablesize
\begin{tabular}{ll|rrrrr}
\toprule
\multicolumn{1}{l}{\textbf{}}       & \textbf{}                      & \multicolumn{5}{c}{$p_{\mathrm{train}}(\ymt = y | \ysp = y)$}              \\
\multicolumn{1}{l}{\textbf{Method}} & \textbf{Accuracy}              & 0.5          & 0.6          & 0.7          & 0.8          & 0.9          \\
\hline
\multirow{6}{*}{JSE}   & $\ymt = 0$, $\ysp = 0$ & 86.40 (0.35) & 87.56 (0.71) & 87.52 (0.36) & 88.32 (1.57) & 90.36 (0.43) \\
                       & $\ymt = 0$, $\ysp = 1$ & 82.72 (1.23) & 82.24 (0.56) & 81.12 (1.18) & 82.56 (0.77) & 81.84 (1.37) \\
                       & $\ymt = 1$, $\ysp = 0$ & 82.04 (0.70) & 84.48 (1.23) & 82.72 (1.10) & 79.12 (1.74) & 74.40 (1.55) \\
                       & $\ymt = 1$, $\ysp = 1$ & 92.32 (0.37) & 92.68 (0.30) & 92.76 (0.23) & 92.08 (0.57) & 92.84 (0.37) \\
                       & Worst-group            & 80.88 (0.90) & 81.76 (0.56) & 80.00 (0.96) & 78.68 (1.41) & 74.36 (1.51) \\
                       & Average                & 85.87 (0.23) & 86.74 (0.15) & 86.03 (0.33) & 85.52 (0.21) & 84.86 (0.49) \\
                       \hline
\multirow{6}{*}{ERM}   & $\ymt = 0$, $\ysp = 0$ & 86.36 (0.56) & 89.52 (0.90) & 91.52 (0.43) & 93.96 (0.66) & 95.28 (0.35) \\
                       & $\ymt = 0$, $\ysp = 1$ & 82.80 (1.29) & 80.76 (0.41) & 78.40 (1.27) & 75.00 (1.69) & 68.88 (1.38) \\
                       & $\ymt = 1$, $\ysp = 0$ & 82.68 (1.10) & 82.24 (1.31) & 76.76 (1.29) & 63.80 (1.82) & 55.32 (2.15) \\
                       & $\ymt = 1$, $\ysp = 1$ & 92.12 (0.38) & 93.96 (0.37) & 94.68 (0.63) & 95.88 (0.45) & 96.24 (0.47) \\
                       & Worst-group            & 81.28 (1.18) & 80.00 (0.34) & 75.96 (1.19) & 63.80 (1.82) & 55.32 (2.15) \\
                       & Average                & 85.99 (0.33) & 86.62 (0.26) & 85.34 (0.47) & 82.16 (0.81) & 78.93 (0.68) \\
                       \hline
\multirow{6}{*}{INLP}  & $\ymt = 0$, $\ysp = 0$ & 85.92 (0.70) & 81.88 (0.82) & 73.64 (0.69) & 65.64 (1.48) & 58.24 (1.40) \\
                       & $\ymt = 0$, $\ysp = 1$ & 83.08 (1.16) & 85.28 (0.57) & 85.88 (0.46) & 87.96 (0.57) & 87.48 (1.00) \\
                       & $\ymt = 1$, $\ysp = 0$ & 82.64 (0.98) & 88.36 (1.18) & 89.68 (0.58) & 88.76 (0.67) & 86.48 (0.96) \\
                       & $\ymt = 1$, $\ysp = 1$ & 91.60 (0.46) & 87.16 (0.74) & 80.60 (0.56) & 68.08 (0.97) & 59.80 (1.01) \\
                       & Worst-group            & 81.28 (0.81) & 81.88 (0.82) & 73.64 (0.69) & 64.72 (0.92) & 57.40 (1.26) \\
                       & Average                & 85.81 (0.27) & 85.67 (0.28) & 82.45 (0.30) & 77.61 (0.26) & 73.00 (0.52) \\
                       \hline
\multirow{6}{*}{LEACE} & $\ymt = 0$, $\ysp = 0$ & 85.48 (0.73) & 81.52 (0.63) & 72.92 (0.71) & 65.84 (0.99) & 59.64 (0.86) \\
                       & $\ymt = 0$, $\ysp = 1$ & 83.92 (0.92) & 85.80 (0.62) & 86.12 (0.70) & 88.04 (0.29) & 85.48 (1.11) \\
                       & $\ymt = 1$, $\ysp = 0$ & 83.60 (0.87) & 88.80 (1.30) & 89.48 (0.34) & 87.64 (0.60) & 85.68 (0.51) \\
                       & $\ymt = 1$, $\ysp = 1$ & 91.32 (0.45) & 86.52 (0.69) & 80.60 (0.85) & 71.64 (1.68) & 63.20 (0.89) \\
                       & Worst-group            & 82.52 (0.79) & 81.52 (0.63) & 72.92 (0.71) & 65.84 (0.99) & 59.64 (0.86) \\
                       & Average                & 86.08 (0.33) & 85.66 (0.13) & 82.28 (0.29) & 78.29 (0.19) & 73.50 (0.45) \\
                       \hline
\multirow{6}{*}{RLACE} & $\ymt = 0$, $\ysp = 0$ & 85.52 (0.77) & 81.28 (0.87) & 72.64 (0.78) & 64.64 (1.09) & 57.92 (1.29) \\
                       & $\ymt = 0$, $\ysp = 1$ & 83.68 (1.14) & 85.84 (0.75) & 86.36 (0.61) & 89.36 (0.60) & 88.12 (1.34) \\
                       & $\ymt = 1$, $\ysp = 0$ & 83.44 (0.86) & 89.32 (1.22) & 90.60 (0.57) & 90.16 (0.46) & 87.88 (1.28) \\
                       & $\ymt = 1$, $\ysp = 1$ & 91.76 (0.39) & 86.60 (0.66) & 80.60 (0.67) & 69.96 (1.47) & 59.56 (1.33) \\
                       & Worst-group            & 81.96 (0.78) & 81.28 (0.87) & 72.64 (0.78) & 64.64 (1.09) & 57.20 (1.03) \\
                       & Average                & 86.10 (0.34) & 85.76 (0.19) & 82.55 (0.11) & 78.53 (0.29) & 73.37 (0.62) \\
                       \hline
\multirow{6}{*}{ADV}   & $\ymt = 0$, $\ysp = 0$ & 88.04 (0.54) & 87.48 (0.63) & 91.44 (0.15) & 93.44 (0.64) & 95.56 (0.23) \\
                       & $\ymt = 0$, $\ysp = 1$ & 83.72 (1.62) & 82.16 (0.37) & 77.92 (1.21) & 76.36 (0.76) & 69.44 (1.60) \\
                       & $\ymt = 1$, $\ysp = 0$ & 82.80 (2.06) & 77.04 (2.29) & 72.84 (1.75) & 66.52 (2.08) & 57.04 (2.06) \\
                       & $\ymt = 1$, $\ysp = 1$ & 91.68 (0.96) & 93.40 (0.47) & 94.00 (0.36) & 95.96 (0.52) & 96.40 (0.61) \\
                       & Worst-group            & 81.32 (1.24) & 76.64 (1.97) & 72.84 (1.75) & 66.52 (2.08) & 57.04 (2.06) \\
                       & Average                & 86.56 (0.54) & 85.02 (0.51) & 84.05 (0.63) & 83.07 (0.63) & 79.61 (1.08)\\
                                    \bottomrule
\end{tabular}
\label{tab:CelebA}
\end{table}

\begin{table}[H]
\caption{\textbf{Results for the MultiNLI dataset}: Table shows the average, worst-group, and per-group accuracy on a test set where $p_{\mathrm{OOD}}(\ymt = y | \ysp = y) = 0.5$, with $y \in \{0, 1\}$,  as a function of $p_\mathrm{train}(\ymt = y  | \ysp = y )$. Each accuracy is obtained by averaging over 5 runs. Standard error is reported between brackets. }
\vskip 0.15in
 \centering \tablesize
\begin{tabular}{ll|rrrrr}
\toprule
\multicolumn{1}{l}{\textbf{}}       & \textbf{}                      & \multicolumn{5}{c}{$p_{\mathrm{train}}(\ymt = y | \ysp = y)$}              \\
\multicolumn{1}{l}{\textbf{Method}} & \textbf{Accuracy}              & 0.5          & 0.6          & 0.7          & 0.8          & 0.9          \\
\hline
\multirow{6}{*}{JSE}                & $\ymt = 0$, $\ysp = 0$ & 87.49 (0.69) & 88.83 (0.52) & 90.34 (0.36) & 92.53 (0.47) & 93.94 (0.56) \\
                                    & $\ymt = 0$, $\ysp = 1$ & 87.31 (0.42) & 84.30 (0.66) & 82.24 (0.98) & 78.21 (0.92) & 73.97 (1.22) \\
                                    &  $\ymt = 1$, $\ysp = 0$  & 81.74 (0.54) & 79.66 (0.35) & 76.46 (0.35) & 70.93 (1.10) & 63.42 (1.26) \\
                                    & $\ymt = 1$, $\ysp = 1$  & 81.81 (0.31) & 84.86 (0.46) & 86.14 (0.86) & 87.47 (0.42) & 88.18 (0.96) \\
                                    & Worst-group            & 81.38 (0.41) & 79.66 (0.35) & 76.46 (0.35) & 70.93 (1.10) & 63.42 (1.26) \\
                                    & Average                & 84.59 (0.40) & 84.42 (0.23) & 83.80 (0.38) & 82.28 (0.32) & 79.88 (0.33) \\
                                    \hline
\multirow{6}{*}{ERM}                & $\ymt = 0$, $\ysp = 0$ & 87.42 (0.69) & 88.99 (0.55) & 91.06 (0.37) & 94.30 (0.43) & 97.06 (0.29) \\
                                    & $\ymt = 0$, $\ysp = 1$ & 87.20 (0.49) & 83.73 (0.60) & 80.48 (1.11) & 70.24 (1.91) & 51.84 (3.09) \\
                                    & $\ymt = 1$, $\ysp = 0$  & 81.87 (0.48) & 79.65 (0.34) & 75.84 (0.27) & 65.92 (1.84) & 52.05 (1.37) \\
                                    & $\ymt = 1$, $\ysp = 1$  & 81.92 (0.31) & 85.39 (0.55) & 87.57 (0.59) & 91.60 (0.66) & 96.13 (0.70) \\
                                    & Worst-group            & 81.50 (0.35) & 79.65 (0.34) & 75.84 (0.27) & 65.26 (1.72) & 49.70 (2.22) \\
                                    & Average                & 84.60 (0.39) & 84.44 (0.24) & 83.74 (0.31) & 80.52 (0.62) & 74.27 (0.73) \\
                                    \hline
\multirow{6}{*}{INLP}               & $\ymt = 0$, $\ysp = 0$ & 87.44 (0.69) & 88.96 (0.57) & 90.50 (0.41) & 75.09 (4.41) & 56.45 (1.77) \\
                                    & $\ymt = 0$, $\ysp = 1$ & 87.20 (0.49) & 83.84 (0.63) & 81.95 (0.77) & 65.95 (4.04) & 58.26 (4.10) \\
                                    & $\ymt = 1$, $\ysp = 0$  & 81.86 (0.49) & 79.73 (0.33) & 76.51 (0.34) & 62.10 (4.29) & 50.64 (3.92) \\
                                    & $\ymt = 1$, $\ysp = 1$  & 81.94 (0.32) & 85.30 (0.52) & 86.43 (0.67) & 63.12 (7.22) & 54.54 (3.12) \\
                                    & Worst-group            & 81.47 (0.36) & 79.73 (0.33) & 76.51 (0.34) & 57.62 (4.97) & 48.13 (3.12) \\
                                    & Average                & 84.61 (0.39) & 84.46 (0.22) & 83.85 (0.36) & 66.56 (4.51) & 54.97 (2.06) \\
                                    \hline
\multirow{6}{*}{LEACE} & $\ymt = 0$, $\ysp = 0$ & 86.95 (0.56) & 83.25 (0.58) & 79.56 (0.78) & 75.49 (0.80) & 69.39 (0.85) \\
                       & $\ymt = 0$, $\ysp = 1$ & 87.07 (0.41) & 89.36 (0.56) & 91.13 (0.40) & 91.60 (0.38) & 91.77 (0.37) \\
                       & $\ymt = 1$, $\ysp = 0$  & 81.95 (0.69) & 85.11 (0.25) & 87.81 (0.51) & 88.87 (0.70) & 90.15 (0.46) \\
                       & $\ymt = 1$, $\ysp = 1$  & 81.93 (0.37) & 78.29 (0.67) & 75.39 (1.00) & 70.76 (1.65) & 68.05 (1.30) \\
                       & Worst-group            & 81.43 (0.50) & 78.29 (0.67) & 75.39 (1.00) & 70.76 (1.65) & 67.63 (1.11) \\
                       & Average                & 84.47 (0.40) & 84.00 (0.15) & 83.47 (0.32) & 81.68 (0.48) & 79.84 (0.30)\\
                       \hline
\multirow{6}{*}{RLACE}              & $\ymt = 0$, $\ysp = 0$ & 87.16 (0.72) & 88.90 (0.47) & 81.89 (0.82) & 76.61 (1.60) & 69.33 (1.95) \\
                                    & $\ymt = 0$, $\ysp = 1$ & 87.24 (0.55) & 84.02 (0.72) & 91.09 (0.39) & 92.02 (0.69) & 93.70 (0.38) \\
                                    & $\ymt = 1$, $\ysp = 0$  & 81.64 (0.51) & 79.81 (0.31) & 84.59 (1.12) & 86.27 (0.87) & 88.90 (1.04) \\
                                    & $\ymt = 1$, $\ysp = 1$  & 82.16 (0.32) & 85.01 (0.43) & 72.16 (1.18) & 66.69 (1.51) & 61.15 (1.20) \\
                                    & Worst-group            & 81.52 (0.44) & 79.81 (0.31) & 72.16 (1.18) & 66.69 (1.51) & 61.15 (1.20) \\
                                    & Average                & 84.55 (0.43) & 84.43 (0.25) & 82.43 (0.35) & 80.40 (0.64) & 78.27 (0.41)\\
                                    \bottomrule
\end{tabular}
\label{tab:MultiNLI}
\end{table}

\begin{table}[H]
\caption{\textbf{Results for the Toy dataset for} $\rho \in \{0.0, 0.1, 0.2, 0.3, 0.4\}$.  Table shows the average, worst-group, and per-group accuracy on a test set without spurious correlation, as a function of the spurious correlation in the training data. Each accuracy is obtained by averaging over 100 runs. Standard error is reported between brackets. }
\vskip 0.15in

\centering \tablesize
\begin{tabular}{ll|rrrrr}
\toprule
\multicolumn{1}{l}{\textbf{}}       & \textbf{}                      & \multicolumn{5}{c}{$\rho$}              \\
\multicolumn{1}{l}{\textbf{Method}} & \textbf{Accuracy}              & 0.0         & 0.1          & 0.2          & 0.3         & 0.4         \\
\hline
\multirow{6}{*}{JSE}                &   $\ymt = 0$,  $\ysp = 0$ & 83.71 (0.17) & 83.62 (0.18) & 83.46 (0.18) & 83.36 (0.18) & 83.40 (0.16) \\
                                    & $\ymt = 0$,  $\ysp = 1$ & 83.67 (0.16) & 83.56 (0.17) & 83.36 (0.19) & 83.31 (0.18) & 83.24 (0.19) \\
                                    &   $\ymt = 1$, $\ysp = 0$ & 83.73 (0.19) & 83.71 (0.20) & 83.79 (0.19) & 83.71 (0.18) & 83.63 (0.20) \\
                                    & $\ymt = 1$, $\ysp = 1$ & 83.77 (0.15) & 83.83 (0.16) & 83.81 (0.16) & 83.84 (0.17) & 83.82 (0.17) \\
                                    & Worst-group                    & 81.95 (0.12) & 81.86 (0.13) & 81.81 (0.13) & 81.82 (0.13) & 81.68 (0.12) \\
                                    & Average                        & 83.73 (0.09) & 83.69 (0.09) & 83.61 (0.09) & 83.56 (0.09) & 83.53 (0.09) \\
                                    \hline
\multirow{6}{*}{ERM}                & $\ymt = 0$, $\ysp = 0$ & 83.55 (0.16) & 83.46 (0.18) & 83.42 (0.19) & 83.52 (0.21) & 83.85 (0.23) \\
                                    &  $\ymt = 0$, $\ysp = 1$ & 83.82 (0.17) & 83.73 (0.17) & 83.38 (0.19) & 83.17 (0.22) & 82.44 (0.29) \\
                                    &  $\ymt = 1$, $\ysp = 0$ & 83.89 (0.19) & 83.83 (0.20) & 83.72 (0.20) & 83.50 (0.21) & 82.82 (0.35) \\
                                    & $\ymt = 1$, $\ysp = 1$ & 83.61 (0.16) & 83.72 (0.16) & 83.83 (0.16) & 83.92 (0.20) & 84.22 (0.21) \\
                                    & Worst-group                    & 81.90 (0.13) & 81.80 (0.13) & 81.63 (0.13) & 81.41 (0.15) & 80.72 (0.25) \\
                                    & Average                        & 83.74 (0.08) & 83.70 (0.08) & 83.60 (0.09) & 83.54 (0.09) & 83.35 (0.10) \\
                                    \hline
\multirow{6}{*}{INLP}               & $\ymt = 0$, $\ysp = 0$ & 83.76 (0.18) & 83.07 (0.29) & 78.93 (0.80) & 72.41 (1.16) & 64.74 (1.17) \\
                                    &  $\ymt = 0$, $\ysp = 1$ & 83.64 (0.17) & 83.44 (0.24) & 80.08 (0.74) & 74.74 (1.10) & 67.58 (1.24) \\
                                    &  $\ymt = 1$, $\ysp = 0$ & 83.60 (0.22) & 83.66 (0.25) & 80.52 (0.73) & 75.42 (1.08) & 68.16 (1.21) \\
                                    & $\ymt = 1$, $\ysp = 1$ & 83.79 (0.16) & 83.43 (0.23) & 79.52 (0.82) & 72.99 (1.14) & 65.16 (1.16) \\
                                    & Worst-group                    & 81.72 (0.14) & 81.26 (0.24) & 77.32 (0.81) & 70.58 (1.19) & 61.41 (1.24) \\
                                    & Average                        & 83.70 (0.09) & 83.41 (0.18) & 79.77 (0.75) & 73.90 (1.09) & 66.45 (1.11) \\
                                    \hline
\multirow{6}{*}{LEACE} & $\ymt = 0$, $\ysp = 0$ & 84.05 (0.17) & 80.84 (0.20) & 77.17 (0.20) & 73.39 (0.21) & 69.41 (0.21) \\
                       & $\ymt = 0$, $\ysp = 1$ & 83.67 (0.18) & 85.95 (0.16) & 87.71 (0.14) & 89.06 (0.14) & 89.93 (0.14) \\
                       & $\ymt = 1$, $\ysp = 0$  & 83.64 (0.19) & 86.23 (0.18) & 88.09 (0.15) & 89.38 (0.14) & 90.38 (0.13) \\
                       & $\ymt = 1$, $\ysp = 1$  & 83.69 (0.18) & 80.95 (0.20) & 77.60 (0.20) & 73.79 (0.21) & 70.10 (0.21) \\
                       & Worst-group            & 81.82 (0.13) & 79.74 (0.18) & 76.26 (0.17) & 72.38 (0.17) & 68.55 (0.17) \\
                       & Average                & 83.77 (0.09) & 83.50 (0.08) & 82.65 (0.08) & 81.42 (0.09) & 79.98 (0.10)\\
                       \hline
\multirow{6}{*}{RLACE}              & $\ymt = 0$, $\ysp = 0$ & 83.69 (0.21) & 81.44 (0.23) & 78.52 (0.26) & 75.37 (0.26) & 72.31 (0.27) \\
                                    &  $\ymt = 0$, $\ysp = 1$ & 83.49 (0.20) & 85.42 (0.18) & 86.90 (0.17) & 88.25 (0.14) & 89.08 (0.13) \\
                                    &  $\ymt = 1$, $\ysp = 0$ & 83.57 (0.25) & 85.43 (0.23) & 87.13 (0.18) & 88.47 (0.16) & 89.42 (0.14) \\
                                    & $\ymt = 1$, $\ysp = 1$ & 83.60 (0.21) & 81.67 (0.25) & 79.27 (0.26) & 75.98 (0.28) & 72.99 (0.28) \\
                                    & Worst-group                    & 81.20 (0.16) & 80.30 (0.21) & 77.72 (0.24) & 74.53 (0.25) & 71.38 (0.25) \\
                                    & Average                        & 83.59 (0.09) & 83.49 (0.09) & 82.96 (0.10) & 82.02 (0.11) & 80.95 (0.11)\\
                                    \bottomrule
\end{tabular}
\label{tab:Toy_1}
\end{table}

\begin{table}[H]
 \caption{\textbf{Results for the Toy dataset for} $\rho \in \{0.5, 0.6, 0.7, 0.8, 0.9\}$.  Table shows the average, worst-group, and per-group accuracy on a test set without spurious correlation, as a function of the spurious correlation in the training data. Each accuracy is obtained by averaging over 100 runs. Standard error is reported between brackets. }
\label{tab:Toy_2}
\vskip 0.15in
\centering \tablesize
\begin{tabular}{ll|rrrrr}
\toprule
\multicolumn{1}{l}{\textbf{}}       & \textbf{}                      & \multicolumn{5}{c}{$\rho$}              \\
\multicolumn{1}{l}{\textbf{Method}} & \textbf{Accuracy}              & 0.5          & 0.6          & 0.7          & 0.8          & 0.9          \\
\hline
\multirow{6}{*}{JSE}   & $\ymt = 0$, $\ysp = 0$ & 83.27 (0.16) & 83.31 (0.17) & 83.25 (0.18) & 83.30 (0.18) & 83.49 (0.26) \\
                       &  $\ymt = 0$, $\ysp = 1$ & 83.20 (0.18) & 83.06 (0.18) & 82.99 (0.20) & 83.06 (0.21) & 81.87 (0.45) \\
                       &  $\ymt = 1$, $\ysp = 0$ & 83.57 (0.20) & 83.54 (0.21) & 83.48 (0.22) & 83.31 (0.20) & 82.35 (0.47) \\
                       & $\ymt = 1$, $\ysp = 1$ & 83.67 (0.18) & 83.71 (0.18) & 83.76 (0.18) & 83.65 (0.19) & 84.05 (0.25) \\
                       & Worst-group                    & 81.54 (0.12) & 81.55 (0.13) & 81.42 (0.14) & 81.44 (0.14) & 80.27 (0.42) \\
                       & Average                        & 83.43 (0.09) & 83.41 (0.10) & 83.38 (0.09) & 83.33 (0.10) & 82.94 (0.18) \\
                       \hline
\multirow{6}{*}{ERM}   & $\ymt = 0$, $\ysp = 0$ & 83.61 (0.22) & 83.76 (0.27) & 84.60 (0.31) & 86.00 (0.31) & 88.72 (0.28) \\
                       &  $\ymt = 0$, $\ysp = 1$ & 82.76 (0.28) & 82.10 (0.40) & 80.39 (0.53) & 77.80 (0.77) & 69.97 (0.88) \\
                       &  $\ymt = 1$, $\ysp = 0$ & 83.03 (0.30) & 82.50 (0.42) & 80.62 (0.60) & 77.78 (0.79) & 70.11 (0.94) \\
                       & $\ymt = 1$, $\ysp = 1$ & 83.93 (0.22) & 84.16 (0.24) & 84.97 (0.28) & 85.99 (0.32) & 88.86 (0.26) \\
                       & Worst-group                    & 80.76 (0.22) & 80.06 (0.36) & 78.51 (0.53) & 76.01 (0.77) & 68.41 (0.91) \\
                       & Average                        & 83.35 (0.09) & 83.15 (0.12) & 82.67 (0.17) & 81.90 (0.26) & 79.43 (0.35) \\
                       \hline
\multirow{6}{*}{INLP}  & $\ymt = 0$, $\ysp = 0$ & 59.94 (1.03) & 58.54 (0.90) & 55.66 (0.81) & 54.71 (0.68) & 53.06 (0.69) \\
                       &  $\ymt = 0$, $\ysp = 1$ & 62.83 (1.19) & 61.77 (1.26) & 58.31 (1.29) & 56.64 (1.27) & 56.01 (1.53) \\
                       &  $\ymt = 1$, $\ysp = 0$ & 62.82 (1.19) & 60.43 (1.31) & 58.98 (1.26) & 57.09 (1.26) & 58.23 (1.48) \\
                       & $\ymt = 1$, $\ysp = 1$ & 60.52 (1.01) & 57.87 (0.95) & 56.60 (0.82) & 55.09 (0.65) & 55.14 (0.70) \\
                       & Worst-group                    & 55.51 (1.06) & 53.06 (0.95) & 50.06 (0.77) & 48.59 (0.55) & 47.06 (0.56) \\
                       & Average                        & 61.58 (0.94) & 59.71 (0.89) & 57.44 (0.76) & 55.93 (0.68) & 55.67 (0.74) \\
                       \hline
\multirow{6}{*}{LEACE} & $\ymt = 0$, $\ysp = 0$ & 65.63 (0.20) & 62.01 (0.22) & 58.69 (0.22) & 55.42 (0.23) & 52.64 (0.25) \\
                       & $\ymt = 0$, $\ysp = 1$ & 90.76 (0.12) & 91.21 (0.12) & 91.49 (0.12) & 91.70 (0.12) & 91.63 (0.14) \\
                       & $\ymt = 1$, $\ysp = 0$  & 91.02 (0.13) & 91.61 (0.12) & 91.82 (0.13) & 92.11 (0.12) & 92.05 (0.14) \\
                       & $\ymt = 1$, $\ysp = 1$  & 65.99 (0.23) & 62.41 (0.21) & 58.94 (0.25) & 55.85 (0.26) & 52.96 (0.29) \\
                       & Worst-group            & 64.60 (0.17) & 60.95 (0.18) & 57.37 (0.18) & 54.02 (0.19) & 50.97 (0.20) \\
                       & Average                & 78.37 (0.10) & 76.83 (0.09) & 75.27 (0.10) & 73.80 (0.09) & 72.35 (0.10)\\
                       \hline
\multirow{6}{*}{RLACE} & $\ymt = 0$, $\ysp = 0$ & 68.59 (0.27) & 64.77 (0.27) & 60.84 (0.29) & 57.30 (0.28) & 53.89 (0.30) \\
                       &  $\ymt = 0$, $\ysp = 1$ & 89.97 (0.14) & 90.60 (0.13) & 91.02 (0.12) & 91.21 (0.13) & 90.71 (0.30) \\
                       &  $\ymt = 1$, $\ysp = 0$ & 90.22 (0.13) & 90.87 (0.13) & 91.30 (0.13) & 91.65 (0.14) & 90.89 (0.39) \\
                       & $\ymt = 1$, $\ysp = 1$ & 69.32 (0.33) & 65.42 (0.31) & 61.70 (0.31) & 57.85 (0.32) & 54.39 (0.35) \\
                       & Worst-group                    & 67.65 (0.27) & 63.87 (0.27) & 59.74 (0.27) & 55.89 (0.25) & 52.31 (0.27) \\
                       & Average                        & 79.53 (0.13) & 77.92 (0.13) & 76.22 (0.13) & 74.50 (0.13) & 72.48 (0.20)\\
                                    \bottomrule
\end{tabular}

\end{table}

\newpage
\section{Additional Results for Toy dataset}
\label{sec:add_results_Toy}
\subsection{Removing the Orthogonality Assumption}
\label{sec:orth_assumption}

\begin{figure}[H]
\centering
{\input{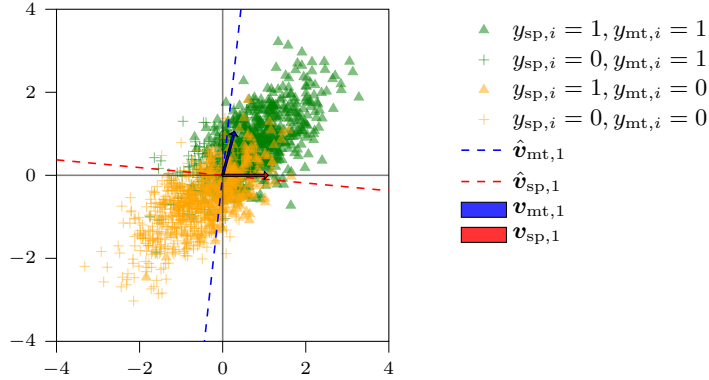}}
\caption{\textbf{Illustration of vectors found by JSE  with non-orthogonal spurious and main-task subspaces}: The blue and red arrow indicate the basis of the spurious and main-task subspaces, which have an angle of  $75^{\circ}$. Data is based on a single simulation from the $d(=20)$-dimensional Toy dataset of Section~\ref{sec:datasets} with $\rho =0.8 $ and sample size $n = $ 2,000. 
}
\label{fig:orthogonal_illustration}
\end{figure}

In this section, we provide an analysis of how JSE performs when the $\Zsp$ and $\Zmt$ are not orthogonal subspaces. We create an example of the Toy dataset where the orthogonality assumption does not hold, by changing the angle of $\boldsymbol{w}_{\mathrm{sp}}$ and $\boldsymbol{w}_{\mathrm{mt}}$ to $75^{\circ}$. Let $a = \cos(\frac{15\pi}{180}), b = \sin(\frac{15\pi}{180})$, and  $\boldsymbol{w}_{\mathrm{sp}} = \begin{pmatrix} \gamma, 0,  0, \dots, 0 \end{pmatrix}^{\top}$ and $\boldsymbol{w}_{\mathrm{mt}} = \begin{pmatrix} \frac{\gamma}{ 1 + \frac{a}{b}}, \frac{\gamma}{ 1 + \frac{b}{a}},  0 \dots, 0 \end{pmatrix}^{\top} $. The main-task labels are now determined by a linear combination of the spurious and main-task directions. In Figure \ref{fig:orthogonal_illustration}, we illustrate the subspaces found by JSE for this scenario. JSE finds a spurious and main-task vector that are slightly different from the basis vectors, yet orthogonal. This illustrates that when the $\Zsp$ and $\Zmt$ are not orthogonal subspaces, JSE finds two orthogonal subspaces that best fit given the data.

\setlength\figureheight{6cm}
\setlength\figurewidth{6cm}

\begin{figure}[H]
\centering
{\input{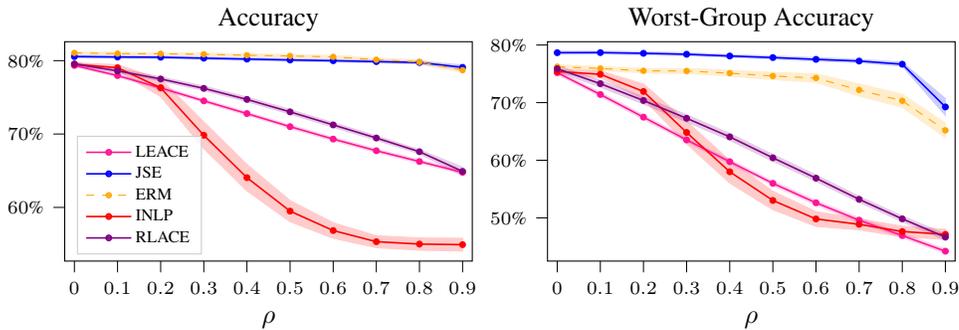}}
\vspace{-0.25cm}
\caption{\textbf{OOD generalization for Toy dataset, when the angle between the spurious and main-task vector is $75^{\circ}$}: We plot the (worst-group) accuracy on a test set without spurious correlation, as a function of the spurious correlation in the training data. Each accuracy is obtained by averaging over 100 runs. The shaded area reflects the 95\% confidence interval.}
\label{fig:non_orth}
\end{figure}

Figure \ref{fig:non_orth} shows the performance for the Toy dataset when the subspaces are not orthogonal. Compared to the case where the subspaces are orthogonal, the performance of JSE is slightly worse. It removes a small part of the main-task direction, and leaves a small part of the spurious direction. However, it still outperforms other concept-removal methods. These methods perform relatively worse, because they also remove part of the main-task direction - even when there is no correlation between the spurious and main-task direction. When the spurious direction is removed, the part of the main-task direction that is non-orthogonal to it is also removed.

\setlength\figureheight{4.5cm}
\setlength\figurewidth{7cm}

\subsection{Finite-sample Estimation Noise for JSE and INLP} \label{sec:finite_sample_noise}

\begin{figure}[H]
\centering
\input{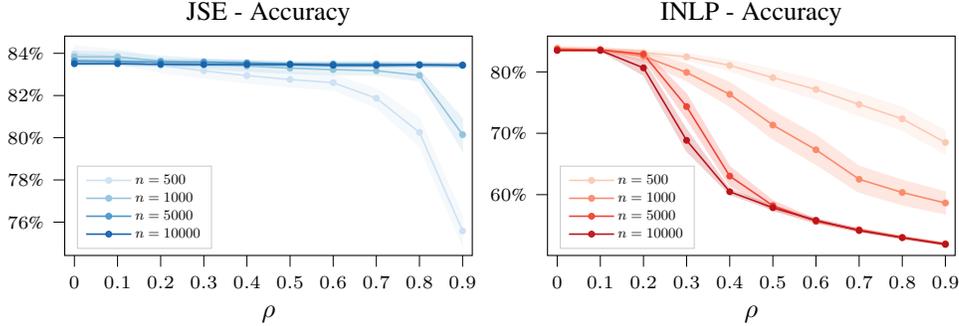}
\vspace{-0.25cm}
\caption{\textbf{Effect of training set size and spurious correlation strength for Toy dataset}: We plot the accuracy on a test set without spurious correlation, as a function of the spurious correlation in the training data. Each accuracy is obtained by averaging over 100 runs. The shaded area reflects the 95\% confidence interval.} 
\label{fig:finite_sample_noise}
\end{figure}

In this section, we briefly illustrate how the performance of JSE and INLP is affected by the interaction of (i) the size of the training set, and (ii) the correlation between the spurious and main-task features. Figure \ref{fig:finite_sample_noise} shows the result of applying JSE and INLP to the Toy dataset for different sizes of the dataset. For JSE, we observe that with a limited sample size and a high spurious correlation, it is harder to separate the spurious and main-task features. We attribute this to finite-sample noise:  our method finds two orthogonal vectors that fit well in the training data, but they are less likely to align with the data-generating process. Interestingly, INLP becomes worse as the sample size increases. With a greater sample size, it is more likely that INLP assigns the main-task feature as belonging to the spurious subspace, since its predictive ability of the spurious concept label is more likely to be detected. 

\section{The JSE Algorithm}
\label{sec:details_algorithm}

In this section, we first provide a more detailed version of the JSE algorithm, initially described in Section~\ref{sec:iterative}. We briefly investigate two modifications  of the algorithm in Section \ref{sec:symmetry}. Finally, we perform a brief ablation on the Waterbirds for how JSE performs without first applying PCA to the embeddings in Section \ref{sec:PCA}.

\subsection{Detailed Description of the JSE Algorithm}
\label{sec:detailed_description_algorithm}

In Equation \ref{eq:estimation} we note that we optimize over the coefficients for two logistic regressions, while enforcing an orthogonality constraint. Here, we briefly explain how this constraint is enforced. We solve the following unconstrained problem using stochastic gradient descent (SGD):
\begin{align*} 
    \whatsp, \whatmt, \bhatsp, \bhatmt = 
    \argmin_{\substack{ \wsp, \wmt, \bsp, \bmt}}
    \sum^{n}_{i = 1} 
    \mathcal{L}_{\mathrm{BCE}}(\yhatspsub{i}, \yspsub{i}) +   \mathcal{L}_{\mathrm{BCE}}(\yhatmtsub{i}, \ymtsub{i}),
\end{align*}
where we define the predictions as follows:
\begin{align*}
 \yhatspsub{i} &= \mathrm{Logit}^{-1} \left(\bz_i^{\top} \wsp + b_{\mathrm{sp}}\right),\\
 \yhatmtsub{i} &=  \mathrm{Logit}^{-1} \left(\bz_i^{\top} (\boldsymbol{I} - \boldsymbol{P}_{\wsp}) \wmt + b_{\mathrm{mt}}\right), \\
 \boldsymbol{P}_{\wsp} &= \wsp \left(\wsp^{\top}\wsp\right)^{-1}\wsp^{\top}.
\end{align*}
By definition, $(\boldsymbol{I} - \boldsymbol{P}_{\wsp}) \wmt$ is orthogonal to $\wsp$. Therefore, the predictions for each $\ymtsub{i}$ are based upon a set of coefficients that is orthogonal to $\wsp$. 
By weighing both losses equally, there is no reason for this optimization problem to favor the loss for one set of labels over the other.


In Algorithm \ref{alg:JSE_detailed} we provide the exact same procedure as in Algorithm \ref{alg:JSE_basic}, but in greater detail. 

\begin{algorithm}[H]
\caption{JSE algorithm to estimate orthonormal bases for $\Zsp$ and $\Zmt.$ The calculation of the test statistics is discussed in Section \ref{sec:test_stat}}
\label{alg:JSE_detailed}
\begin{algorithmic}
\REQUIRE a sample $\left\{ \ymtsub{k}, \yspsub{k}, \bz_k \right\}_{k=1}^n$ consisting of two binary labels and a vector $\bz_k \in \mathbb{R}^d.$
\STATE Initialize a $(n \times d)$-dimensional embedding matrix $\bZ = \begin{pmatrix}\bz_1 \, \bz_2 \, \cdots \, \bz_n \end{pmatrix}^{\top}.$
\STATE Initialize $\Zsportho \leftarrow \bZ.$
\STATE Choose a significance level $\alpha$ for the test statistics, resulting in critical value $t_{1-\alpha}$
\FOR{$i = 1, ..., d$}
    \STATE $\Zremain \leftarrow \Zsportho$ 
    \FOR{$j =1, ..., d$}
        \STATE 
        \STATE Estimate $\whatsp$ and $\whatmt$ with \Eqref{eq:estimation}, using embeddings $\Zremain.$
        \STATE 
        \STATE Define normalizations $\vhatspsub{i} \leftarrow \whatsp / || \whatsp ||$ and $\vhatmtsub{j} \leftarrow \whatmt / || \whatmt ||.$
        \STATE 
        \STATE  Estimate $ \displaystyle \hat{\gamma}_{\mathrm{sp}}, \hat{b}_{\mathrm{sp}} = \argmin_{\substack{\gamma_{\mathrm{sp}}, b_{\mathrm{sp}} }} \sum^n_{h=1} \mathcal{L}_{\mathrm{BCE}}(\yhatspsub{h}^{(\vhatmtsub{j})}  , \yspsub{h})$, 
        where $\displaystyle \yhatspsub{h}^{(\vhatmtsub{j})} = \mathrm{Logit}^{-1}\left(\gamma_\mathrm{sp} \,\bz_{h, \mathrm{remain}}^{\top} \vhatmtsub{j} + b_\mathrm{sp}\right)$
        \STATE 
        \STATE Calculate $t_{\mathrm{mt}, \mathrm{rnd}},  t^{(\vhatmtsub{j})}_{\mathrm{mt}, \mathrm{sp}}$ using $ \displaystyle \hat{\gamma}_{\mathrm{sp}}, \hat{b}_{\mathrm{sp}}$  (see Section \ref{sec:test_stat})
        \STATE 
        \IF{($t_{\mathrm{mt}, \mathrm{rnd}} < -t_{1-\alpha}$ ) and ($t^{(\vhatmtsub{j})}_{\mathrm{mt}, \mathrm{sp}} > t_{1-\alpha}$)}
            \STATE Projection $\Zremain \leftarrow \Zsportho (\boldsymbol{I} - \Vhatmt\Vhatmt^{\top})$, where  $\Vhatmt = \begin{pmatrix}\vhatmtsub{1} \, \vhatmtsub{2} \, \cdots \, \vhatmtsub{j} \end{pmatrix}$
        \ELSE
            \STATE \textbf{break}
        \ENDIF
    \ENDFOR
    \STATE 
    \STATE  Estimate $ \displaystyle \hat{\gamma}_{\mathrm{mt}}, \hat{b}_{\mathrm{mt}} = \sum^n_{h=1}\argmin_{\substack{\gamma_{\mathrm{mt}}, b_{\mathrm{mt}} }} \mathcal{L}_{\mathrm{BCE}}(\yhatmtsub{h}^{(\vhatspsub{i})}  , \ymtsub{h})$, where $\displaystyle \yhatmtsub{h}^{(\vhatspsub{i})} = \mathrm{Logit}^{-1}\left(\gamma_{\mathrm{mt}} \zsportho \vhatspsub{i} + b_\mathrm{sp}\right)$
        \STATE 
        \STATE  Calculate $t_{\mathrm{sp}, \mathrm{rnd}},  t^{(\vhatspsub{i})}_{\mathrm{mt}, \mathrm{sp}}$ using $\displaystyle \hat{\gamma}_{\mathrm{mt}}, \hat{b}_{\mathrm{mt}}$ (see Section \ref{sec:test_stat})
        \STATE 
    \IF{($t_{\mathrm{sp}, \mathrm{rnd}} < -t_{1-\alpha}$ ) and ($t^{(\vhatspsub{i})}_{\mathrm{mt}, \mathrm{sp}} < -t_{1-\alpha}$)}
        \STATE Projection $\Zsportho \leftarrow \bZ (\boldsymbol{I} - \Vhatsp\Vhatsp^{\top}),$ where $\Vhatsp = \begin{pmatrix}\vhatspsub{1} \, \vhatspsub{2} \, \cdots \, \vhatspsub{i} \end{pmatrix}.$
        \STATE $\hat{\boldsymbol{v}}'_{\mathrm{mt},{\ell}} \leftarrow \vhatmtsub{\ell},$ for $\ell=1,2,\ldots,\ell_{0},$ with $\ell_{0} = j.$
    \ELSE
        \STATE \textbf{break}
    \ENDIF
\ENDFOR
\STATE \textbf{return} Bases $\{\vhatspsub{m}\}_{m=1}^{i-1}$ and $\{\hat{\boldsymbol{v}}'_{\mathrm{mt},{\ell}}\}_{\ell=1}^{\ell_{0}-1}.$
\end{algorithmic}
\end{algorithm}

\subsection{Modifications of the JSE Algorithm}
\label{sec:symmetry}

In this section, we briefly compare the formulation of the JSE algorithm to two alternatives. 
\begin{itemize}
    \item \textbf{Swapping the loops:} one might consider interchanging the inner loop and the outer loop of the JSE algorithm. Now, at each step, first the spurious vectors $\vhatspsub{1}, \vhatspsub{2}, \ldots, \vhatspsub{\dsp}$ are projected out before the main-task vector (at that step) is estimated. 
    \item \textbf{Projecting onto main-task subspace}: instead of projecting $\boldsymbol{z}$ onto the orthogonal complement of $\Zsp$, one could be interested in projecting onto $\Zmt$. In this case, rather than the transformation $(\boldsymbol{I} - \Vsp \Vsp^{\top}) \bz$, one uses the transformed embeddings $(\Vmt \Vmt^{\top}) \bz$. 
\end{itemize}

In Table \ref{tab:JSE_ablation}   we compare these two alternative formulations of the algorithm for the Waterbirds dataset. The performance for these two versions is similar to the JSE algorithm as outlined in Section \ref{sec:iterative}. We suggest that in practice, which version of the algorithm should be used could depend on (1) whether the user wants to remove a certain spurious concept, or isolate certain main-task features, and (2) the dimensionality of both subspaces. For example, if the dimension of the main-task subspace is much lower than that of the spurious concept subspace, it might be worthwhile to project onto  $\zmt$ to improve the bias-variance trade-off, rather than removing $\zsp$. 

\begin{table}[H]
\caption{\textbf{Results for the Waterbirds dataset for different versions of the JSE algorithm}: Table shows the average, worst-group, and per-group accuracy on a test set where $p_{\mathrm{OOD}}(\ymt = y | \ysp = y) = 0.5$,  with $y \in \{0, 1\}$. Each accuracy is obtained by averaging over 5 runs. Standard error is reported between brackets. }
\label{tab:JSE_ablation}
\vskip 0.15in

 \centering \tablesize
\begin{tabular}{cllllll}
\toprule
\multicolumn{1}{l}{\textbf{}}       & \textbf{}                      & \multicolumn{5}{c}{$p_{\mathrm{train}}(\ymt = y | \ysp = y)$}              \\
\multicolumn{1}{l}{\textbf{Method}} & \textbf{Accuracy}              & 0.5          & 0.6          & 0.7          & 0.8          & 0.9          \\
\hline
\multirow{6}{*}{\begin{tabular}[c]{@{}c@{}}JSE with\\ projecting\\ onto $\zmt$\end{tabular}}  & $\ymt = 0$, $\ysp = 0$ & 90.98 (0.45) & 91.90 (0.46) & 89.06 (0.72) & 89.39 (0.24) & 92.39 (0.65) \\
                                                                                                             & $\ymt = 0$, $\ysp = 1$ & 88.52 (0.59) & 89.38 (0.39) & 89.37 (0.32) & 91.54 (0.23) & 89.22 (0.91) \\
                                                                                                             & $\ymt = 1$, $\ysp = 0$  & 91.00 (0.18) & 90.81 (0.49) & 92.15 (0.41) & 91.56 (0.25) & 88.32 (0.80) \\
                                                                                                             & $\ymt = 1$, $\ysp = 1$  & 89.72 (0.40) & 88.94 (0.39) & 90.12 (0.38) & 88.01 (0.48) & 88.82 (0.75) \\
                                                                                                             & Worst-group            & 88.43 (0.55) & 88.47 (0.17) & 88.61 (0.45) & 88.01 (0.48) & 86.86 (0.26) \\
                                                                                                             & Average                & 89.89 (0.35) & 90.47 (0.24) & 89.64 (0.33) & 90.31 (0.09) & 90.31 (0.42) \\
                                                                                                             \hline
\multirow{6}{*}{\begin{tabular}[c]{@{}c@{}}JSE with\\ Swapped \\ Loops\end{tabular}} & $\ymt = 0$, $\ysp = 0$ & 91.59 (0.72) & 90.71 (1.91) & 91.16 (0.32) & 91.49 (0.64) & 90.83 (0.67) \\
                                                                                                             & $\ymt = 0$, $\ysp = 1$ & 88.39 (0.79) & 87.65 (1.80) & 89.05 (0.72) & 90.40 (0.35) & 89.18 (0.90) \\
                                                                                                             & $\ymt = 1$, $\ysp = 0$  & 89.91 (0.33) & 90.53 (0.54) & 90.81 (0.29) & 90.28 (0.20) & 89.28 (0.61) \\
                                                                                                             & $\ymt = 1$, $\ysp = 1$  & 89.25 (0.46) & 90.16 (0.65) & 89.28 (0.26) & 88.82 (0.47) & 87.91 (0.55) \\
                                                                                                             & Worst-group            & 87.83 (0.59) & 87.16 (1.58) & 88.34 (0.46) & 88.75 (0.44) & 87.11 (0.30) \\
                                                                                                             & Average                & 89.90 (0.50) & 89.44 (1.31) & 90.09 (0.35) & 90.64 (0.19) & 89.69 (0.41)\\
                                                                                           
                                                           \bottomrule
\end{tabular}
\end{table}

\subsection{Using JSE With and Without First Applying PCA to the Embeddings}
\label{sec:PCA}

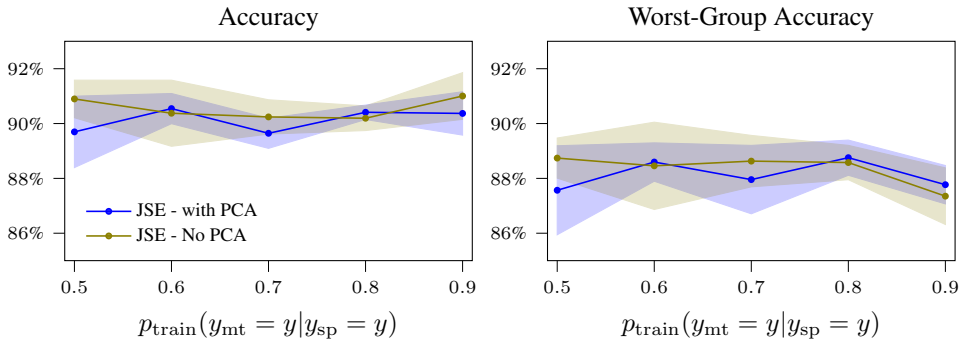
\begin{figure}[H]
\centering
\begin{tikzpicture}

\definecolor{darkgray176}{RGB}{176,176,176}
\definecolor{darkblue}{RGB}{0,100,115}
\definecolor{darkyellow}{RGB}{139, 128, 0}

\begin{groupplot}[group style={group size=2 by 1}]
\nextgroupplot[
height=\figureheight,
legend cell align={left},
legend style={
  fill opacity=0.8,
  draw opacity=1,
  text opacity=1,
  at={(0.03,0.03)},
  anchor=south west,
  draw=none,
  nodes={scale=0.8}
},
tick align=outside,
tick pos=left,
title={Accuracy},
width=\figurewidth,
x grid style={darkgray176},
xlabel={\(\displaystyle p_{\mathrm{train}}(\ymt = y | \ysp = y) \)},
xmin=0.49, xmax=0.91,
xtick style={color=black},
y grid style={darkgray176},
ymin=85, ymax=93,
ytick style={color=black},
yticklabel={\pgfmathprintnumber\tick\%},
title style={yshift=-0.2cm,}
]

\path [draw=blue, fill=blue, opacity=0.2]
(axis cs:0.5,90.9988598267963)
--(axis cs:0.5,88.393613870866)
--(axis cs:0.6,89.9943578978303)
--(axis cs:0.7,89.0949223701444)
--(axis cs:0.8,90.1464335490002)
--(axis cs:0.9,89.5778144807606)
--(axis cs:0.9,91.1608825712414)
--(axis cs:0.9,91.1608825712414)
--(axis cs:0.8,90.6751044224964)
--(axis cs:0.7,90.193996601871)
--(axis cs:0.6,91.0964261750456)
--(axis cs:0.5,90.9988598267963)
--cycle;

\path [draw=darkyellow, fill=darkyellow, opacity=0.2]
(axis cs:0.5,91.5828299134164)
--(axis cs:0.5,90.2121305853934)
--(axis cs:0.6,89.1702140959091)
--(axis cs:0.7,89.6158015695933)
--(axis cs:0.8,89.7442190930114)
--(axis cs:0.9,90.1528256781012)
--(axis cs:0.9,91.8561488740533)
--(axis cs:0.9,91.8561488740533)
--(axis cs:0.8,90.6423909381166)
--(axis cs:0.7,90.8674585851308)
--(axis cs:0.6,91.5822873918228)
--(axis cs:0.5,91.5828299134164)
--cycle;

\addplot [semithick, blue, mark=*, mark size=1, mark options={solid}]
table {%
0.5 89.6962368488312
0.6 90.545392036438
0.7 89.6444594860077
0.8 90.4107689857483
0.9 90.369348526001
};
\addlegendentry{JSE - with PCA}

\addplot [semithick, darkyellow, mark=*, mark size=1, mark options={solid}]
table {%
0.5 90.8974802494049
0.6 90.376250743866
0.7 90.2416300773621
0.8 90.193305015564
0.9 91.0044872760773
};
\addlegendentry{JSE - No PCA}

\nextgroupplot[
height=\figureheight,
legend cell align={left},
legend style={
  at={(0.03,0.03)},
  anchor=south west,
  draw=none,
  fill opacity=0.0,
  draw opacity=1,
  text opacity=1,
  nodes={scale=0.8}
},
tick align=outside,
tick pos=left,
title={Worst-Group Accuracy},
width=\figurewidth,
x grid style={darkgray176},
xlabel={\(\displaystyle p_{\mathrm{train}}(\ymt = y | \ysp = y) \)},
xmin=0.49, xmax=0.91,
xtick style={color=black},
y grid style={darkgray176},
ymin=85, ymax=93,
ytick style={color=black},
yticklabel={\pgfmathprintnumber\tick\%},
title style={yshift=-0.2cm,}
]
\path [draw=blue, fill=blue, opacity=0.2]
(axis cs:0.5,89.1907119750977)
--(axis cs:0.5,85.9433212280273)
--(axis cs:0.6,87.8955078125)
--(axis cs:0.7,86.7073593139648)
--(axis cs:0.8,88.1130065917969)
--(axis cs:0.9,87.069221496582)
--(axis cs:0.9,88.4702072143555)
--(axis cs:0.9,88.4702072143555)
--(axis cs:0.8,89.3975219726562)
--(axis cs:0.7,89.2039413452148)
--(axis cs:0.6,89.2966003417969)
--(axis cs:0.5,89.1907119750977)
--cycle;

\path [draw=darkyellow, fill=darkyellow, opacity=0.2]
(axis cs:0.5,89.4719470526161)
--(axis cs:0.5,88.0089642022668)
--(axis cs:0.6,86.8652312074463)
--(axis cs:0.7,87.6913162359765)
--(axis cs:0.8,87.9514162065127)
--(axis cs:0.9,86.3192941679366)
--(axis cs:0.9,88.3842733369462)
--(axis cs:0.9,88.3842733369462)
--(axis cs:0.8,89.2042997358701)
--(axis cs:0.7,89.568098436875)
--(axis cs:0.6,90.0503419126709)
--(axis cs:0.5,89.4719470526161)
--cycle;

\addplot [semithick, blue, mark=*, mark size=1, mark options={solid}]
table {%
0.5 87.5670166015625
0.6 88.5960540771484
0.7 87.9556503295898
0.8 88.7552642822266
0.9 87.7697143554688
};

\addplot [semithick, darkyellow, mark=*, mark size=1, mark options={solid}]
table {%
0.5 88.7404556274414
0.6 88.4577865600586
0.7 88.6297073364258
0.8 88.5778579711914
0.9 87.3517837524414
};

\end{groupplot}

\end{tikzpicture}
\caption{\textbf{Effect of using JSE with and without PCA for the Waterbirds dataset}:  We plot the (worst-group) accuracy on an OOD test set where $p_\mathrm{OOD}(\ymt = y  | \ysp = y ) = 0.5$, as a function of $p_\mathrm{train}(\ymt = y  | \ysp = y ).$ Each accuracy is obtained by averaging over 5 runs. The shaded area reflects the 95\% confidence interval. When using PCA, the dimension of the embeddings is reduced to $d=300$.  }
\label{fig:PCA}
\end{figure}

For our results in Section \ref{sec:OOD_generalization}, we first apply PCA to the embeddings to reduce their respective dimension. This section serves to show that it is possible for JSE to be successful without first applying PCA to the embeddings. We compare the performance of JSE on the Waterbirds dataset with and without PCA in Figure \ref{fig:PCA}, and observe that there is little to no difference in performance. However, the version using PCA is computationally more efficient due to the reduced dimensionality (2048 for the original last-layer embeddings of Resnet, vs. 300 after PCA). Aplying PCA to the embeddings can potentially be useful in cases where there is a limited set of datapoints available, and the dimensionality of the embeddings is very large.

\newpage
\section{Details on the Testing Procedure of JSE}
\label{sec:details_tests}

In this section, we provide a detailed explanation of the testing procedure used in the JSE algorithm for breaking the for-loops.  Section \ref{sec:test_notation} will  introduce notation, and a general set-up that is applicable for each test. In Section \ref{sec:derivation} we provide the  derivations that are needed for our subsequent test statistics. In Section \ref{sec:test_stat} we define the test statistics that are used throughout the JSE method. In Section \ref{sec:weighting_ablation} we study the effect of our choice to equally weight across the four groups in the data when measuring the test statistics, and in Section \ref{sec:Delta} we investigate how one can adjust the test in cases where one set of labels is much harder to predict than the other.

\subsection{Notation and Set-up}
\label{sec:test_notation}

Recall from Section \ref{sec:testing} that we are interested in testing the difference between two binary cross-entropies (BCE). For example, we can define the difference in BCE between a logistic regression that is trained on our embeddings $\boldsymbol{z}$ to predict the spurious concept labels $\ysp$ and one that just uses an intercept (referred to as `random' classifier).

For a particular sample, one could simply estimate this difference by weighting the individual observations equally. However, for our tests, we noticed that there was a considerable improvement if the difference was weighted equally within the subgroups specified by the pair of labels $(y_{\mathrm{mt}}, y_{\mathrm{sp}})$. This is examined in Section \ref{sec:weighting_ablation}. Below, we formulate how one, in general, can test for such a weighted difference between BCE's of two classifiers. 

Based on the main task and spurious labels $y_{\mathrm{mt}}, y_{\mathrm{sp}}$, we define four groups: (1) $y_{\mathrm{mt}} = 0,y_{\mathrm{sp}} = 0 $, (2) $y_{\mathrm{mt}} = 0,y_{\mathrm{sp}} = 1 $, (3) $y_{\mathrm{mt}} = 1,y_{\mathrm{sp}} = 0 $, and (4) $y_{\mathrm{mt}} = 1,y_{\mathrm{sp}} = 1 $. The spurious and main-task labels are used to define a new random variable $G \in \{1, 2, 3, 4 \}$, indicating group membership. The probability of a group $g$ is noted as $\pi_g$.

We consider the difference between the binary cross-entropy (BCE) as a random variable $d$. We assume that this difference is a mixture of four random variables: $d_1,d_2, d_3,d_4 $. The random variable $d_g$ corresponds to the difference in BCE for group $g$.  We assume that the expectation and variance of each of these four random variables is different, e.g. $\mathbb{E}[d_g] \neq \mathbb{E}[d_{h}]$,  $\mathrm{Var}[d_g] \neq \mathrm{Var}[d_{h}]$, for $g \neq h$. However, we do assume that they are independent: $\mathrm{Cov}(d_g, d_{h}) = 0$. We use the use the notation $\mathbb{E}[d_g] = \mu_g$ and $\mathrm{Var}[d_g] = \sigma^2_g$. 

We are interested in the weighted average of $d_1,d_2, d_3,d_4 $, where each group receives an equal weight. Concretely, we can define a new random variable 
\begin{align*}
    d_w = \frac{1}{4}\sum^{4}_{g=1}d_g.
\end{align*}
The subscript of $w$ will be used to refer to the equally weighted sum of difference $d$. We are interested in the following hypotheses 
\begin{align*}
    H_0: \mathbb{E}[d_w] = 0, \quad H_1: \mathbb{E}[d_w] < 0. 
\end{align*}
We draw a sample of $n$ independent and identical  (IID) observations from $d$ and $G$. 

Because we observe $G$, we know for each observation  which random variable we are observing - e.g. if $G = 1$, we observe $d_1$. This means that after drawing the $n$ observations, we observe the $n_g$ observations for group $g$. The observations of $d$ from group $g$ are denoted as $d_{g, i}$ for $i = 1, 2, ..., n_g$. However, we do not know the value of each $\pi_g$. We do assume that each $\pi_g$ is strictly positive. 

\subsection{Derivation of Test Statistic}
\label{sec:derivation}
The expectation of $d_w$ is
\begin{align*}
    \mu_w=\mathbb{E}[d_w ] &= \frac{1}{4}\mathbb{E}\left[\sum^{4}_{g=1}d_g \right]
    =  \frac{1}{4}\sum^{4}_{g=1}\mu_g.
\end{align*}
Let $\bar{d}_w$ denote an estimator of $d_w$
\begin{align*}
  \overallestimator = \overallestimatorsum\text{,} \quad \text{with}  \quad \bar{d}_g = \samplemeang{g},
\end{align*}
where $n_g$ is a random variable. We can show this estimator is unbiased via the law of total expectation.
\begin{align*}
    \mathbb{E}[\bar{d}_g] &= \mathbb{E}\left[\frac{1}{n_g}\sum^{n_g}_{i=1}d_{i, g}\right]\\
    &=  \mathbb{E}\left[\mathbb{E}\left[\frac{1}{n_g}\sum^{n_g}_{i=1}d_{i, g} | n_g\right]\right]\\
    &= \mathbb{E}\left[\frac{1}{n_g}\sum^{n_g}_{i=1} \mathbb{E}\left[d_{i, g} | n_g\right]\right]\\
    &= \mathbb{E}\left[\frac{1}{n_g}n_g \mu_g \right] \tag{Since they are IID}\\
    &=\mu_g, \\
    \mathbb{E}[\bar{d}_{w}] &= \frac{1}{4}\sum^{4}_{g=1} \mu_g.
\end{align*}
We now turn to the variance of $\bar{d}_{w}$
\begin{align*}
    \mathrm{Var}(\bar{d}_{w}) &=  \mathrm{Var}\left(\frac{1}{4}\sum^4_{g=1}\bar{d}_g\right) \\
    &= \frac{1}{16}\mathrm{Var}\left(\sum^4_{g=1}\bar{d}_g\right) \\
    &=  \frac{1}{16} \sum^{4}_{g=1}\sum^{4}_{h=1} \mathrm{Cov}(\bar{d}_g, \bar{d}_h ).
\end{align*}
First, it is shown first show that $\mathrm{Cov}(\bar{d}_g, \bar{d}_h ) = 0$ for $g \neq h$ via the law of total covariance:
\begin{align*}
    \mathrm{Cov}(\bar{d}_g, \bar{d}_h ) = \mathbb{E}[\mathrm{Cov}(\bar{d}_g, \bar{d}_h | n_g, n_h)] + \mathrm{Cov}(\mathbb{E}[\bar{d}_g | n_g, n_h], \mathbb{E}[\bar{d}_h| n_g, n_h])
\end{align*}
Since the sample means in expectation are the constants $\mu_g$, $\mu_h$, their covariance is 0:
\begin{align*}
    \mathrm{Cov}(\mathbb{E}[\bar{d}_g | n_g, n_h], \mathbb{E}[\bar{d}_h | n_g, n_h]) =  \mathrm{Cov}(\mu_g, \mu_h) = 0.
\end{align*}
Next, we define
\begin{align*}
    \mathbb{E}[\mathrm{Cov}(\bar{d}_g, \bar{d}_h | n_g, n_h)] &=  \mathbb{E}\left[\frac{1}{n_g}\frac{1}{n_h}\mathrm{Cov}\left(\sum^{n_g}_{i=1}d_{i, g}, \sum^{n_h}_{i=1}d_{i, h}  | n_g, n_h\right)\right]\\
    &=  \mathbb{E}\left[\frac{1}{n_g}\frac{1}{n_h}\times 0\right] = 0,
\end{align*}
This last step can be made because we assume  $d_{g,i}$  is independent of $d_{h,j}$ for $g \neq h$, $i = 1, ..., n_g$ and $j = 1, ..., n_h$. The variance of  $\overallestimator$ becomes
\begin{align*}
    \var{\overallestimator} = \frac{1}{16}\sum^{4}_{g=1}\var{\bar{d}_g}.
\end{align*}
We can define $\var{\bar{d}_g}$ via the law of total variance
\begin{align*}
    \var{\bar{d}_g} &= \mathrm{Var}\left(\frac{1}{n_g}\sum^{n_g}_{i=1}d_{i, g} \right)\\
    &= \mathbb{E}\left[\mathrm{Var}\left(\frac{1}{n_g}\sum^{n_g}_{i=1}d_{i, g} |  n_g\right)\right]+ \var{\mathbb{E}\left[\frac{1}{n_g}\sum^{n_g}_{i=1}d_{i, g}  | n_g \right]}\\
    &=\mathbb{E}\left[\mathrm{Var}\left(\frac{1}{n_g}\sum^{n_g}_{i=1}d_{i, g}  | n_g\right)\right] + \var{\mu_g}\\
    &= \mathbb{E}\left[\mathrm{Var}\left(\frac{1}{n_g}\sum^{n_g}_{i=1}d_{i, g}  | n_g\right)\right] \tag{Since variance of constant is 0 }\\
    &= \mathbb{E}\left[\frac{1}{n_g^2}\sum^{n_g}_{i=1}\var{d_{i, g}}\right]\\
    &=  \mathbb{E}[\frac{1}{n_g^2}n_g\sigma^2_g] \tag{Since they are IID}\\
    &= \mathbb{E}\left[\frac{1}{n_g}\right]\sigma^2_g.
\end{align*}
For $n \rightarrow \infty$, $n_g$ approximately follows a binomial distribution with $\mathbb{E}[n_g] = n\pi_g$ and variance $n\pi_g(1-\pi_g)$. We define a second order Taylor expansion to approximate $\frac{1}{n_g}$ around $\mathbb{E}[n_g] = n\pi_g$
\begin{align*}
    \mathbb{E}\left[\frac{1}{n_g}\right] \approx \frac{1}{n\pi_g} + \frac{(1-\pi_g)}{n^2\pi_g^2}.
\end{align*}
This means:
\begin{align*}
     \var{\bar{d}_g} \approx& \left(\frac{1}{n\pi_g} + \frac{(1-\pi_g)}{n^2\pi_g^2}\right)\sigma^2_g \\
     &= \frac{1}{n\pi_g}\sigma^2_g + \mathcal{O}(n^{-2}).
\end{align*}
Using this, we approximate the variance of the weighted sum $d_w$ via
\begin{align*}
    \mathrm{Var}(\overallestimator) \approx \frac{1}{16}\sum^{4}_{g=1} \frac{1}{n\pi_g}\sigma^2_g.
\end{align*}

Using the expectation and variance of $\overallestimator$, we now proceed to its asymptotic distribution. Assuming that (i) the sample means $\bar{d}_g$ are based on IID random variables and (ii)  $\sigma^2_g$ is bounded, we can use the central limit theorem (CLT) for each of the four sample means 
\begin{align*}
    \sqrt{n}\left(\bar{d}_g - \mu_g\right) \xrightarrow{\littlespace d \littlespace} \mathcal{N}\left(0, \frac{\sigma^2_g}{\pi_g}\right). 
\end{align*}
 This holds because $lim_{n\to\infty}n_g/n = \pi_g>0$ for all $g$; see for instance \citet{renyi1957asymptotic}. Given that the CLT holds for each sample mean, joint convergence follows by independence of the sample means. Hence, the distribution of the linear combination directly follows 
\begin{align*}
    \sqrt{n}(\overallestimator - \mathbb{E}[d_w])=\frac{1}{4}\sum_{g=1}^{4} \sqrt{n} (\bar{d}_g-\mu_g) \xrightarrow{\littlespace d \littlespace} \mathcal{N}\left(0, \frac{1}{16}\sum^{4}_{g=1} \frac{\sigma^2_g}{\pi_g}\right). 
\end{align*}
Hence, the variance of $\overallestimator$
\begin{align*}
 \mathrm{Var}(\overallestimator) \approx \frac{1}{16}\sum^{4}_{g=1} \frac{\sigma^2_g}{n \pi_g}
\end{align*}
can be consistently estimated via
\begin{align*}
 \widehat{\mathrm{Var}}(\overallestimator) = \frac{1}{16}\sum^{4}_{g=1} \frac{s^2_g}{n \hat{\pi}_g}=\frac{1}{16}\sum^{4}_{g=1} \frac{s^2_g}{n_g},
\end{align*}
with $\hat{\pi}_g=n_g/n$, and $s^2_g = \frac{1}{n_g - 1}\sum^{n_g}_{i=1}(d_{g, i} - \bar{d}_g)^2$. Using this, we can define a test statistic $t_{w}$, which for $n \rightarrow \infty$
\begin{align*}
    t_{w} = \frac{\overallestimator - \mathbb{E}[d_w]}{\widehat{\mathrm{Var}}(\overallestimator)} \xrightarrow{\littlespace d \littlespace} \mathcal{N}(0, 1).
\end{align*}

\subsection{Test statistics for JSE}
\label{sec:test_stat}

In the previous section, we defined a test statistic for the equally weighted weighted average of $d$. Here, we will use this derivation for the test statistics in the inner and outer loop of JSE. 

We start with the first criterion, namely that the vector $\vsp$ ($\vmt$) is informative about the spurious label (main-task label). This criterion is operationalised as follows: the $\vsp$ should contain more information about $\ysp$ than a majority-rule `random classifier'.  The coefficients for a logistic regression can be written as a combination of a unit vector and a scalar: $\boldsymbol{w} = \boldsymbol{v} \gamma$. Consider a logistic regression model $\yhatsp^{(\vsp)} = \mathrm{Logit}^{-1}\left(\gamma_\mathrm{sp} \bz^{\top} \vsp + b_\mathrm{sp}\right).$ This is a predictor for the label $\ysp$ based on the embeddings projected onto $\vsp$. Let $\yhatsp^{(\mathrm{rnd})}$ denote a random classifier. We can define the difference between these two classifiers as
\begin{equation*}
    d_{\mathrm{sp}, \mathrm{rnd}}^{(\vsp)} = \mathcal{L}_{\mathrm{BCE}}(\yhatsp^{(\vsp)}, \ysp) - \mathcal{L}_{\mathrm{BCE}}(\yhatsp^{(\mathrm{rnd})}, \ysp).
\end{equation*}
The first criterion translates into the following hypothesis 
\begin{equation*}
\label{eq:hypotheses_1}
    H_0: \mathbb{E}[d_{w, \mathrm{sp}, \mathrm{rnd}}^{(\vsp)}] = 0  \quad \text{versus} \quad
    H_1:  \mathbb{E}[d_{w, \mathrm{sp}, \mathrm{rnd}}^{(\vsp)}] < 0,
\end{equation*}
which means that under the null hypothesis, there is no difference in the BCE of these two classifiers. Under the alternative hypothesis, the BCE of a random classifier is higher. 

We use the following test statistic, where under the null hypothesis
\begin{align*}
\label{eq:test_stat_1}
    t_{\mathrm{sp}, \mathrm{rnd}} = \frac{\bar{d}_{w, \mathrm{sp}, \mathrm{rnd}}^{(\vsp)}}{\mathrm{Var}\left(\bar{d}_{w, \mathrm{sp}, \mathrm{rnd}}^{(\vsp)}\right)} \xrightarrow{\littlespace d \littlespace} \mathcal{N}(0, 1). 
\end{align*}

The previous test can also be defined for a main-task vector $\vmt$ and main-task labels $\ymt$. Consider a  logistic regression model $\yhatmt^{(\vmt)} = \mathrm{Logit}^{-1}\left(\gamma_\mathrm{mt} \bz^{\top} \vmt + b_\mathrm{mt}\right)$. The difference between the BCE of this classifier and a random classifier $\yhatmt^{(\mathrm{rnd})}$ is
\begin{equation*}
    d_{\mathrm{mt}, \mathrm{rnd}}^{(\vmt)} = \mathcal{L}_{\mathrm{BCE}}(\yhatmt^{(\vmt)}, \ymt) - \mathcal{L}_{\mathrm{BCE}}(\yhatmt^{(\mathrm{rnd})}, \ymt),
\end{equation*}
which we can use to test the hypothesis
\begin{equation*}
    H_0: \mathbb{E}[d_{w, \mathrm{mt}, \mathrm{rnd}}^{(\vmt)}] = 0  \quad \text{versus} \quad
    H_1:  \mathbb{E}[d_{w, \mathrm{mt}, \mathrm{rnd}}^{(\vmt)}] < 0.
\end{equation*}
For these hypotheses, we define a test statistic similar to the previous one, only now for the main-task vector and labels
\begin{align*}
    t_{\mathrm{mt}, \mathrm{rnd}} = \frac{\bar{d}_{w, \mathrm{mt}, \mathrm{rnd}}^{(\vmt)}}{\mathrm{Var}\left(\bar{d}_{w, \mathrm{mt}, \mathrm{rnd}}^{(\vmt)}\right)} \xrightarrow{\littlespace d \littlespace} \mathcal{N}(0, 1).
\end{align*}

We now turn to the second criterion, which is that the $\vsp$ is more predictive of the spurious concept than the main-task concept (and vice-versa for $\vmt$). This criterion is operationalised as follows: The BCE of a spurious vector $\vsp$ should be lower for the spurious concept than the main-task, and the vice-versa. We compare the BCE's of $\yhatsp^{(\vsp)}$ and $\yhatmt^{(\vsp)} = \mathrm{Logit}^{-1}\left(\gamma_\mathrm{mt}' \,\bz^{\top} \vsp + b_\mathrm{mt}'\right),$ where the latter is a predictor for the main-task label, based on the embeddings projected onto $\vsp$. The model parameters $\gamma_\mathrm{mt}$ and $b_\mathrm{mt}$ are to be trained by minimizing the BCE. We define the difference
\begin{equation*}
    d_{\mathrm{sp}, \mathrm{mt}}^{(\vsp)} = \mathcal{L}_{\mathrm{BCE}}(\yhatsp^{(\vsp)}, \ysp) - \mathcal{L}_{\mathrm{BCE}}(\yhatmt^{(\vsp)}, \ymt), 
\end{equation*}
and use this difference to test the following hypotheses
\begin{equation*}
\label{test:compare_sp}
    H_0: \mathbb{E}[d_{w, \mathrm{sp}, \mathrm{mt}}^{(\vsp)}] = \Delta  \quad \text{versus} \quad
    H_1:  \mathbb{E}[d_{w, \mathrm{sp}, \mathrm{mt}}^{(\vsp)}] < \Delta,
\end{equation*}
where the parameter $\Delta$ can be used to adjust for the fact that one label is harder to predict than the other. We give an example of its usefulness in Appendix~\ref{sec:Delta}, and a heuristic for setting the parameter value. Under the null hypothesis, the difference in the BCE for the spurious concept and main-task labels,
for a logistic regression  based on the embeddings projected onto $\vsp$, is $\Delta$. We can use the following test statistic, where under the null hypothesis
\begin{align*}
      t^{(\vsp)}_{\mathrm{sp}, \mathrm{mt}} = \frac{\bar{d}_{w, \mathrm{sp}, \mathrm{mt} }^{(\vsp)}- \Delta}{\mathrm{Var}\left(\bar{d}_{w, \mathrm{sp}, \mathrm{mt}}^{(\vsp)}\right)} \xrightarrow{\littlespace d \littlespace} \mathcal{N}(0, 1).
\end{align*}
We can then conduct the same test, but now for the main-task vector $\vmt$ instead of $\vsp$. Define $\yhatsp^{(\vmt)} = \mathrm{Logit}^{-1}\left(\gamma_\mathrm{sp}' \,\bz^{\top} \vmt + b_\mathrm{sp}'\right)$. This test uses the following difference 
\begin{align*}
    d_{\mathrm{sp}, \mathrm{mt}}^{(\vmt)} = \mathcal{L}_{\mathrm{BCE}}(\yhatsp^{(\vmt)}, \ysp) - \mathcal{L}_{\mathrm{BCE}}(\yhatmt^{(\vmt)}, \ymt),
\end{align*}
and the hypotheses become 
\begin{equation*}
\label{test:compare_mt}
    H_0: \mathbb{E}[d_{w, \mathrm{sp}, \mathrm{mt}}^{(\vmt)}] = \Delta  \quad \text{versus} \quad
    H_1:  \mathbb{E}[d_{w, \mathrm{sp}, \mathrm{mt}}^{(\vmt)}] > \Delta.
\end{equation*}
Under $H_1$ we now test if the difference is greater than $\Delta$ compared to the previous test. We use the following test statistic:
\begin{align*}
    t^{(\vmt)}_{\mathrm{sp}, \mathrm{mt}} = \frac{\bar{d}_{w, \mathrm{sp}, \mathrm{mt} }^{(\vmt)}- \Delta}{\mathrm{Var}\left(\bar{d}_{w, \mathrm{sp}, \mathrm{mt}}^{(\vmt)}\right)}  \xrightarrow{\littlespace d \littlespace} \mathcal{N}(0, 1).
\end{align*}

For each of the test statistics, given a large enough sample size, we can acquire our critical values from the standard normal distribution for a given significance level $\alpha$. 

In general, we used the validation set for the test statistics in order to mitigate the effect of overfitting. Unless otherwise mentioned, we set $\alpha = 0.05$ and $\Delta = 0$.  

\subsection{Weighted Average vs. Unweighted Average for Test Statistics of JSE }
\label{sec:weighting_ablation}

As stated in the previous section,  we use a weighted average for the test statistics of JSE, where the difference in BCE's is weighted equally across the four combinations of $\ysp$ and $\ymt$. We argue that this is helpful in distinguishing whether or not a spurious concept vector $\vsp$ contains a spurious concept or not (vice versa for $\vmt$). 

For example, consider a sample where 90\% of the water or landbirds coincide with a water or land background ($p_{\mathrm{train}}(\ymt = y | \ysp = y) = 0.9$), and we are interested in determining if a vector $\boldsymbol{v}$ contains information about the spurious or main-task features. If $\boldsymbol{v}$ contains information about the background features, a logistic regression for the label $\ysp$ based on the embeddings projected onto $\boldsymbol{v}$  will have a low BCE for spurious concept labels. However, this logistic regression will likely also have a low BCE for the main-task labels, due to the correlation between the labels. We can address this problem by weighting the BCE equally across the four combinations of $\ysp$ and $\ymt$, where the groups with a small sample (and where $\ysp$ and $\ymt$ do not coincide) will have a greater influence on the overall average

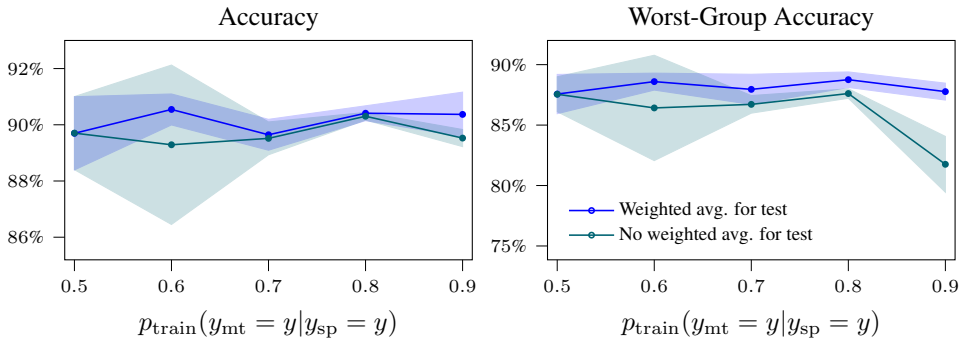
\begin{figure}[H]
\centering
\begin{tikzpicture}

\definecolor{darkgray176}{RGB}{176,176,176}
\definecolor{darkblue}{RGB}{0,100,115}
\definecolor{lightgray204}{RGB}{204,204,204}

\begin{groupplot}[group style={group size=2 by 1}]
\nextgroupplot[
height=\figureheight,
legend cell align={left},
legend style={
  fill opacity=0.8,
  draw opacity=1,
  text opacity=1,
  at={(0.03,0.03)},
  anchor=south west,
  draw=none,
  nodes={scale=0.8}
},
tick align=outside,
tick pos=left,
title={Accuracy},
width=\figurewidth,
x grid style={darkgray176},
xlabel={\(\displaystyle p_{\mathrm{train}}(\ymt = y | \ysp = y) \)},
xmin=0.49, xmax=0.91,
xtick style={color=black},
y grid style={darkgray176},
ymin=85.1953992310816, ymax=93,
ytick style={color=black},
yticklabel={\pgfmathprintnumber\tick\%},
title style={yshift=-0.2cm,}
]

\path [draw=blue, fill=blue, opacity=0.2]
(axis cs:0.5,90.9988598267963)
--(axis cs:0.5,88.393613870866)
--(axis cs:0.6,89.9943578978303)
--(axis cs:0.7,89.0949223701444)
--(axis cs:0.8,90.1464335490002)
--(axis cs:0.9,89.5778144807606)
--(axis cs:0.9,91.1608825712414)
--(axis cs:0.9,91.1608825712414)
--(axis cs:0.8,90.6751044224964)
--(axis cs:0.7,90.193996601871)
--(axis cs:0.6,91.0964261750456)
--(axis cs:0.5,90.9988598267963)
--cycle;

\path [draw=darkblue, fill=darkblue, opacity=0.2]
(axis cs:0.5,91.0030676247319)
--(axis cs:0.5,88.4032176612178)
--(axis cs:0.6,86.45319261162)
--(axis cs:0.7,88.932914910397)
--(axis cs:0.8,90.1678603745196)
--(axis cs:0.9,89.2231347957265)
--(axis cs:0.9,89.8310610851634)
--(axis cs:0.9,89.8310610851634)
--(axis cs:0.8,90.439665927818)
--(axis cs:0.7,90.1005695485263)
--(axis cs:0.6,92.1246477165965)
--(axis cs:0.5,91.0030676247319)
--cycle;

\addplot [semithick, blue, mark=*, mark size=1, mark options={solid}]
table {%
0.5 89.6962368488312
0.6 90.545392036438
0.7 89.6444594860077
0.8 90.4107689857483
0.9 90.369348526001
};

\addplot [semithick, darkblue, mark=*, mark size=1, mark options={solid}]
table {%
0.5 89.7031426429749
0.6 89.2889201641083
0.7 89.5167422294617
0.8 90.3037631511688
0.9 89.5270979404449
};

\nextgroupplot[
height=\figureheight,
legend cell align={left},
legend style={
  at={(0.03,0.03)},
  anchor=south west,
  draw=none,
  fill opacity=0.0,
  draw opacity=1,
  text opacity=1,
  nodes={scale=0.8}
},
tick align=outside,
tick pos=left,
title={Worst-Group Accuracy},
width=\figurewidth,
x grid style={darkgray176},
xlabel={\(\displaystyle p_{\mathrm{train}}(\ymt = y | \ysp = y) \)},
xmin=0.49, xmax=0.91,
xtick style={color=black},
y grid style={darkgray176},
ymin=73.8540648310368, ymax=92.0,
ytick style={color=black},
yticklabel={\pgfmathprintnumber\tick\%},
title style={yshift=-0.2cm,}
]
\path [draw=blue, fill=blue, opacity=0.2]
(axis cs:0.5,89.1907119750977)
--(axis cs:0.5,85.9433212280273)
--(axis cs:0.6,87.8955078125)
--(axis cs:0.7,86.7073593139648)
--(axis cs:0.8,88.1130065917969)
--(axis cs:0.9,87.069221496582)
--(axis cs:0.9,88.4702072143555)
--(axis cs:0.9,88.4702072143555)
--(axis cs:0.8,89.3975219726562)
--(axis cs:0.7,89.2039413452148)
--(axis cs:0.6,89.2966003417969)
--(axis cs:0.5,89.1907119750977)
--cycle;

\path [draw=darkblue, fill=darkblue, opacity=0.2]
(axis cs:0.5,88.9951705932617)
--(axis cs:0.5,86.1051864624023)
--(axis cs:0.6,82.0615158081055)
--(axis cs:0.7,85.9859313964844)
--(axis cs:0.8,87.2185287475586)
--(axis cs:0.9,79.4475326538086)
--(axis cs:0.9,84.0636672973633)
--(axis cs:0.9,84.0636672973633)
--(axis cs:0.8,88.0009841918945)
--(axis cs:0.7,87.4420013427734)
--(axis cs:0.6,90.7769393920898)
--(axis cs:0.5,88.9951705932617)
--cycle;

\addplot [semithick, blue, mark=*, mark size=1, mark options={solid}]
table {%
0.5 87.5670166015625
0.6 88.5960540771484
0.7 87.9556503295898
0.8 88.7552642822266
0.9 87.7697143554688
};
\addlegendentry{Weighted avg. for test}

\addplot [semithick, darkblue, mark=*, mark size=1, mark options={solid}]
table {%
0.5 87.550178527832
0.6 86.4192276000977
0.7 86.7139663696289
0.8 87.6097564697266
0.9 81.7555999755859
};
\addlegendentry{No weighted avg. for test}

\end{groupplot}

\end{tikzpicture}
\vspace{-0.25cm}
\caption{\textbf{Effect of using an equally weighted average for the tests of JSE for the Waterbirds dataset}:  We plot the (worst-group) accuracy on an OOD test set where $p_\mathrm{OOD}(\ymt = y  | \ysp = y ) = 0.5$, as a function of $p_\mathrm{train}(\ymt = y  | \ysp = y ).$ Each accuracy is obtained by averaging over 5 runs. The shaded area reflects the 95\% confidence interval.}
\label{fig:weighted_average_waterbirds}
\end{figure}

We verify this argument empirically by comparing two versions of JSE: one with the equally weighted average for the test statistics (as presented in this paper), and one which uses a simple average. Figure \ref{fig:weighted_average_waterbirds} shows the results of applying these two versions of the JSE algorithm to the Waterbirds dataset. When the spurious correlation is high, the version with a simple average performs worse in terms of both overall and worst-group accuracy.

\subsection{Adjusting for Different Difficulty in Predicting Labels}
\label{sec:Delta}

A potential issue with comparing two BCE's is that it might be fundamentally harder to predict one set of labels over the other. Consider the example of distinguishing cows vs. penguins as given in the introduction. If it is  much easier to predict the background than the animal type, then we might wrongly attribute spurious vectors to $\Zmt$. If the spurious concept is always easier to predict, then even if $\boldsymbol{v}$ represents the main-task features (e.g.~the animal shape), the vector might be attributed to the spurious concept subspace, since it has a lower binary cross-entropy for the spurious concept label than the main task label. To address this, we add the $\Delta$ term when testing for the second criterion mentioned in Section \ref{sec:testing}. By having a non-zero $\Delta$, we can account for the fact that one binary cross-entropy is always likely to be lower (or higher) than the other.

This naturally leads to the question how one should determine $\Delta$. We provide a simple heuristic. First, we optimize \Eqref{eq:estimation} and obtain a first pair of spurious and main-task vectors, $\vhatsp, \vhatmt$.
We compare the difference between these two orthogonal vectors through measuring the following:
\begin{align*}
    d^*_{\mathrm{sp}, \mathrm{mt}} = \mathcal{L}_{\mathrm{BCE}}(\yhatsp^{(\vhatsp)}, \ysp) - \mathcal{L}_{\mathrm{BCE}}(\yhatmt^{\vhatmt}, \ymt).
\end{align*}
We measure the weighted average of this term, defined $ \bar{d}^*_{w, \mathrm{sp}, \mathrm{mt}}$, for the validation set. This gives an   indication if one set of labels is harder to predict than the other, and its value can be used to set $\Delta$.  In order to demonstrate the usefulness of $\Delta$ and the heuristic, consider the Toy dataset, outlined Section~\ref{sec:datasets}. In the original set-up, both sets of labels were equally hard to predict, since $\gamma_{\mathrm{sp}} = \gamma_{\mathrm{mt}}= 3$ for both. We change that for this section, and set $\gamma_{\mathrm{sp}} = 6$, $ \gamma_{\mathrm{mt}} = 2$, making the spurious concept labels much more separable than the main-task labels.

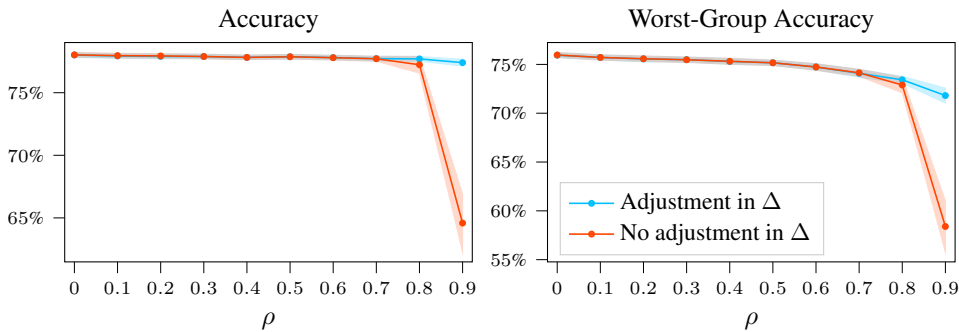
\begin{figure}[H]
\centering
\begin{tikzpicture}

\definecolor{darkgray176}{RGB}{176,176,176}
\definecolor{deepskyblue}{RGB}{0,191,255}
\definecolor{lightgray204}{RGB}{204,204,204}
\definecolor{orangered}{RGB}{255,69,0}

\begin{groupplot}[group style={group size=2 by 1}]
\nextgroupplot[
height=\figureheight,
legend cell align={left},
legend style={
  fill opacity=0.8,
  draw opacity=1,
  text opacity=1,
  at={(0.03,0.03)},
  anchor=south west,
  draw=lightgray204
},
tick align=outside,
tick pos=left,
title={ Accuracy},
width=\figurewidth,
x grid style={darkgray176},
xlabel={\(\displaystyle \rho\)},
xmin=-0.0225, xmax=0.9225,
xtick style={color=black},
y grid style={darkgray176},
ymin=61.4862704357358, ymax=79.0089848987297,
ytick style={color=black},
xtick={0, 0.1, 0.2, 0.3, 0.4, 0.5, 0.6, 0.7, 0.8, 0.9},
title style={yshift=-0.2cm},
yticklabel={\pgfmathprintnumber\tick\%}
]
\path [draw=deepskyblue, fill=deepskyblue, opacity=0.2]
(axis cs:0,78.1972004901097)
--(axis cs:0,77.821799086459)
--(axis cs:0.1,77.7468823857444)
--(axis cs:0.2,77.7077292564759)
--(axis cs:0.3,77.7055377610267)
--(axis cs:0.4,77.646530520798)
--(axis cs:0.5,77.6917387018858)
--(axis cs:0.6,77.6203653265858)
--(axis cs:0.7,77.5287387301111)
--(axis cs:0.8,77.5121552496836)
--(axis cs:0.9,77.130285115699)
--(axis cs:0.9,77.6757149977883)
--(axis cs:0.9,77.6757149977883)
--(axis cs:0.8,77.8888442963674)
--(axis cs:0.7,77.9162623952246)
--(axis cs:0.6,77.9936345170116)
--(axis cs:0.5,78.0562610854448)
--(axis cs:0.4,78.0034692169416)
--(axis cs:0.3,78.0754620186746)
--(axis cs:0.2,78.0902703639618)
--(axis cs:0.1,78.1231180958611)
--(axis cs:0,78.1972004901097)
--cycle;

\path [draw=orangered, fill=orangered, opacity=0.2]
(axis cs:0,78.2124978776845)
--(axis cs:0,77.8375018362132)
--(axis cs:0.1,77.7755277376503)
--(axis cs:0.2,77.754799617884)
--(axis cs:0.3,77.7065148822081)
--(axis cs:0.4,77.6416473567963)
--(axis cs:0.5,77.6895640210018)
--(axis cs:0.6,77.6095369057247)
--(axis cs:0.7,77.5161327480562)
--(axis cs:0.8,76.5635680037939)
--(axis cs:0.9,62.282757456781)
--(axis cs:0.9,66.8882425470337)
--(axis cs:0.9,66.8882425470337)
--(axis cs:0.8,77.901431314329)
--(axis cs:0.7,77.908868396353)
--(axis cs:0.6,77.9834629102162)
--(axis cs:0.5,78.0544358178272)
--(axis cs:0.4,77.9983522712707)
--(axis cs:0.3,78.0764848717439)
--(axis cs:0.2,78.1262000664498)
--(axis cs:0.1,78.1474727172524)
--(axis cs:0,78.2124978776845)
--cycle;

\addplot [semithick, deepskyblue, mark=*, mark size=1, mark options={solid}]
table {%
0 78.0094997882843
0.1 77.9350002408028
0.2 77.8989998102188
0.3 77.8904998898506
0.4 77.8249998688698
0.5 77.8739998936653
0.6 77.8069999217987
0.7 77.7225005626678
0.8 77.7004997730255
0.9 77.4030000567436
};
\addplot [semithick, orangered, mark=*, mark size=1, mark options={solid}]
table {%
0 78.0249998569489
0.1 77.9615002274513
0.2 77.9404998421669
0.3 77.891499876976
0.4 77.8199998140335
0.5 77.8719999194145
0.6 77.7964999079704
0.7 77.7125005722046
0.8 77.2324996590614
0.9 64.5855000019073
};

\nextgroupplot[
height=\figureheight,
legend cell align={left},
legend style={
  fill opacity=0.8,
  draw opacity=1,
  text opacity=1,
  at={(0.03,0.03)},
  anchor=south west,
  draw=lightgray204
},
tick align=outside,
tick pos=left,
title={Worst-Group Accuracy},
width=\figurewidth,
x grid style={darkgray176},
xlabel={\(\displaystyle \rho\)},
xmin=-0.0225, xmax=0.9225,
xtick style={color=black},
y grid style={darkgray176},
ymin=54.7721283050576, ymax=77.2405906588637,
ytick style={color=black},
xtick={0, 0.1, 0.2, 0.3, 0.4, 0.5, 0.6, 0.7, 0.8, 0.9},
title style={yshift=-0.2cm},
yticklabel={\pgfmathprintnumber\tick\%}
]
\path [draw=deepskyblue, fill=deepskyblue, opacity=0.2]
(axis cs:0,76.1987402453656)
--(axis cs:0,75.6960198864703)
--(axis cs:0.1,75.4408051870163)
--(axis cs:0.2,75.2698032203127)
--(axis cs:0.3,75.2144660638914)
--(axis cs:0.4,75.0057580245912)
--(axis cs:0.5,74.8862108787698)
--(axis cs:0.6,74.4082748509379)
--(axis cs:0.7,73.6846005488471)
--(axis cs:0.8,73.0588905558411)
--(axis cs:0.9,71.0379832875909)
--(axis cs:0.9,72.5836445200263)
--(axis cs:0.9,72.5836445200263)
--(axis cs:0.8,73.7842796101745)
--(axis cs:0.7,74.4971621464654)
--(axis cs:0.6,75.0694473170308)
--(axis cs:0.5,75.4426618018942)
--(axis cs:0.4,75.6113989578307)
--(axis cs:0.3,75.7604332281008)
--(axis cs:0.2,75.8539913353514)
--(axis cs:0.1,75.9877531625931)
--(axis cs:0,76.1987402453656)
--cycle;

\path [draw=orangered, fill=orangered, opacity=0.2]
(axis cs:0,76.2192969155089)
--(axis cs:0,75.7079796958192)
--(axis cs:0.1,75.4470674135743)
--(axis cs:0.2,75.2980795724067)
--(axis cs:0.3,75.2029100803502)
--(axis cs:0.4,75.0148732796399)
--(axis cs:0.5,74.8865545032522)
--(axis cs:0.6,74.4018985293733)
--(axis cs:0.7,73.7864953845685)
--(axis cs:0.8,72.0917982442408)
--(axis cs:0.9,55.7934220484124)
--(axis cs:0.9,61.001721079029)
--(axis cs:0.9,61.001721079029)
--(axis cs:0.8,73.6873613016576)
--(axis cs:0.7,74.494388404494)
--(axis cs:0.6,75.0631588436736)
--(axis cs:0.5,75.4429437877634)
--(axis cs:0.4,75.6205026969226)
--(axis cs:0.3,75.7471173854701)
--(axis cs:0.2,75.8649148123589)
--(axis cs:0.1,75.9747923276367)
--(axis cs:0,76.2192969155089)
--cycle;

\addplot [semithick, deepskyblue, mark=*, mark size=1, mark options={solid}]
table {%
0 75.947380065918
0.1 75.7142791748047
0.2 75.561897277832
0.3 75.4874496459961
0.4 75.3085784912109
0.5 75.164436340332
0.6 74.7388610839844
0.7 74.0908813476562
0.8 73.4215850830078
0.9 71.8108139038086
};
\addlegendentry{Adjustment in $\Delta$}
\addplot [semithick, orangered, mark=*, mark size=1, mark options={solid}]
table {%
0 75.9636383056641
0.1 75.7109298706055
0.2 75.5814971923828
0.3 75.4750137329102
0.4 75.3176879882812
0.5 75.1647491455078
0.6 74.7325286865234
0.7 74.1404418945312
0.8 72.8895797729492
0.9 58.3975715637207
};
\addlegendentry{No adjustment in $\Delta$}
\end{groupplot}

\end{tikzpicture}
\vspace{-0.25cm}
\caption{\textbf{Effect of adjusting $\Delta$ for JSE}: We plot the  (worst-group) accuracy on a test set without spurious correlation, as a function of the spurious correlation in the training data. Each accuracy is obtained by averaging over 100 runs. The shaded area reflects the 95\% confidence interval.} 
\label{fig:adjustment}
\end{figure}

In Figure \ref{fig:adjustment} we show how JSE performs on a Toy dataset where the separability of the spurious and main-task labels differs. If we do not adjust $\Delta$ (e.g. keep it 0), we observe that overall and worst-group accuracy drop once the correlation between the spurious and main-task features becomes high ($\rho = 0.9$). If we adjust $\Delta$ via our heuristic, this problem is avoided.

\section{Grad-CAM Images for Waterbirds dataset}
\label{sec:gradcam_fig}

\begin{figure}[H]
\vspace{-0.2cm}
    \centering
\includegraphics[scale=0.15]{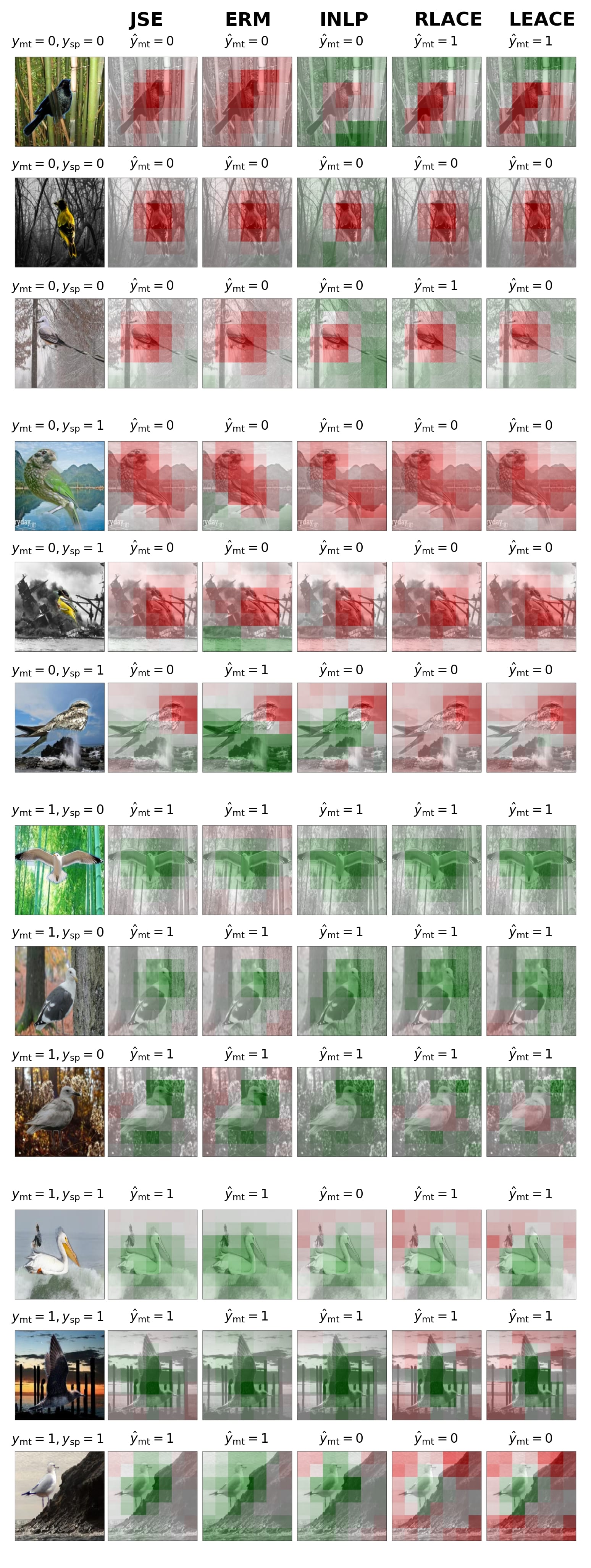}
    \vspace{-0.4cm}
    \caption{\textbf{Grad-CAM images for the Waterbirds dataset}: Red (green) patches indicate contribution
towards a prediction \\ $\ymt=0$ ($\ymt=1$). The
model and respective concept-removal methods were trained on
data where $p_{\mathrm{train}}(\ymt = y|\ysp = y) = 0.9$.}
    \label{fig:gradcam_WB_all}

\end{figure}

 For each method, we start with the same model, finetuned on the Waterbirds dataset. Then, after applying PCA ($d=300$), concept removal, and re-training the last linear layer, we apply Grad-CAM to the last convolutional layer of the ResNet50 model.  The model and concept-removal methods uses data were $p_\mathrm{train}(\ymt = y  | \ysp = y ) = 0.9$. We use the \texttt{Captum} library for calculating the Grad-CAM saliency maps.  From each of the four groups in the data (defined by the combinations of $\ymt, \ysp$), we sample 3 images at random. The results are shown in Figure \ref{fig:gradcam_WB_all}. For each group (landbirds on land, landbirds on water, waterbirds on land, waterbirds on water) we observe that JSE ignores the background, while other methods rely on the background for prediction.

\section{Concept Removal of Black and White CelebA Images}
\label{sec:CelebA_removal}

\begin{figure}[H]
    \centering
    \includegraphics[scale=0.05]{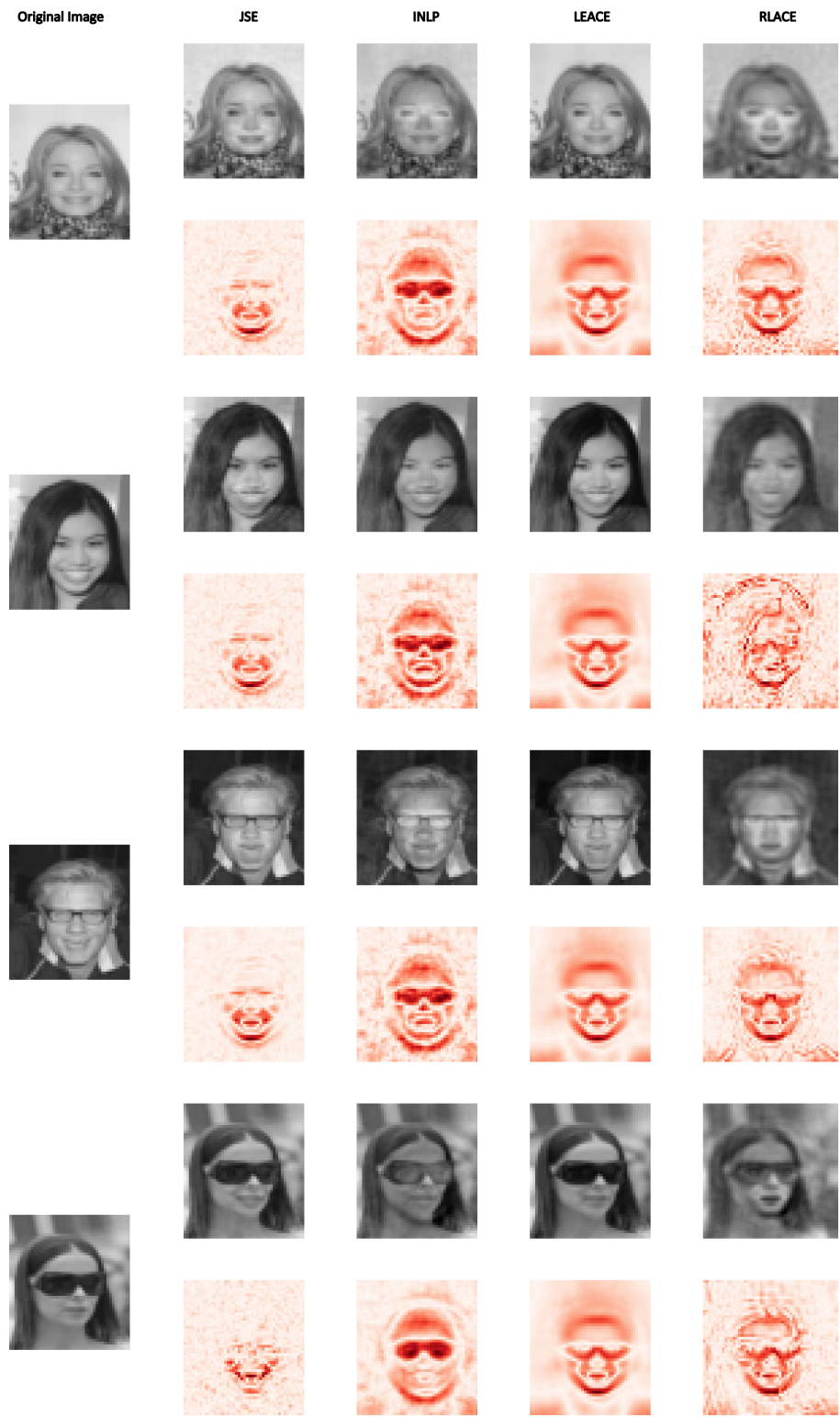}
    \caption{\textbf{Application of concept-removal methods to raw pixel data (image 1-4)}:  The first row shows the image, after it is transformed by the concept-removal method. The second row shows the absolute difference between the transformed and original image to indicate which pixels have been changed.}
        \label{fig:CelebA_removal_1}

\end{figure}

\begin{figure}[H]
    \centering
    \includegraphics[scale=0.05]{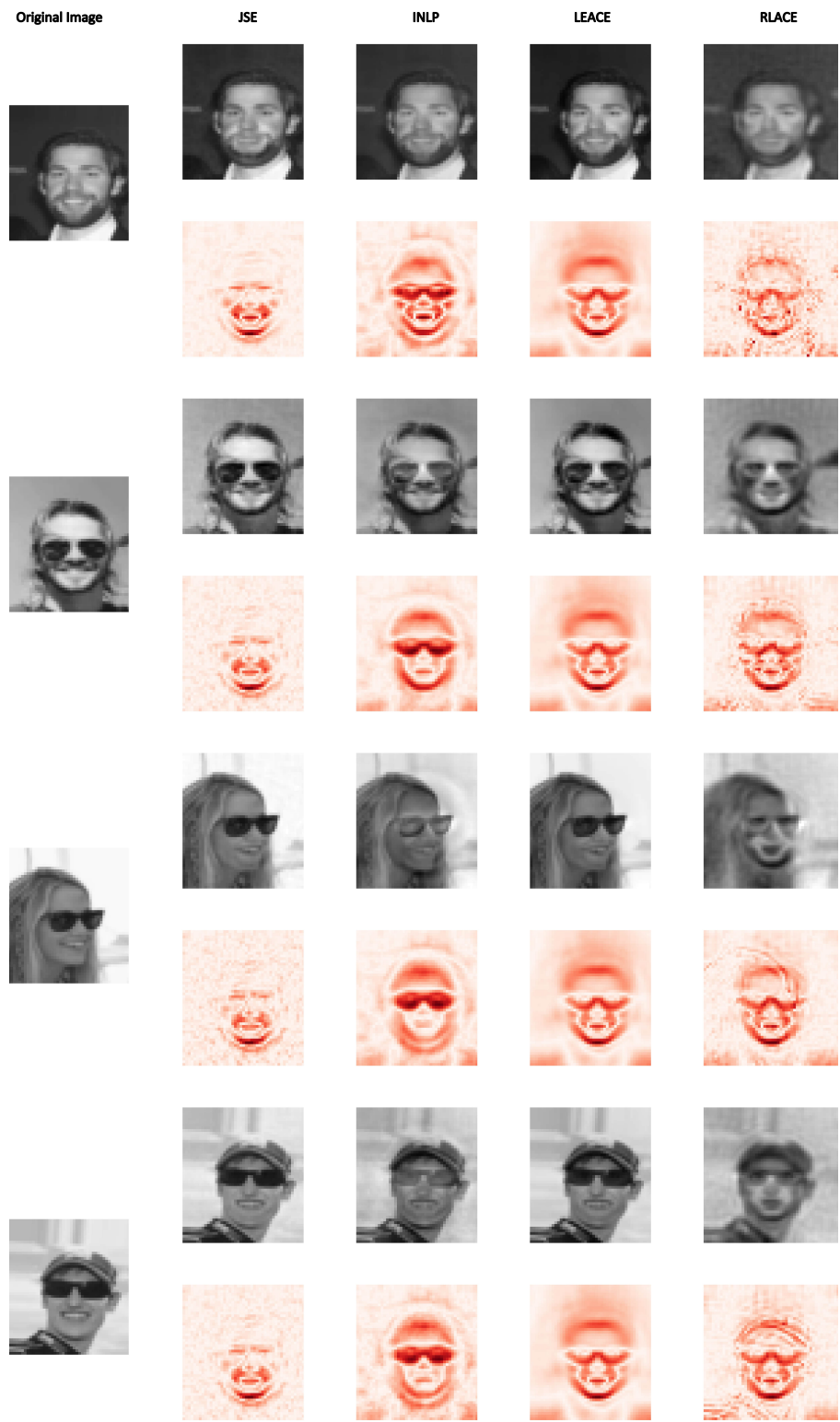}
    \caption{\textbf{Application of concept-removal methods to raw pixel data (image 5-8)}:  The first row shows the image, after it is transformed by the concept-removal method. The second row shows the absolute difference between the transformed and original image to indicate which pixels have been changed.}
        \label{fig:CelebA_removal_2}
\end{figure}

For each method, we work with 8,000 images in the training set, and 2,000 in the validation set. The black and white images are sampled such that $p(\ymt = y) = 0.5$. We set $p(\ymt = y | \ysp = y) = 0.8$ to create a spurious correlation between the main-task features and spurious concept features. We run JSE, INLP and RLACE with a learning rate of 0.01, weight decay 0f 0.01, and a batch size of 128, and optimize via the outlined training procedure in Section \ref{sec:training_procedure}.   When using the original dimension $(2500)$, we were unable to get RLACE to converge. For this method, we used PCA to the reduce the dimensionality of the images ($d=500$) before applying the projection. The vectors are then transformed back to the original dimension. The results are shown for several images across Figure \ref{fig:CelebA_removal_1} and \ref{fig:CelebA_removal_2}.

\section{Comparing JSE to other Instance-reweighting Methods}
\label{sec:OOD_other_algorithms}

In this section, we compare JSE to four powerful instance-reweighting methods: (i) group-weighted empirical risk minimization (GW-ERM), which uses a sampling scheme such that the spurious concept and main-task labels are balanced \mbox{\citep{idrissi2022simple}}, (ii)  group distributional robust optimization (GDRO), which aims to minimize the worst-group loss over possible combinations of the spurious and main-task labels \mbox{\citep{sagawa_2020}}, (iii) sub-group resampling (SUBG), in which we create a new sample of the dataset where each group is equal in size to the smallest group, and them apply ERM \citep{sagawa_2020B}, and finally (iv) just train twice (JTT), which puts a greater weight on samples that were wrongly identified by an initial model \mbox{\citep{Liu_2021}}. This last method does not require the use of spurious concept labels, except for the validation set.

Concept-removal methods such as JSE, but also INLP, RLACE and LEACE are applied post-hoc, e.g. when the full model has already been trained. For a fair comparison in terms of computational resources, each method is applied to the last layer of a model that was finetuned on data with the spurious correlation. Previous work finds that this strategy is particular effective at dealing with spurious correlations  \mbox{\citep{kirichenko_2022}}. The results of this comparison with the on benchmark datasets is discussed in Section \ref{sec:instance_reweighting_full}, and in Section \ref{sec:instance_reweighting_limited} we compare the methods when limited spurious concept labels are available.

\setlength\figureheight{5cm}
\setlength\figurewidth{5cm}

\newpage

\subsection{Comparison on Benchmark Datasets}
\label{sec:instance_reweighting_full}

\begin{figure}[H]
\centering
\vspace{-0.25cm}
\input{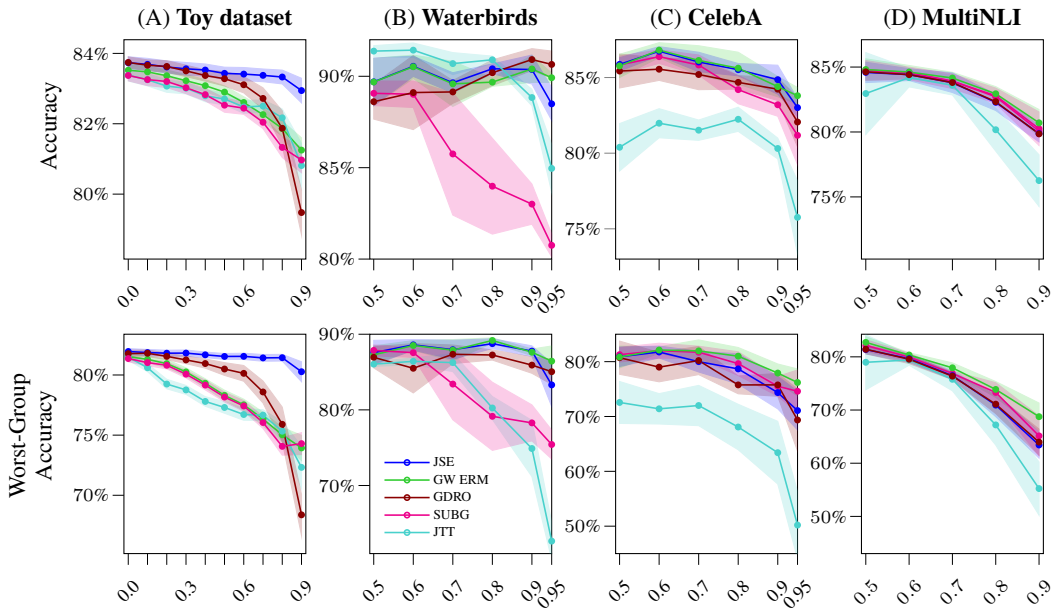}
\caption{\textbf{OOD generalization, compared to instance-reweighting methods}: We plot the (worst-group) accuracy on a test set without spurious correlation, as a function of the spurious correlation in the training set ($\rho$ for the Toy dataset, $p_\mathrm{train}(\ymt = y  | \ysp = y )$ for the other datasets). Averages based on 100, 5, 5 and 5 runs, respectively. The shaded area reflects the 95\% confidence interval.}
\label{fig:OOD_generalization_instance_reweighting}
\vspace{-0.25cm}
\end{figure}

We show the comparison between JSE and instance-reweighting methods with the datasets considered in this paper (Toy, Waterbirds, CelebA, multiNLI) in Figure \ref{fig:OOD_generalization_instance_reweighting}. The numerical results are shown in Table \ref{tab:WB_instance_reweighting_table}-\ref{tab:toy_instance_reweighting_table_2}. JSE is comparable or outperforms other instance-reweighting methods, except in the case where the spurious correlation is very high - e.g. $p(\ymt = y | \ysp = y) = 0.95$ for the Waterbirds and CelebA datasets. \\

We suspect the relatively worse performance of JSE at $p(\ymt = y | \ysp = y) = 0.95$ is related to finite-sample estimation noise. If the spurious correlation strength is high, there are relatively less datapoints where the spurious vector is not predictive of the main-task, and vice-versa - making it harder for JSE to separate the spurious and main-task features. We further describe this issue in Section \ref{sec:finite_sample_noise} of the appendix using the Toy dataset.\\ 

Furthermore, we emphasize a conceptual difference between JSE and instance-reweighting methods. JSE removes the spurious embedding, which makes it robust to changes at the more fundamental level of the conditional distribution $p(\ymt | \zsp).$ The other methods aim for robustness against changes in $p(\ymt | \ysp).$ This explains why for the Toy dataset JSE outperforms other methods, because in this setting  the conditional distribution $p(\ymt | \zsp)$ is modified for the OOD data.

\subsection{Comparison with Limited Spurious Concept Labels}
\label{sec:instance_reweighting_limited}

In this section, we evaluate the ability of JSE and instance-reweighting methods to deal with spurious correlations when limited spurious concept labels are available. We compare methods that rely on such labels for training (and thus leave out JTT). In the experiment, there is a limited budget of size $n_{\mathrm{sp}}$ to gather spurious concept labels - e.g. one can label 1500, 1000, or 500 images with $\ysp$. The limited dataset (of size $n_{\mathrm{sp}}$) exhibits the same spurious correlation as in the original training set. In the case of GW-ERM, GDRO, this means we can only train on a limited set of labels. For SUBG, this means we still create a dataset where each group has the size of the smallest group - only now based on the smaller dataset. For JSE, we first estimate the spurious concept subspace based on the limited dataset. We use this estimate to remove the spurious concept subspace from all the embeddings, and subsequently train the model to predict $\ymt$ on the full dataset.  \\

\begin{figure}[H]
\centering
\input{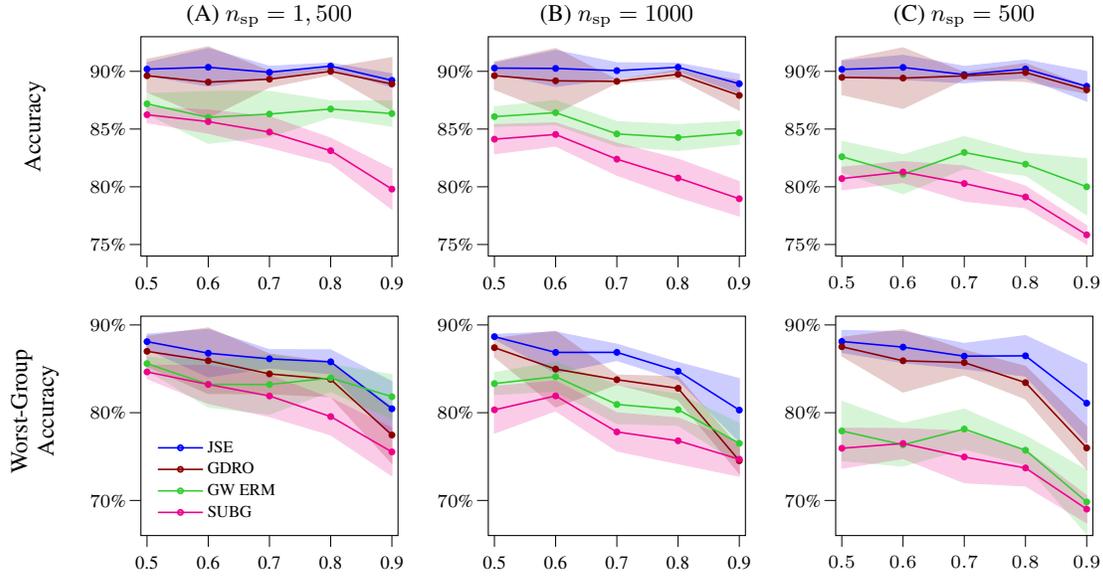}
\caption{\textbf{Performance with limited available spurious concept labels for the Waterbirds dataset}:  We plot the (worst-group) accuracy on an OOD test set where $p_\mathrm{OOD}(\ymt = y  | \ysp = y ) = 0.5$, as a function of $p_\mathrm{train}(\ymt = y  | \ysp = y ).$ Each accuracy is obtained by averaging over 5 runs. The shaded area reflects the 95\% confidence interval.}
\label{fig:sample_sizes_WB}
\end{figure} 
\vspace{-0.3cm}
\begin{figure}[H]
\centering
\input{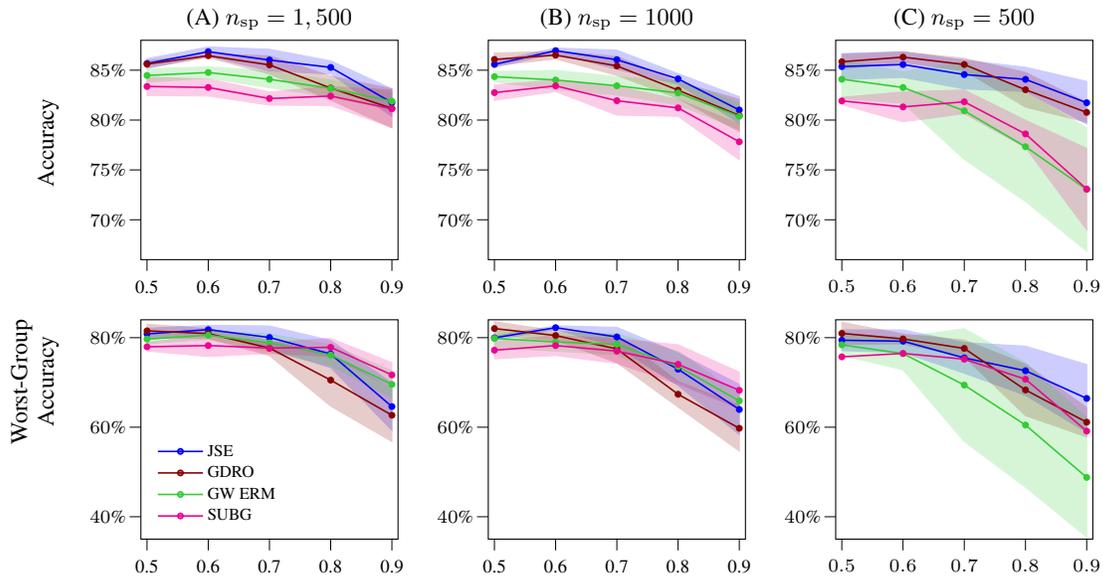}
\caption{\textbf{Performance with limited available spurious concept labels for the CelebA dataset}:  We plot the (worst-group) accuracy on an OOD test set where $p_\mathrm{OOD}(\ymt = y  | \ysp = y ) = 0.5$, as a function of $p_\mathrm{train}(\ymt = y  | \ysp = y ).$ Each accuracy is obtained by averaging over 5 runs. The shaded area reflects the 95\% confidence interval.}
\label{fig:sample_sizes_CelebA}
\end{figure} 

\vspace{-0.3cm}
In Figures  \ref{fig:sample_sizes_WB} and \ref{fig:sample_sizes_CelebA} we compare JSE to instance-reweighting methods on the Waterbirds and CelebA datasets with varying numbers of avilable spurious concept labels. For the Waterbirds dataset, JSE outperforms other methods in overall accuracy. In terms of worst-group accuracy, it is on average slightly better than GW-ERM  for $n_{\mathrm{sp}} = 1000$ and $n_{\mathrm{sp}} = 500$. For the CelebA dataset, JSE performs similar or slightly better than GDRO. In terms of worst-group accuracy,  GW-ERM and SUBG perform on average better than JSE and GDRO for $n_{\mathrm{sp}} = 1500$ and $n_{\mathrm{sp}} = 1000$, although the variance of the worst-group accuracy is quite high.  When  $n_{\mathrm{sp}} = 500$, JSE on average outperforms GW-ERM and SUBG in terms of overall and worst-group accuracy. Overall, the results indicate that JSE can have an advantage when limited spurious concept labels are available.


\begin{table}[H]
\caption{\textbf{Results of instance-reweighting methods for the Waterbirds dataset}: Table shows the average, worst-group, and per-group accuracy on a test set where $p_{\mathrm{OOD}}(\ymt = y | \ysp = y) = 0.5$, with $y \in \{0, 1\}$,  as a function of $p_\mathrm{train}(\ymt = y  | \ysp = y ).$ Each accuracy is obtained by averaging over 5 runs. Standard error is reported between brackets. }
\label{tab:WB_instance_reweighting_table}
\centering \tablesize
\vskip 0.15in

\begin{tabular}{clllllll}
\toprule
\multicolumn{1}{l}{\textbf{}}       & \textbf{}              & \multicolumn{5}{c}{$p_{\mathrm{train}}(\ymt = y | \ysp = y)$}              \\
\multicolumn{1}{l}{\textbf{Method}} & \textbf{Accuracy}      & 0.5          & 0.6          & 0.7          & 0.8          & 0.9   & 0.95       \\
\hline
\multirow{6}{*}{JSE}   & $\ymt = 0$, $\ysp = 0$ & 90.72 (0.83) & 92.43 (0.44) & 90.58 (0.27) & 91.46 (0.76) & 91.85 (0.75)  & 91.08 (0.64)  \\
                       & $\ymt = 0$, $\ysp = 1$ & 88.16 (1.08) & 89.05 (0.50) & 87.96 (0.64) & 89.64 (0.50) & 89.56 (0.76) &  83.99 (1.71) \\
                       & $\ymt = 1$, $\ysp = 0$  & 91.43 (0.24) & 90.16 (0.46) & 91.25 (0.28) & 90.25 (0.41) & 89.63 (0.81) & 84.98 (2.09) \\
                       & $\ymt = 1$, $\ysp = 1$  & 89.75 (0.57) & 89.56 (0.33) & 90.69 (0.32) & 89.60 (0.53) & 88.75 (0.66) & 90.44 (0.80) \\
                       & Worst-group            & 87.57 (0.83) & 88.60 (0.36) & 87.96 (0.64) & 88.76 (0.33) & 87.77 (0.36) & 83.30 (1.46) \\
                       & Average                & 89.70 (0.66) & 90.55 (0.28) & 89.64 (0.28) & 90.41 (0.13) & 90.37 (0.40) & 88.49 (0.46)\\
                       \hline
                       
\multirow{6}{*}{GW-ERM} & $\ymt = 0$, $\ysp = 0$ & 90.16 (0.71) & 91.95 (0.50) & 89.35 (1.28) & 89.42 (0.28) & 91.29 (0.77) & 92.57 (0.71) \\
                        & $\ymt = 0$, $\ysp = 1$ & 88.79 (1.21) & 89.34 (0.58) & 89.16 (0.66) & 89.61 (0.11) & 90.42 (0.66) & 88.57 (0.33) \\
                        & $\ymt = 1$, $\ysp = 0$  & 91.71 (0.29) & 90.65 (0.33) & 91.87 (0.57) & 90.93 (0.10) & 89.69 (0.56) & 87.04 (1.30) \\
                        & $\ymt = 1$, $\ysp = 1$  & 89.13 (0.63) & 89.53 (0.57) & 89.63 (0.16) & 89.60 (0.16) & 87.91 (0.48) & 88.26 (0.24) \\
                        & Worst-group            & 87.27 (0.60) & 88.51 (0.33) & 87.89 (0.72) & 89.18 (0.18) & 87.62 (0.22) & 86.44 (1.02) \\
                        & Average                & 89.69 (0.66) & 90.52 (0.33) & 89.59 (0.63) & 89.68 (0.12) & 90.40 (0.38) & 89.92 (0.13) \\
                        \hline
\multirow{6}{*}{SUBG} & $\ymt = 0$, $\ysp = 0$ & 88.86 (0.54) & 89.58 (0.55) & 84.06 (2.29) & 79.42 (2.53) & 78.28 (1.22) & 76.76 (1.40) \\
                      & $\ymt = 0$, $\ysp = 1$ & 88.87 (0.57) & 88.20 (0.56) & 84.24 (2.52) & 84.39 (1.37) & 83.48 (0.59) & 79.28 (1.30) \\
                      & $\ymt = 1$, $\ysp = 0$ & 90.81 (0.39) & 90.16 (0.50) & 92.74 (0.57) & 92.49 (0.67) & 92.12 (0.45) & 91.68 (0.26) \\
                      & $\ymt = 1$, $\ysp = 1$ & 88.75 (0.33) & 88.79 (0.55) & 90.00 (0.96) & 90.09 (0.97) & 88.85 (0.46) & 89.00 (0.97) \\
                      & Worst-group            & 87.86 (0.37) & 87.57 (0.32) & 83.41 (2.41) & 79.15 (2.30) & 78.28 (1.22) & 75.42 (0.98) \\
                      & Average                & 89.07 (0.34) & 89.02 (0.37) & 85.75 (1.71) & 83.99 (1.34) & 83.01 (0.56) & 80.75 (0.32)\\
\hline
\multirow{6}{*}{GDRO}   & $\ymt = 0$, $\ysp = 0$ & 88.20 (1.11) & 89.35 (2.64) & 89.60 (0.70) & 92.46 (0.75) & 94.02 (0.61) & 93.53 (0.99) \\
                        & $\ymt = 0$, $\ysp = 1$ & 87.73 (0.56) & 88.31 (1.01) & 87.95 (0.85) & 88.44 (1.01) & 90.28 (0.82) & 90.30 (0.52)\\
                        & $\ymt = 1$, $\ysp = 0$  & 91.65 (0.26) & 90.75 (0.69) & 91.28 (0.41) & 88.75 (0.70) & 86.17 (0.68) & 85.39 (1.17) \\
                        & $\ymt = 1$, $\ysp = 1$  & 90.12 (0.42) & 89.44 (0.69) & 89.63 (0.70) & 89.81 (0.82) & 86.98 (0.65) & 87.01 (0.67)\\
                        & Worst-group            & 86.94 (0.71) & 85.50 (1.66) & 87.35 (0.62) & 87.26 (0.34) & 85.92 (0.55) & 85.05 (0.85)\\
                        & Average                & 88.61 (0.49) & 89.11 (1.03) & 89.14 (0.32) & 90.19 (0.36) & 90.91 (0.30) & 90.65 (0.37)\\
                        \hline
\multirow{6}{*}{JTT}    & $\ymt = 0$, $\ysp = 0$ & 92.63 (0.37) & 95.14 (0.37) & 94.29 (0.38) & 97.08 (0.21) & 98.16 (0.26) & 98.93 (0.12) \\
                        & $\ymt = 0$, $\ysp = 1$ & 92.53 (0.22) & 90.12 (0.58) & 88.52 (0.72) & 87.96 (0.29) & 82.94 (0.90) & 75.34 (1.98)\\
                        & $\ymt = 1$, $\ysp = 0$  & 88.26 (0.31) & 86.54 (0.30) & 86.64 (0.74) & 80.25 (0.77) & 74.89 (1.89) & 62.65 (1.26) \\
                        & $\ymt = 1$, $\ysp = 1$  & 86.04 (0.14) & 87.88 (0.44) & 89.75 (0.32) & 90.16 (0.46) & 90.75 (0.47) & 91.99 (0.57)\\
                        & Worst-group            & 86.04 (0.14) & 86.42 (0.23) & 86.25 (0.61) & 80.25 (0.77) & 74.89 (1.89) & 62.65 (1.26) \\
                        & Average                & 91.38 (0.15) & 91.43 (0.20) & 90.69 (0.31) & 90.90 (0.10) & 88.84 (0.23) & 84.96 (0.67)\\
                                    \bottomrule
\end{tabular}
\end{table}

\begin{table}[H]
\caption{\textbf{Results instance-reweighting methods for the CelebA dataset}: Table shows the average, worst-group, and per-group accuracy on a test set where $p_{\mathrm{OOD}}(\ymt = y | \ysp = y) = 0.5$, with $y \in \{0, 1\}$, as a function of $p_\mathrm{train}(\ymt = y  | \ysp = y )$. Each accuracy is obtained by averaging over 5 runs. Standard error is reported between brackets. }
\vskip 0.15in

\centering \tablesize
\begin{tabular}{clllllll}
\toprule
\multicolumn{1}{l}{\textbf{}}       & \textbf{}              & \multicolumn{5}{c}{$p_{\mathrm{train}}(\ymt = y | \ysp = y)$}              \\
\multicolumn{1}{l}{\textbf{Method}} & \textbf{Accuracy}      & 0.5          & 0.6          & 0.7          & 0.8          & 0.9 & 0.95          \\
\hline
\multirow{6}{*}{JSE}   & $\ymt = 0$, $\ysp = 0$ & 86.40 (0.35) & 87.56 (0.71) & 87.52 (0.36) & 88.32 (1.57) & 90.36 (0.43) & 91.08 (0.64)  \\
                       & $\ymt = 0$, $\ysp = 1$ & 82.72 (1.23) & 82.24 (0.56) & 81.12 (1.18) & 82.56 (0.77) & 81.84 (1.37) & 78.72 (0.92) \\
                       & $\ymt = 1$, $\ysp = 0$ & 82.04 (0.70) & 84.48 (1.23) & 82.72 (1.10) & 79.12 (1.74) & 74.40 (1.55) & 71.08 (1.80) \\
                       & $\ymt = 1$, $\ysp = 1$ & 92.32 (0.37) & 92.68 (0.30) & 92.76 (0.23) & 92.08 (0.57) & 92.84 (0.37) & 91.16 (0.53)\\
                       & Worst-group            & 80.88 (0.90) & 81.76 (0.56) & 80.00 (0.96) & 78.68 (1.41) & 74.36 (1.51) & 71.08 (1.80) \\
                       & Average                & 85.87 (0.23) & 86.74 (0.15) & 86.03 (0.33) & 85.52 (0.21) & 84.86 (0.49) & 83.01 (0.30)\\
                       \hline
\multirow{6}{*}{GW-ERM} & $\ymt = 0$, $\ysp = 0$ & 86.63 (0.87) & 87.64 (0.50) & 85.88 (0.88) & 86.72 (0.98) & 86.64 (0.62) & 88.08 (0.55) \\
                        & $\ymt = 0$, $\ysp = 1$ & 83.05 (1.15) & 82.72 (0.67) & 82.68 (1.39) & 83.08 (0.76) & 81.36 (1.46) & 78.36 (0.67)\\
                        & $\ymt = 1$, $\ysp = 0$  & 82.66 (1.52) & 84.88 (1.20) & 84.80 (1.08) & 81.00 (0.81) & 78.56 (0.84) & 76.76 (1.50)\\
                        & $\ymt = 1$, $\ysp = 1$  & 92.08 (0.46) & 92.00 (0.40) & 91.16 (0.63) & 91.68 (0.42) & 91.04 (0.75) & 92.00 (1.10) \\
                        & Worst-group            & 80.85 (1.01) & 82.12 (0.30) & 81.96 (1.02) & 81.00 (0.81) & 77.92 (0.85) & 76.20 (1.26)\\
                        & Average                & 85.77 (0.39) & 86.81 (0.25) & 86.13 (0.50) & 85.62 (0.54) & 84.40 (0.34) & 83.80 (0.28)\\
                        \hline
\multirow{6}{*}{SUBG} & $\ymt = 0$, $\ysp = 0$ & 86.72 (0.71) & 87.00 (0.67) & 85.60 (0.75) & 85.44 (0.61) & 86.64 (0.60) & 83.72 (0.61) \\
                      & $\ymt = 0$, $\ysp = 1$ & 82.92 (0.62) & 83.76 (0.62) & 82.84 (0.85) & 82.20 (0.59) & 80.96 (1.13) & 77.96 (0.56) \\
                      & $\ymt = 1$, $\ysp = 0$ & 81.76 (0.79) & 83.40 (1.51) & 84.04 (1.31) & 79.76 (1.18) & 75.52 (0.92) & 75.12 (2.23) \\
                      & $\ymt = 1$, $\ysp = 1$ & 91.92 (0.37) & 91.36 (0.74) & 90.84 (0.26) & 89.40 (0.64) & 89.68 (0.82) & 87.92 (0.95) \\
                      & Worst-group            & 81.28 (0.68) & 82.00 (0.70) & 81.72 (0.56) & 79.64 (1.10) & 75.52 (0.92) & 74.60 (1.98) \\
                      & Average                & 85.83 (0.30) & 86.38 (0.23) & 85.83 (0.34) & 84.20 (0.48) & 83.20 (0.39) & 81.18 (0.99) \\
                      \hline
\multirow{6}{*}{GDRO}   & $\ymt = 0$, $\ysp = 0$ & 85.16 (1.45) & 88.28 (0.67) & 85.68 (1.14) & 88.96 (0.80) & 88.12 (0.94) & 90.64 (1.20)\\
                        & $\ymt = 0$, $\ysp = 1$ & 82.88 (1.29) & 80.68 (0.34) & 80.84 (1.54) & 80.76 (0.70) & 80.68 (1.31) & 73.76 (2.07)\\
                        & $\ymt = 1$, $\ysp = 0$  & 82.56 (1.62) & 80.36 (2.00) & 82.20 (1.80) & 75.76 (0.83) & 75.76 (1.06) & 69.40 (2.91) \\
                        & $\ymt = 1$, $\ysp = 1$  & 91.12 (0.27) & 92.88 (0.61) & 92.08 (0.95) & 93.24 (0.49) & 92.36 (0.92) & 94.44 (0.79)\\
                        & Worst-group            & 80.72 (1.54) & 79.00 (1.37) & 80.20 (1.31) & 75.76 (0.83) & 75.76 (1.06) & 69.36 (2.88)\\
                        & Average                & 85.43 (0.57) & 85.55 (0.38) & 85.20 (0.51) & 84.68 (0.23) & 84.23 (0.38) & 82.06 (0.84)\\
                        \hline
\multirow{6}{*}{JTT}    & $\ymt = 0$, $\ysp = 0$ & 82.16 (1.64) & 85.64 (1.09) & 85.48 (1.27) & 89.20 (0.49) & 91.04 (0.50) & 94.68 (0.63)\\
                        & $\ymt = 0$, $\ysp = 1$ & 83.04 (0.74) & 83.80 (1.06) & 79.08 (1.11) & 79.92 (0.71) & 71.72 (0.85) & 63.04 (2.13)\\
                        & $\ymt = 1$, $\ysp = 0$  & 72.56 (1.95) & 71.40 (1.41) & 72.00 (1.87) & 68.08 (2.12) & 63.40 (2.92) & 50.20 (3.13) \\
                        & $\ymt = 1$, $\ysp = 1$  & 83.80 (2.30) & 87.08 (0.62) & 89.52 (1.29) & 91.76 (0.44) & 95.08 (0.77)& 95.12 (0.45)\\
                        & Worst-group            & 72.56 (1.95) & 71.40 (1.41) & 72.00 (1.87) & 68.08 (2.12) & 63.40 (2.92) & 50.20 (3.13)\\
                        & Average                & 80.39 (0.80) & 81.98 (0.49) & 81.52 (0.34) & 82.24 (0.41) & 80.31 (0.35) & 75.76 (1.23)\\
                                    \bottomrule
\end{tabular}
\end{table}

\begin{table}[H]
\caption{\textbf{Results of instance-reweighting methods for the MultiNLI dataset}: Table shows the average, worst-group, and per-group accuracy on a test set where $p_{\mathrm{OOD}}(\ymt = y | \ysp = y) = 0.5$, with $y \in \{0, 1\}$,  as a function of $p_\mathrm{train}(\ymt = y  | \ysp = y )$. Each accuracy is obtained by averaging over 5 runs. Standard error is reported between brackets. }
\vskip 0.15in
\centering \tablesize

\begin{tabular}{cllllll}
\toprule
\multicolumn{1}{l}{\textbf{}}       & \textbf{}              & \multicolumn{5}{c}{$p_{\mathrm{train}}(\ymt = y | \ysp = y)$}              \\
\multicolumn{1}{l}{\textbf{Method}} & \textbf{Accuracy}      & 0.5          & 0.6          & 0.7          & 0.8          & 0.9          \\
\hline
\multirow{6}{*}{GW-ERM}             & $\ymt = 0$, $\ysp = 0$ & 86.51 (0.50) & 88.69 (0.46) & 90.21 (0.38) & 91.34 (0.40) & 92.54 (0.54) \\
                                    & $\ymt = 0$, $\ysp = 1$ & 86.05 (0.51) & 83.55 (0.74) & 81.20 (0.64) & 77.87 (1.27) & 71.73 (1.87) \\
                                    & $\ymt = 1$, $\ysp = 0$  & 83.46 (0.68) & 80.35 (0.21) & 77.94 (0.47) & 73.87 (0.76) & 68.75 (1.31) \\
                                    & $\ymt = 1$, $\ysp = 1$  & 83.22 (0.43) & 85.87 (0.49) & 87.28 (0.38) & 88.69 (0.78) & 89.79 (0.93) \\
                                    & Worst-group            & 82.72 (0.54) & 80.35 (0.21) & 77.94 (0.47) & 73.87 (0.76) & 68.75 (1.31) \\
                                    & Average                & 84.81 (0.42) & 84.62 (0.26) & 84.16 (0.22) & 82.94 (0.28) & 80.70 (0.52) \\
                                    \hline
\multirow{6}{*}{SUBG} & $\ymt = 0$, $\ysp = 0$ & 86.94 (0.51) & 88.72 (0.50) & 90.54 (0.34) & 91.44 (0.78) & 93.60 (0.62) \\
                      & $\ymt = 0$, $\ysp = 1$ & 86.56 (0.30) & 83.52 (0.70) & 81.01 (0.85) & 78.61 (1.56) & 73.41 (1.54) \\
                      & $\ymt = 1$, $\ysp = 0$  & 82.70 (0.76) & 79.97 (0.36) & 76.88 (0.22) & 73.31 (0.93) & 65.14 (1.92) \\
                      & $\ymt = 1$, $\ysp = 1$  & 82.75 (0.39) & 85.84 (0.45) & 87.18 (0.46) & 87.81 (0.93) & 88.86 (0.87) \\
                      & Worst-group            & 82.13 (0.55) & 79.97 (0.36) & 76.88 (0.22) & 73.31 (0.93) & 65.14 (1.92) \\
                      & Average                & 84.74 (0.39) & 84.51 (0.24) & 83.90 (0.29) & 82.79 (0.30) & 80.25 (0.66) \\
\hline
\multirow{6}{*}{GDRO}               & $\ymt = 0$, $\ysp = 0$ & 87.47 (0.66) & 89.07 (0.64) & 90.75 (0.29) & 92.62 (0.34) & 94.05 (0.46) \\
                                    & $\ymt = 0$, $\ysp = 1$ & 87.50 (0.40) & 84.10 (0.70) & 81.17 (0.95) & 77.49 (1.20) & 71.70 (1.17) \\
                                    & $\ymt = 1$, $\ysp = 0$  & 81.84 (0.54) & 79.55 (0.41) & 76.37 (0.33) & 71.10 (0.90) & 63.97 (1.25) \\
                                    & $\ymt = 1$, $\ysp = 1$  & 81.71 (0.43) & 84.93 (0.62) & 86.94 (0.82) & 88.11 (0.60) & 89.70 (0.86) \\
                                    & Worst-group            & 81.42 (0.44) & 79.55 (0.41) & 76.37 (0.33) & 71.10 (0.90) & 63.97 (1.25) \\
                                    & Average                & 84.63 (0.42) & 84.41 (0.28) & 83.81 (0.34) & 82.33 (0.32) & 79.85 (0.41) \\
                                    \hline
\multirow{6}{*}{JTT}                & $\ymt = 0$, $\ysp = 0$ & 83.52 (1.64) & 88.29 (0.83) & 90.99 (0.26) & 92.10 (2.08) & 96.11 (0.83) \\
                                    & $\ymt = 0$, $\ysp = 1$ & 82.42 (2.40) & 81.97 (1.03) & 78.05 (1.33) & 70.02 (2.20) & 57.92 (3.39) \\
                                    & $\ymt = 1$, $\ysp = 0$  & 82.82 (3.15) & 79.95 (0.48) & 76.00 (1.06) & 68.58 (2.23) & 56.14 (2.28) \\
                                    & $\ymt = 1$, $\ysp = 1$  & 83.09 (2.65) & 86.54 (0.66) & 89.34 (0.32) & 90.00 (2.81) & 94.85 (0.66) \\
                                    & Worst-group            & 78.98 (2.64) & 79.41 (0.50) & 75.78 (1.10) & 67.20 (1.93) & 55.23 (2.55) \\
                                    & Average                & 82.96 (1.61) & 84.19 (0.36) & 83.60 (0.38) & 80.17 (0.84) & 76.26 (1.01) \\
                                    \bottomrule
\end{tabular}
\end{table}

\begin{table}[H]
\caption{\textbf{Results of instance-reweighting methods for the Toy dataset for} $\rho \in \{0.0, 0.1, 0.2, 0.3, 0.4\}$.  Table shows the average, worst-group, and per-group accuracy on a test set without spurious correlation, as a function of the spurious correlation in the training data. Each accuracy is obtained by averaging over 100 runs. Standard error is reported between brackets. }
\centering \tablesize
\vskip 0.15in

\begin{tabular}{cllllll}
\toprule
\multicolumn{1}{l}{\textbf{}}       & \textbf{}              & \multicolumn{5}{c}{$\rho$}              \\
\multicolumn{1}{l}{\textbf{Method}} & \textbf{Accuracy}      & 0.0          & 0.1          & 0.2          & 0.3          & 0.4          \\
\hline
\multirow{6}{*}{GW-ERM}             & $\ymt = 0$ and $\ysp = 0$ & 83.26 (0.20) & 82.64 (0.19) & 82.15 (0.21) & 81.25 (0.22) & 80.44 (0.25) \\
                                    & $\ymt = 0$ and $\ysp = 1$ & 83.56 (0.17) & 83.96 (0.19) & 84.55 (0.18) & 84.69 (0.19) & 85.47 (0.18) \\
                                    & $\ymt = 1$ and $\ysp = 0$ & 83.79 (0.19) & 84.25 (0.20) & 84.56 (0.17) & 85.13 (0.19) & 85.57 (0.20) \\
                                    & $\ymt = 1$ and $\ysp = 1$ & 83.45 (0.17) & 83.05 (0.18) & 82.21 (0.19) & 81.82 (0.18) & 80.87 (0.21) \\
                                    & Worst-group                    & 81.52 (0.14) & 81.30 (0.14) & 80.95 (0.17) & 80.26 (0.17) & 79.31 (0.19) \\
                                    & Average                        & 83.52 (0.09) & 83.48 (0.08) & 83.36 (0.08) & 83.22 (0.09) & 83.08 (0.09) \\
                                    \hline
\multirow{6}{*}{SUBG} & $\ymt = 0$, $\ysp = 0$ & 83.54 (0.20) & 82.66 (0.20) & 82.21 (0.20) & 81.57 (0.20) & 80.64 (0.22) \\
                      & $\ymt = 0$, $\ysp = 1$ & 83.42 (0.20) & 83.80 (0.19) & 84.22 (0.20) & 84.86 (0.24) & 85.00 (0.22) \\
                      & $\ymt = 1$, $\ysp = 0$ & 83.11 (0.17) & 83.84 (0.18) & 84.21 (0.18) & 84.46 (0.21) & 85.17 (0.21) \\
                      & $\ymt = 1$, $\ysp = 1$ & 83.43 (0.17) & 82.73 (0.19) & 82.18 (0.21) & 81.25 (0.21) & 80.56 (0.26) \\
                      & Worst-group            & 81.37 (0.14) & 81.03 (0.13) & 80.81 (0.15) & 80.07 (0.16) & 79.16 (0.20) \\
                      & Average                & 83.37 (0.08) & 83.26 (0.09) & 83.20 (0.09) & 83.02 (0.09) & 82.83 (0.09)
                      \\
                      \hline
\multirow{6}{*}{GDRO}               & $\ymt = 0$ and $\ysp = 0$ & 83.63 (0.16) & 83.49 (0.15) & 83.57 (0.17) & 83.44 (0.20) & 83.45 (0.21) \\
                                    & $\ymt = 0$ and $\ysp = 1$ & 83.93 (0.18) & 83.78 (0.18) & 83.81 (0.20) & 83.49 (0.21) & 83.13 (0.25) \\
                                    & $\ymt = 1$ and $\ysp = 0$ & 83.54 (0.20) & 83.54 (0.20) & 83.42 (0.21) & 83.21 (0.22) & 82.91 (0.26) \\
                                    & $\ymt = 1$ and $\ysp = 1$ & 83.81 (0.19) & 83.80 (0.17) & 83.66 (0.20) & 83.82 (0.21) & 83.94 (0.22) \\
                                    & Worst-group                    & 81.78 (0.15) & 81.81 (0.14) & 81.55 (0.14) & 81.25 (0.16) & 80.95 (0.20) \\
                                    & Average                        & 83.74 (0.08) & 83.66 (0.08) & 83.63 (0.08) & 83.50 (0.08) & 83.37 (0.08) \\
                                    \hline
\multirow{6}{*}{JTT}                & $\ymt = 0$ and $\ysp = 0$ & 83.23 (0.18) & 81.81 (0.23) & 80.60 (0.23) & 80.07 (0.24) & 79.59 (0.30) \\
                                    & $\ymt = 0$ and $\ysp = 1$ & 83.71 (0.17) & 84.83 (0.18) & 85.80 (0.18) & 86.06 (0.19) & 86.22 (0.25) \\
                                    & $\ymt = 1$ and $\ysp = 0$ & 83.12 (0.18) & 84.34 (0.20) & 85.21 (0.17) & 85.64 (0.20) & 85.77 (0.25) \\
                                    & $\ymt = 1$ and $\ysp = 1$ & 83.43 (0.19) & 81.99 (0.17) & 80.66 (0.25) & 80.27 (0.25) & 79.60 (0.30) \\
                                    & Worst-group                    & 81.49 (0.13) & 80.60 (0.16) & 79.25 (0.20) & 78.76 (0.19) & 77.80 (0.20) \\
                                    & Average                        & 83.38 (0.08) & 83.25 (0.09) & 83.07 (0.09) & 83.01 (0.08) & 82.80 (0.09) \\
                                    \bottomrule
\end{tabular}

\end{table}

\begin{table}[H]
\caption{\textbf{Results of instance-reweighting methods for the Toy dataset for} $\rho \in \{0.5, 0.6, 0.7, 0.8, 0.9\}$.  Table shows the average, worst-group, and per-group accuracy on a test set without spurious correlation, as a function of the spurious correlation in the training data. Each accuracy is obtained by averaging over 100 runs. Standard error is reported between brackets. }
\label{tab:toy_instance_reweighting_table_2}

\centering \tablesize
\vskip 0.15in
\begin{tabular}{cllllll}
\toprule
\multicolumn{1}{l}{}                &                                & \multicolumn{5}{c}{$\rho$}                                               \\
\multicolumn{1}{l}{\textbf{Method}} & \textbf{Accuracy}              & 0.5          & 0.6          & 0.7          & 0.8          & 0.9          \\
\hline
\multirow{6}{*}{GW-ERM}             & $\ymt = 0$ and $\ysp = 0$ & 79.55 (0.30) & 78.88 (0.30) & 78.37 (0.41) & 77.57 (0.46) & 79.78 (0.57) \\
                                    & $\ymt = 0$ and $\ysp = 1$ & 85.65 (0.23) & 85.63 (0.27) & 85.73 (0.40) & 85.59 (0.50) & 82.78 (0.78) \\
                                    & $\ymt = 1$ and $\ysp = 0$ & 86.05 (0.22) & 86.19 (0.29) & 86.06 (0.35) & 85.99 (0.47) & 82.69 (0.80) \\
                                    & $\ymt = 1$ and $\ysp = 1$ & 80.33 (0.24) & 79.72 (0.33) & 78.90 (0.37) & 78.28 (0.45) & 79.68 (0.59) \\
                                    & Worst-group                    & 78.31 (0.22) & 77.52 (0.24) & 76.33 (0.29) & 75.03 (0.31) & 73.94 (0.53) \\
                                    & Average                        & 82.90 (0.09) & 82.60 (0.08) & 82.26 (0.10) & 81.86 (0.11) & 81.25 (0.18) \\
                                    \hline
\multirow{6}{*}{SUBG} & $\ymt = 0$, $\ysp = 0$ & 80.03 (0.28) & 78.79 (0.27) & 77.71 (0.31) & 77.63 (0.46) & 78.75 (0.48) \\
                      & $\ymt = 0$, $\ysp = 1$ & 85.19 (0.28) & 86.09 (0.24) & 86.44 (0.31) & 85.10 (0.62) & 83.39 (0.71) \\
                      & $\ymt = 1$, $\ysp = 0$ & 85.02 (0.28) & 86.15 (0.23) & 86.44 (0.25) & 85.29 (0.53) & 83.22 (0.68) \\
                      & $\ymt = 1$, $\ysp = 1$ & 79.92 (0.27) & 78.79 (0.29) & 77.68 (0.33) & 77.42 (0.49) & 78.57 (0.49) \\
                      & Worst-group            & 78.17 (0.20) & 77.42 (0.23) & 76.05 (0.25) & 74.07 (0.37) & 74.29 (0.46) \\
                      & Average                & 82.53 (0.10) & 82.44 (0.09) & 82.04 (0.09) & 81.33 (0.15) & 80.97 (0.18) \\
                      \hline
\multirow{6}{*}{GDRO}               & $\ymt = 0$ and $\ysp = 0$ & 83.64 (0.24) & 83.89 (0.23) & 84.62 (0.29) & 85.70 (0.30) & 88.46 (0.29) \\
                                    & $\ymt = 0$ and $\ysp = 1$ & 83.01 (0.34) & 82.25 (0.39) & 80.94 (0.50) & 77.92 (0.78) & 69.99 (0.90) \\
                                    & $\ymt = 1$ and $\ysp = 0$ & 82.51 (0.33) & 82.07 (0.36) & 80.51 (0.55) & 77.70 (0.76) & 70.26 (0.85) \\
                                    & $\ymt = 1$ and $\ysp = 1$ & 83.93 (0.25) & 84.20 (0.28) & 84.74 (0.31) & 86.10 (0.33) & 89.00 (0.28) \\
                                    & Worst-group                    & 80.50 (0.27) & 80.14 (0.33) & 78.59 (0.47) & 75.90 (0.73) & 68.37 (0.87) \\
                                    & Average                        & 83.28 (0.10) & 83.11 (0.12) & 82.72 (0.15) & 81.87 (0.25) & 79.48 (0.32) \\
                                    \hline
\multirow{6}{*}{JTT}                & $\ymt = 0$ and $\ysp = 0$ & 79.66 (0.33) & 78.52 (0.29) & 78.66 (0.34) & 79.67 (0.48) & 82.96 (0.56) \\
                                    & $\ymt = 0$ and $\ysp = 1$ & 85.97 (0.31) & 86.65 (0.29) & 86.70 (0.25) & 84.78 (0.57) & 78.26 (1.14) \\
                                    & $\ymt = 1$ and $\ysp = 0$ & 85.42 (0.36) & 86.13 (0.34) & 85.97 (0.30) & 84.26 (0.67) & 78.29 (1.11) \\
                                    & $\ymt = 1$ and $\ysp = 1$ & 79.71 (0.36) & 78.64 (0.32) & 78.72 (0.32) & 79.92 (0.46) & 83.64 (0.57) \\
                                    & Worst-group                    & 77.30 (0.25) & 76.71 (0.25) & 76.66 (0.23) & 75.33 (0.40) & 72.32 (0.90) \\
                                    & Average                        & 82.70 (0.09) & 82.48 (0.09) & 82.50 (0.09) & 82.17 (0.13) & 80.81 (0.32) \\
                                    \bottomrule
\end{tabular}
\end{table}

\section{JSE and Multiple Spurious Concepts}
\label{sec:multiple_concepts}

In this section we aim to provide a brief example of how one can use JSE to deal with multiple spurious concepts. In order to do so, we conduct an experiment on the CelebA dataset. Here, the first spurious concept $\ysp^{(1)} \in \{0, 1\}$ is the sex of the person in the image (female-male). We take the version of the dataset where the spurious correlation between the main-task and spurious concept of gender is $p(\ymt = y | \ysp^{(1)} = y) = 0.8$. In the test set, $p(\ymt = y | \ysp^{(1)} = y) = 0.5$. The second spurious concept $\ysp^{(2)} \in \{0, 1\}$ indicates whether or not someone wears glasses or not. This second spurious concept also happens to be spuriously correlated with the main-task of blond vs. non-blond. For instance, in this dataset, $p(\ymt = 1 | \ysp^{(2)} = 1) = 0.18$, and $p(\ymt = 0 | \ysp^{(2)} = 1) = 0.82$. 

To remove the influence of both concepts, we first apply JSE to estimate the subspace related to $\ysp^{(1)}$, and then related to $\ysp^{(2)}$. Then, we project the embeddings on the orthogonal complement of both subspaces. For determining the hyper-parameters of JSE in each instance, we use the same approach as our experiments on OOD generalization.  We compare this approach to using standard ERM, as well as group-weighted ERM (GW-ERM). With group-weighted ERM, the groups are defined over all 8 combinations of $\ymt, \ysp^{(1)}, \ysp^{(2)}$.\\

The results of this experiment are reported in Table \ref{tab:JSE_multiple}. If JSE is applied for both spurious concepts, it improves upon ERM for worst-group accuracy for both spurious concepts. There appears to be a small trade-off with only intervening for a single concept. For instance, if we only remove the subspace related to $\ysp^{(1)}$, the average worst-group accuracy (over combinations of $\ymt,\ysp^{(1)}$) is  80.04, vs. 78.72 if we intervene for both spurious concepts.


\begin{table}[H]
\caption{\textbf{Results of applying JSE to multiple spurious concepts:} Table shows the average worst-group, and per-group accuracy on a test set where $p(\ymt = y | \ysp^{(1)}= y) = 0.5$.  In the training set,  $p(\ymt = y | \ysp^{(1)}= y) = 0.8$. The worst-group accuracy is either taken over combinations of $\ymt $, $\ysp^{(1)} $ or  $\ymt $, $\ysp^{(2)} $. The $n$ refers to the number of samples in the test set. We show three versions of JSE: when removing the subspace related to $\ysp^{(1)}$, $\ysp^{(2)}$, and both. The learning rate used for JSE to remove $\ysp^{(1)}$ was 0.001, with a weight decay of 0.001, and for removing $\ysp^{(2)}$ it was 0.01, with a weight decay of 0.01.  Each accuracy is obtained by averaging over 5 runs. Standard error is reported in brackets.  }
\label{tab:JSE_multiple}
\vspace{0.25cm}
\begin{tabular}{lllllll}
\toprule
\textbf{Accuracy type}               & $n$ & \textbf{Both $\ysp^{(1)}$, $\ysp^{(2)}$} & \textbf{Only for  $\ysp^{(1)}$} & \textbf{Only for  $\ysp^{(2)}$} & \textbf{ERM} & \textbf{GW-ERM} \\
\hline
Overall                              & 2000       & 85.6 (0.2)                                               & 85.99 (0.24)                                                               & 84.02 (0.29)                                                               & 83.68 (0.22) & 84.45 (0.57)    \\
$\ymt = 1$, $\ysp^{(1)} =1$          & 500        & 92.16 (0.72)                                             & 92.2 (0.6)                                                                 & 95.84 (0.33)                                                               & 95.88 (0.45) & 89.64 (0.49)    \\
$\ymt = 1$, $\ysp^{(1)} =0$          & 500        & 78.72 (0.69)                                             & 80.24 (1.12)                                                               & 68.64 (0.7)                                                                & 67.8 (0.64)  & 80.0 (0.48)     \\
$\ymt = 0$, $\ysp^{(1)} =1$          & 500        & 82.64 (0.53)                                             & 83.08 (0.39)                                                               & 77.68 (0.87)                                                               & 76.96 (0.68) & 81.56 (0.73)    \\
$\ymt = 0$, $\ysp^{(1)} =0$          & 500        & 88.88 (0.99)                                             & 88.44 (0.92)                                                               & 93.92 (0.5)                                                                & 94.08 (0.59) & 86.6 (1.96)     \\
Worst-group ($\ymt $, $\ysp^{(1)} $) & -          & 78.72 (0.69)                                             & 80.04 (0.98)                                                               & 68.64 (0.7)                                                                & 67.8 (0.64)  & 79.36 (0.39)    \\
\hline
$\ymt = 1$, $\ysp^{(2)} =1$          & 30         & 73.11 (3.6)                                              & 73.5 (3.4)                                                                 & 75.04 (2.65)                                                               & 63.96 (2.34) & 83.5 (2.66)     \\
$\ymt = 1$, $\ysp^{(2)} =0$          & 970        & 86.0 (0.15)                                              & 86.81 (0.39)                                                               & 82.58 (0.43)                                                               & 82.64 (0.61) & 84.93 (0.33)    \\
$\ymt = 0$, $\ysp^{(2)} =1$          & 75         & 83.94 (4.65)                                             & 84.43 (2.96)                                                               & 88.65 (2.7)                                                                & 90.38 (2.76) & 83.78 (3.88)    \\
$\ymt = 0$, $\ysp^{(2)} =0$          & 925        & 85.95 (0.33)                                             & 85.9 (0.44)                                                                & 85.55 (0.56)                                                               & 85.1 (0.41)  & 84.14 (1.08)    \\
Worst-group ($\ymt $, $\ysp^{(2)} $) & -          & 69.04 (2.82)                                             & 71.84 (3.04)                                                               & 74.15 (2.25)                                                               & 63.96 (2.34) & 78.74 (2.81)     \\
\bottomrule
\end{tabular}
\end{table}

\section{Details on Datasets, Models, and Parameter Selection}
\label{sec:param_details}

\subsection{Datasets} \label{app:datasets}

\textbf{Toy:} For a given dataset size  (e.g. $n=$2,000) the data is split into an 80\% training and 20\% validation set, and a test set of the same size is kept apart for evaluation. We describe the data-generating process of this dataset in Section \ref{sec:datasets}.  

\textbf{Waterbirds}: this dataset  from \citet{sagawa_2020B} is a combination of  the Places dataset \citep{zhou_2016} and the CUB dataset \citep{WelinderEtal2010}. A `water background' is set by selecting an image from the lake and ocean categories in the places dataset, and the `land background' is set based on the broadleaf and bamboo forest categories.  A waterbird/land is then pasted in front of the background. When creating new versions of the dataset, we change the $p(\ymt = y | \ysp = y)$, and keep the size of the training set at 4,775 samples, and 1,199 for the validation set. For the test set, we select 5,796 samples where $p(\ymt = y | \ysp = y) = 0.5$.

For this dataset when training ERM or adversarial removal, we sample in each batch such that $p(\ysp = 1 ) = 0.5$. When training JSE, INLP or RLACE, in each batch we sample such that $p(\ysp = 1 ) = 0.5$. When training ERM on the embeddings transformed by JSE, INLP or RLACE, we sample again such that in each batch  $p(\ymt = 1 ) = 0.5$.

\begin{figure}[H]
    \centering
    \includegraphics[scale=0.35]{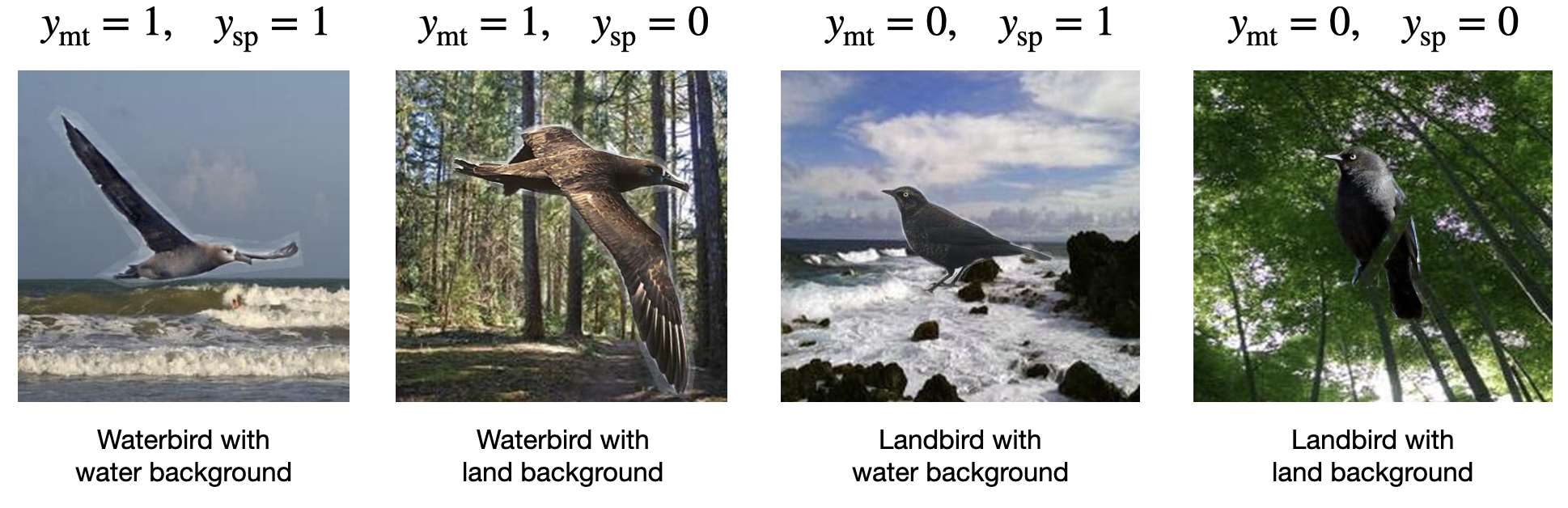}
    \caption{\textbf{Examples of the different images in the Waterbirds dataset}}
    \label{fig:waterbirds_dataset}
\end{figure}

\newpage
\textbf{CelebA}: this dataset contains images of celebrity faces \citep{liu2015faceattributes}. The total size of the dataset is 202,599, from which we sample smaller versions in order to control the strength of the spurious correlation. For these smaller versions, we select 4,500 observations for the training set, 2,000 for the validation set, and 2,000 for the test set. We set the $p(\ymt = y | \ysp = y)$ for the training and validation set, while we set $p(\ymt = y | \ysp = y) = 0.5$ for the test set. We set  $p(\ymt = 1) = 0.5$ for the training, validation and test set.

\begin{figure}[H]
    \centering
    \includegraphics[scale=0.35]{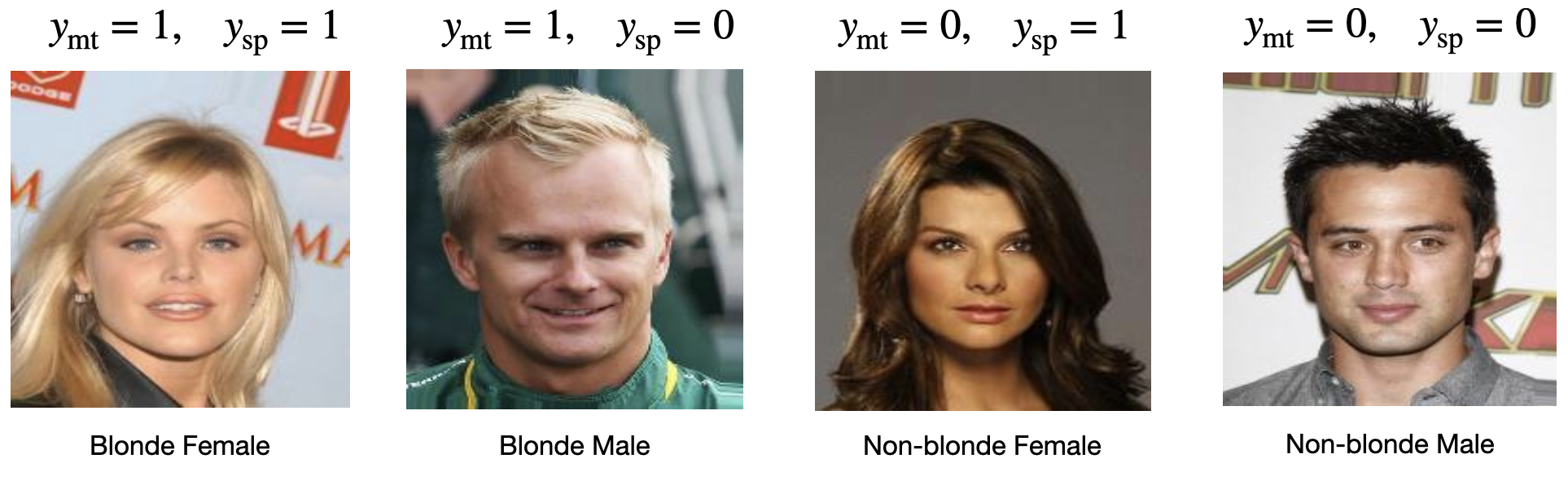}
    \caption{\textbf{Examples of the different images in the CelebA dataset}}
    \label{fig:CelebA_dataset}
\end{figure}

\textbf{MultiNLI}: the MultiNLI dataset \citep{Bowman_MultiNLI} contains pairs of sentences, with examples shown in Table \ref{fig:MultiNLI_examples}. The sentences are pasted together with a [SEP] token in between. We change the dependent variable to a binary label, with $\ymt = 1$ indicating the first sentence (the premise) contradicting the second sentence (the hypothesis), and $\ysp$ otherwise. We chose for the punctuation marks as a spurious correlation rather than the commonly used `negation word' (see for instance \citet{sagawa_2020}) since it is questionable to what extent a negation word such as 'no', 'never' or 'none' is spuriously associated with a contradiction  \citep{joshi2022spurious}. The original dataset contains 206,175 pairs of sentences. We sample smaller versions of the dataset where $p(\ymt = 1) = 0.5$ and change the   $p(\ymt = y | \ysp = y)$ for the training and validation set. For the test set, we set  $p(\ymt = y | \ysp = y) = 0.5$.  The training set contains 50,000 datapoints and the validation and test set each contain 5,000 datapoints.

\begin{table}[H]
\caption{
\textbf{Examples from sentence pairs in the MultiNLI dataset.}}
\label{fig:MultiNLI_examples}
\vskip 0.15in
\small
\centering \tablesize
\begin{tabular}{cccc}
\bottomrule
\textbf{$y_{\mathrm{sp}}$}                              & \textbf{$y_{\mathrm{mt}}$}                              & \textbf{Premise}                                                                                                                                                               & \textbf{Hypothesis}                                                                                                                 \\
\hline 
\cellcolor[HTML]{D0CECE}                    & \cellcolor[HTML]{D0CECE}                    &                                                                                                                                                                                &                                                                                                                                     \\
\multirow{-2}{*}{\cellcolor[HTML]{D0CECE}0} & \cellcolor[HTML]{D0CECE}                    & \multirow{-2}{*}{\begin{tabular}[c]{@{}c@{}}Conceptually cream skimming has \\ two basic dimensions - product and geography.\end{tabular}}                                     & \multirow{-2}{*}{\begin{tabular}[c]{@{}c@{}}Product and geography are what\\  make cream skimming work !!\end{tabular}}             \\
\cline{3-4}
                                            & \cellcolor[HTML]{D0CECE}                    &                                                                                                                                                                                &                                                                                                                                       \\
                                          
\multirow{-2}{*}{1}                         & \multirow{-4}{*}{\cellcolor[HTML]{D0CECE}0} & \multirow{-2}{*}{\begin{tabular}[c]{@{}c@{}}One of our number will carry \\ out your instructions minutely.\end{tabular}}                                                      & \multirow{-2}{*}{\begin{tabular}[c]{@{}c@{}}A member of my team will execute\\  your orders with immense precision.\end{tabular}}   \\
\hline 
\cellcolor[HTML]{D0CECE}                    &                                             &                                                                                                                                                                                &                                                                                                                                     \\
\multirow{-2}{*}{\cellcolor[HTML]{D0CECE}0} &                                             & \multirow{-2}{*}{Fun for adults and children.}                                                                                                                                 & \multirow{-2}{*}{Fun for only children.}                                                                                           \\
 \cline{3-4}
                                            &                                             &                                                                                                                                                                                &                                                                                                                                     \\
\multirow{-2}{*}{1}                         & \multirow{-4}{*}{1}                         & \multirow{-2}{*}{\begin{tabular}[c]{@{}c@{}}This analysis pooled estimates from these  two studies\\ to develop a C-R function linking PM to chronic bronchitis.\end{tabular}} & \multirow{-2}{*}{\begin{tabular}[c]{@{}c@{}}The analysis proves that there is \\ no link between PM and bronchitis !!\end{tabular}}\\
\bottomrule
\end{tabular}

\end{table}

For simplicity, we perform no data augmentation for any of the datasets.  When training logistic regressions on the last-layer representations, we demean the data based on the mean of the training set. This is based on previous work from \citet{rudin_2020}, which states that demeaning is a necessary step before determining concept vectors.

\subsection{Models \& Training Procedure}
\label{sec:training_procedure}

\textbf{Models}: For the Waterbirds and CelebA dataset, we use the ResNet50 architecture implemented in the \verb+torchvision+ package: \verb+torchvision.models.ResNet50(pretrained=True)+. More details on the model can be found in the original paper from  \citet{He_2015}. We finetune the model using the parameters of \citet{kirichenko_2022}. For waterbirds, this means using a learning rate of $10^{-3}$, a weight decay of $10^{-3}$, a batch size of 32, and for 100 epochs without early stopping. For CelebA, this means using a learning rate of $10^{-3}$, a weight decay of $10^{-4}$, a batch size of 128, and for 50 epochs without early stopping. We use stochastic gradient descent (SGD) with a momentum parameter of 0.9. After this, PCA is applied to the embeddings of the layer of the architecture, to reduce the dimensionality from 2048 to 300. 

For the MultiNLI dataset, we use the base BERT model implemented  in the \verb+transformers+ package \citep{huggingface_transformers}: \verb+BertModel.from_pretrained("bert-base-uncased")+. The model was pre-trained on BookCorpus, a dataset consisting of 11,038 unpublished books, as well as the English Wikipedia (excluding lists, tables and headers). More details on the model can be found in the original paper: \citet{devlin-etal-2019-bert}.  For finetuning the BERT model on MultiNLI we use early stopping, and stop the procedure if we observe no improvement after 1 epoch. We use the Adam optimizer \citep{Kingma_adam} with the standard settings in \textit{Pytorch}. When finetuning, we train for a maximum of 10 epochs, use a batch size of 32, a learning rate of $10^{-5}$, and a weight decay of $10^{-4}$.

\textbf{Training procedure of concept-removal methods}: For JSE, ERM INLP and RLACE, we  use early stopping to prevent overfitting. If we observe no improvement on the validation set for a certain model after 5 epochs, the training procedure is stopped and the model with the lowest loss on the validation set is selected. We use stochastic gradient descent (SGD) with a momentum parameter of 0.9 for  JSE, ERM INLP and RLACE, and train for a maximum of 50 epochs. 


\subsection{Implementation Details \& Parameter Selection}

We only assume access to a validation dataset that follows the same distribution as the training dataset. This means that the conditional probability $p(y_{\mathrm{mt}} = y | y_{\mathrm{sp}} = y)$ is the same across the training and validation set. For experiments where we change the  $p(y_{\mathrm{mt}} = y | y_{\mathrm{sp}} = y)$ in the training set, we do not select new parameters for each case, but select them based on the scenario where  $p(y_{\mathrm{mt}} = y | y_{\mathrm{sp}} = y) = 0.9$.

For JSE, ERM, INLP, GDRO, JTT, ERM-GW and SUBG we set the batch size at 128. If we work with these methods with the smaller datasets in Section \ref{sec:instance_reweighting_limited}, we set the batch size to 64. We also do this for RLACE, except in the case of the MultiNLI dataset - we only observed convergence for this dataset and method with a batch size of 512.  After setting the batch size, we select the best combination of the learning rate and weight decay.  For the learning rate, we assess the values $10^{-1}$, $10^{-2}$, $10^{-3}$ and $10^{-4}$. For the weight decay, we assess the values 0,  $10^{-3}$, $10^{-2}$, $10^{-1}$ and $1$. In the case of the Toy dataset we always set the weight decay to 0. For each method,  we use the weighted binary cross-entropy on the validation set to measure performance. For the Toy dataset, the performance is measured across 10 runs, and for the  Waterbirds, CelebA and multiNLI dataset the performance is measured across 5 runs, each time with a different finetuned model. 

Below, we detail how the parameters were selected for each method, as well as implementation details for RLACE and adversarial removal. The selected combinations of the learning rate and weight decay can be found in Table \ref{tab:hyper_param_selection}. 

\textbf{ERM}: We select the learning rate and weight-decay combination that has the best performance for the main-task labels. These parameters are also used when fitting a logistic regression on the transformed representations from JSE, INLP or RLACE. We keep the parameters the same for group-weighted ERM and subgroup sampling. 

\textbf{INLP}:  we select the combination of parameters that has the best performance for the spurious concept labels, based on the first spurious concept vector found by INLP. We continue projecting out the spurious concept vectors found by INLP until the accuracy of the spurious concept classifier is no better than a majority rule classifier. Whether or not the BCE is statistically significantly different from that of a majority rule classifier is tested via an $t$-test of the difference in the BCE's for both classifiers, where the critical value is determined based on $\alpha = 0.05$. 

\textbf{JSE}: we select the combination of parameters that has the best performance for the spurious concept and main-task concept labels, weighing each equally. This performance is based on the first set of spurious and main-task concept vectors. The critical value of the tests in Section \ref{sec:details_tests} is determined based on $\alpha=0.05$. 

\textbf{LEACE}: we use the original code from \citet{belrose2023leace} for implementation of the method.  

\textbf{RLACE}:  we use the original code from \citet{rafvogel_2022}, and run the algorithm for a maximum of 50,000 iterations. For the spurious concept classifier and optimizing the projection matrix,  we use the same parameters as INLP. We optimize the projection matrix until the accuracy of the classifier is lower than 51\%. Similar to \citet{rafvogel_2022}, in each case we find a matrix of rank 1.

\textbf{Adversarial removal}:  The weight of the adversary loss is set to $\lambda = 1$. The entire architecture is finetuned, and the adversary is trained using the gradient reversal method \citep{pmlr-v37-ganin15}.  For the vision datasets, we observe that the accuracy of the adversary converges to below 55\%, which is commonly accepted as success for the method. After the adversarial removal method, we apply standard ERM including PCA, as for the other methods.  For MultiNLI, we experimented extensively with hyper-parameters in order to get both a high accuracy on the main task, and the accuracy of the adversary to converge to below 55\%. We did not observe this, even after lowering the weight of the adversary loss from 1 to $10^{-1}$ or even $10^{-2}$. We used the Adam optimizer when performing adversarial removal.

\textbf{GDRO:} we fix the hyper-parameter $\eta_g =0.1$, similar to \citet{idrissi2022simple}. We grid-search the best combination of the learning rate, weight decay and the hyper-parameter of $C$, with possible values of 0, 1, 2, 3, 4, 5.  We provided initial parameters for the GDRO model based on an fitted ERM model. 

\textbf{JTT:} We grid-search the best combination of the learning rate, weight decay and hyper-parameter of $\lambda$, which is the weight for the misclassified samples. For $\lambda$, we consider the values of 2, 5, 10, and 25.

\begin{table}[H]
\caption{\textbf{Selected combinations of learning rate and weight decay.} For the $C$ parameter of GDRO, we use the values $1, 2, 5$ and $5$ for respectively the Waterbirds, CelebA, multiNLI and Toy datasets. For the $\lambda$ parameter of JTT, we use the values of $2, 5, 10, 2$ for respectively the Toy, Waterbirds, CelebA, and multiNLI datasets.}
\label{tab:hyper_param_selection}
\vskip 0.15in
\centering \tablesize
\setlength{\extrarowheight}{3pt}

\begin{tabular}{cl|rr}
\toprule
\multicolumn{1}{l}{\textbf{Dataset}} & \textbf{Method} & \textbf{Learning Rate}     & \textbf{Weight Decay}      \\
\hline
\multirow{4}{*}{Toy}                 & JSE             & $10^{-2}$                       & 0                        \\
                                     & ERM             & $10^{-1}$                        & 0                        \\
                                     & INLP            & $10^{-1}$                        & 0                        \\
                                     & RLACE           & $10^{-1}$                        & 0                        \\
                                     & GDRO            &  $10^{-3}$    & 0  \\
                                     & JTT             &  $10^{-3}$  & 0 \\
                                     \hline
\multirow{5}{*}{Waterbirds}          & JSE             & $10^{-3}$                      & $10^{-3}$                      \\
                                     & ERM             & $10^{-2}$                       & $10^{-2}$                       \\
                                     & INLP            & $10^{-2}$                        & $10^{-3}$                      \\
                                     & RLACE           & $10^{-2}$                        & $10^{-3}$                      \\
                                     & ADV             & $10^{-3}$ & $10^{-4}$\\
                                       & GDRO            &  $10^{-3}$     & $10^{-1}$   \\
                                     & JTT             &  $10^{-2}$  &  $10^{-2}$\\
                                     \hline
\multirow{5}{*}{CelebA}              & JSE             & $10^{-2}$                       & $10^{-3}$                      \\
                                     & ERM             & $10^{-2}$                       & $10^{-2}$                       \\
                                     & INLP            & $10^{-2}$                       & $10^{-3}$                      \\
                                     & RLACE           & $10^{-2}$                       & $10^{-3}$                      \\
                                     & ADV             & $10^{-3}$ & $10^{-4}$\\
                                       & GDRO            &  $10^{-3}$    &  $10^{-3}$  \\
                                     & JTT             &  $10^{-2}$  & $10^{-2}$ \\
                                     \hline
\multirow{4}{*}{MultiNLI}            & JSE             & $10^{-2}$                       & $10^{-2}$                       \\
                                     & ERM             & $10^{-2}$                       & $1$                          \\
                                     & INLP            & $10^{-2}$                       & $10^{-3}$                      \\
                                     & RLACE           & $10^{-2}$                       & $10^{-2}$   \\
                                       & GDRO            &   $10^{-4}$   & $1$  \\
                                     & JTT             &  $10^{-4}$  & $10^{-3}$ \\
                                     \bottomrule
\end{tabular}

\end{table}

\end{document}